\RequirePackage{fix-cm}
\documentclass[twocolumn]{svjour3} 
\smartqed  
\usepackage[misc]{ifsym}
\usepackage{cite}
\usepackage{colortbl}
\usepackage{color}
\usepackage{bbm}
\usepackage{booktabs}
\usepackage{multirow}
\usepackage{array}
\usepackage{lipsum}
\usepackage{adjustbox}
\usepackage{graphicx}
\usepackage{amsmath}
\usepackage[colorlinks, linkcolor=red, anchorcolor=blue, citecolor=green]{hyperref}
\usepackage{float}
\usepackage{subfloat}
\usepackage{rotating}
\usepackage{makecell}
\usepackage[abs]{overpic}
\usepackage[noend]{algpseudocode}
\usepackage{algorithmicx,algorithm}
\usepackage{amssymb}
\usepackage{nccmath}
\usepackage{pifont}
\usepackage{multirow}
\usepackage{appendix}
\usepackage{placeins}
\usepackage{colortbl}
\usepackage{rotating}
\usepackage{array}
\usepackage{color}
\usepackage{xcolor}
\usepackage{pifont}

\usepackage{amsmath,amssymb,amsthm}
\usepackage{verbatim}
\usepackage{units}
\usepackage{textcomp}
\usepackage{bbding}
\usepackage{mwe}
\usepackage{textcomp}
\usepackage{bbding}
\usepackage{mwe}
\usepackage{subcaption}
\usepackage{xr}
\usepackage{tikz}
\usepackage{pgfplots}
\usepackage[rightcaption]{sidecap}
\definecolor{darkred}{rgb}{0.7, 0, 0}
\definecolor{darkblue}{rgb}{0.0, 0.0, 0.7}
\definecolor{darkgreen}{rgb}{0, 0.4, 0}
\newcommand{\cmark}{\textcolor{darkgreen}{\ding{51}}}%
\newcommand{\xmark}{\textcolor{darkred}{\ding{55}}}%
\newcommand{\Seg}{\mathcal{S}}
\newcommand{\Dis}{\mathcal{D}}
\newcommand{\Gen}{\mathcal{G}}
\pgfplotsset{compat=1.17}
\begin{document}

\title{USIS: Unsupervised Semantic Image Synthesis}


\author{George Eskandar \and Mohamed Abdelsamad \and Karim Armanious  \and Bin Yang}


\institute{George Eskandar, Mohamed Abdelsamad , Karim Armanious, Bin Yang \at
            Institute of Signal Processing and System Theory, University of Stuttgart, Stuttgart, Germany\\
            \email{george.eskandar@iss.uni-stuttgart.de} 
}

\date{Received: date / Accepted: date}
\maketitle
\begin{sloppypar}
\begin{abstract}
Semantic Image Synthesis (SIS) is a subclass of image-to-image translation where a photorealistic image is synthesized from a segmentation mask. SIS has mostly been addressed as a supervised problem. However, state-of-the-art methods depend on a huge amount of labeled data and cannot be applied in an unpaired setting. On the other hand, generic unpaired image-to-image translation frameworks underperform in comparison, because they color-code semantic layouts and feed them to traditional convolutional networks, which then learn correspondences in appearance instead of semantic content. In this initial work, we propose a new Unsupervised paradigm for Semantic Image Synthesis (USIS) as a first step towards closing the performance gap between paired and unpaired settings. Notably, the framework deploys a SPADE generator that learns to output images with visually separable semantic classes using a self-supervised segmentation loss. Furthermore, in order to match the color and texture distribution of real images without losing high-frequency information, we propose to use whole image wavelet-based discrimination. We test our methodology on 3 challenging datasets and demonstrate its ability to generate multimodal photorealistic images with an improved quality in the unpaired setting.  
\begin{figure*}
    \captionsetup[subfloat]{position=top, labelformat=empty, skip=0pt}
    
    \centering
    \begin{minipage}{1.0\textwidth}
    \subfloat[Label]{\includegraphics[width=  0.24\textwidth, height=   0.12\textwidth]{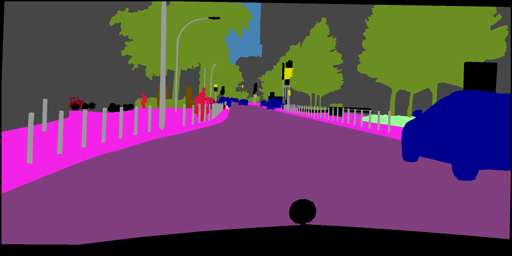}} \hfill
    \subfloat[CUT\cite{park2020cut}]{\includegraphics[width=  0.24\textwidth, height=   0.12\textwidth]{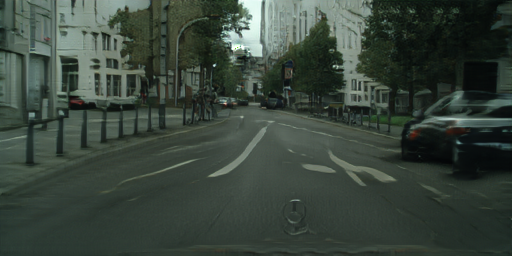}} \hfill
    \subfloat[USIS]{\includegraphics[width=  0.24\textwidth, height=   0.12\textwidth]{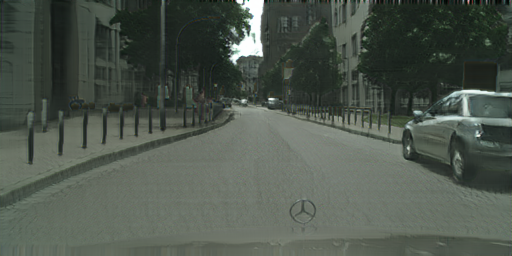}} \hfill
    \subfloat[OASIS\cite{schonfeld2021you}]{\includegraphics[width= 0.24\textwidth, height=   0.12\textwidth]{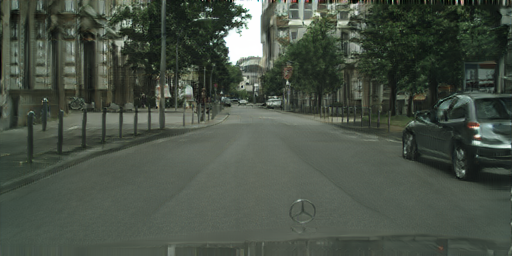}} \vfill
    
    \subfloat[]{\includegraphics[width=  0.24\textwidth, height=   0.12\textwidth]{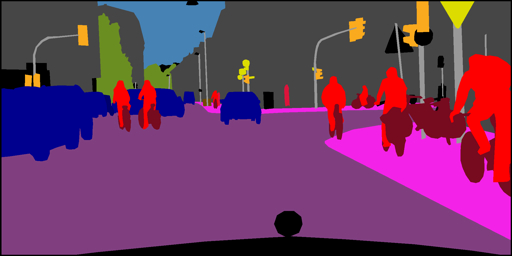}} \hfill
    \subfloat[]{\includegraphics[width=  0.24\textwidth, height=   0.12\textwidth]{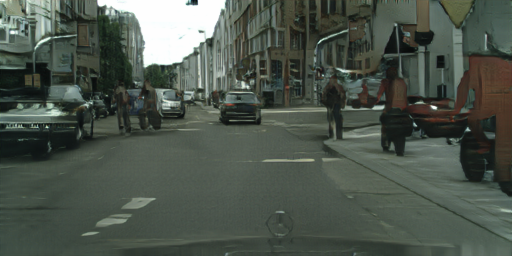}} \hfill
    \subfloat[]{\includegraphics[width=  0.24\textwidth, height=   0.12\textwidth]{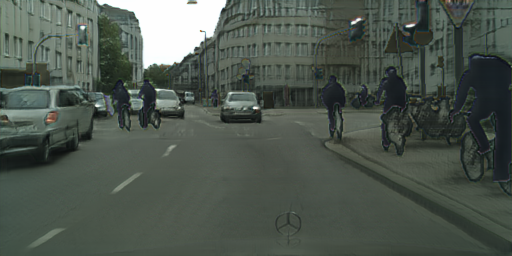}} \hfill
    \subfloat[]{\includegraphics[width= 0.24\textwidth, height=   0.12\textwidth]{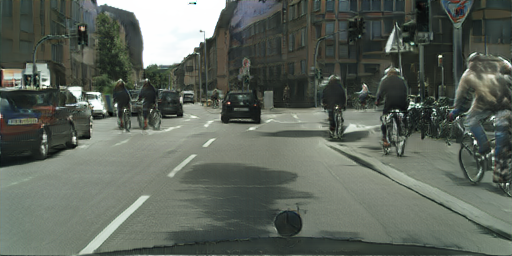}} \vfill
    \end{minipage}
   \caption{Our model USIS is a first step towards bridging the performance gap between paired and unpaired image-to-image translation in the context of SIS. CUT\cite{park2020cut} is the state-of-the-art in unpaired GANs while OASIS\cite{schonfeld2021you} is the state-of-the-art in paired SIS.}
   \label{fig:teaser}
\end{figure*}

\end{abstract}
\vspace{-0.5cm}
\section{Introduction}
\label{sec:1}
\vspace{-1em}
Semantic image synthesis (SIS) is the task of generating high resolution images from user-specified semantic layouts. It is a recent application of Generative Adversarial Networks (GANs) that was introduced by Pix2PixHD\cite{wang2018high} in 2017. In 2019, Spatially Adaptive Normalization or SPADE\cite{park2019semantic} was proposed as a better alternative generator design for the task and since then, the field has significantly grown. SIS opens the door to an extensive range of applications such as content creation and semantic manipulation by editing, adding, removing or changing the appearance of an object. By allowing concept artists and art directors to brainstorm their designs efficiently, it can play a pivotal role in graphics design. In addition, SIS can be used as a data augmentation tool for deep learning models, by generating training data conditioned on desired scenarios which might be hard to capture or reproduce in real-life (example: edge cases in autonomous driving like accidents).\par
In contrast to graphics engine, semantic image synthesis doesn't need either specialized training to use or intricate information like 3D Geometry, materials or light transport simulation\cite{10.5555/3044800}, because they learn directly from the collected real data. Moreover, a problem commonly seen in graphic engines is that the synthesized images look visually different from real data and models trained only on synthetic images do not generalize well due to the domain gap between the two data distributions\cite{Sankaranarayanan2018LearningFS}. That is why SIS can be of a significant advantage because it bypasses these problems. 
However, the problem of semantic image synthesis has mostly been addressed as a supervised learning problem. Although state-of-the-art methods\cite{schonfeld2021you} can produce visually appealing high resolution images, they suffer from several drawbacks. Most importantly, they depend on a lot of annotated paired data which is expensive and time-consuming to acquire: the average annotation time for one frame in the Cityscapes dataset\cite{cordts2016cityscapes} is 1 hour. Furthermore, as labeled datasets are usually smaller than unlabeled datasets, supervised training restricts the generator's learned distribution to the distribution of real images which have labels. Unpaired training allows the usage of a larger number of real images, which exhibit a greater variation. 

Unpaired conditional GAN frameworks\cite{zhu2017unpaired, huang2018multimodal, lee2018diverse, park2020cut, fu2019geometry, benaim2017one} can be used for SIS, but they suffer from several drawbacks: (1) these models approximate labels as images by color-coding each class in the semantic layout, which creates an artificial mapping between labels and images, (2) the unsupervised losses in the state-of-the-art force relationships between the labels and images that do not preserve the semantic content in the case of SIS and (3) the normalization layers in the architecture of unsupervised models wash away the semantic labels as noted in \cite{park2019semantic}. Consequently, the generated samples from these models suffer from poor quality. Another downside is the inability to generate realistic images when the number of classes in the training dataset is too big.

In this work, we propose an unsupervised framework which can synthesize realistic images from labels without the use of paired data. To our knowledge, this is the first paper to address this topic explicitly. The USIS or Unsupervised Semantic Image Synthesis framework can be trained on any 2 unpaired datasets generalizing the use of SIS. For instance, the labels from GTA-V dataset\cite{Richter_2016_ECCV} could be used in training along with realistic images from Cityscapes\cite{cordts2016cityscapes} or KITTI\cite{Geiger2012CVPR}. This is not possible in the supervised setting, as the models trained on GTA-V labels can only produce GTA-V like images. By virtue of its design, the unpaired setting can help eliminate dataset biases and push the model towards a better multimodal generation. The generated samples along with their labels can thus be used as a data augmentation technique for semantic segmentation models. USIS also performs better on datasets with a large number of classes.

In the following, we first review the related works to our proposed framework. Then, after going through some preliminaries, a formal definition of the USIS task is given and some typical problems in the previous unpaired GANs are exposed and analyzed. An unsupervised paradigm for SIS is introduced, which involves an adversarial training between a generator and a whole image wavelet-based discriminator, and a cooperative training between the generator and a UNet\cite{schonfeld2020u} segmentation network. More precisely, the discriminator fosters the generator to match the distribution of the real images dataset by making a real/fake decision while the Unet gives a pixel-level feedback to the generator, by classifying each pixel in the generated image into N classes using a cross entropy loss. This self-supervised segmentation loss encourages the generator to produce images that are semantically aligned with the labels and prevents the bias towards generating classes with bigger areas at the expense of fine and small classes. In addition, upon observing that convolutional networks are biased towards low-frequencies\cite{Chen2021SSDGANMT, Durall2020WatchYU, Dzanic2020FourierSD}, we provide the wavelet decomposition of the real and fake images as input to our discriminator so that it makes a decision based on higher spatial frequencies in the image. Next, extensive experiments on 3 image datasets (COCO-stuff\cite{caesar2018coco}, Cityscapes\cite{cordts2016cityscapes} and ADE20K\cite{zhou2017scene}) in an unpaired setting are conducted to showcase the ability of our model to generate a high diversity of photorealistic images and close the gap between supervised and unsupervised methods in SIS. Various ablation studies on the components of our model are subsequently performed. Finally, we show the performance of our model on a practical use case by generating high quality Cityscapes from GTA-V labels.   

\section{Related Works}
\label{sec:2}
In this section, we will review the related works to our research.

{\bf Generative Adversarial Networks (GANs)} GANs\cite{Goodfellow2014GenerativeAN} can be trained to generate images. In GANs, two networks compete against each other in a minimax game. A generator tries to fool the discriminator into classifying the generated outputs as real. The past 5 years have witnessed great advances in the quality and resolution of the generated images. ProGAN \cite{Karras2018ProgressiveGO} introduced the concept of progressive growing where GANs start with a few layers and are trained with low-resolution images and as the training continues more layers are added progressively to reach a higher resolution. StyleGAN\cite{Karras2019ASG} built upon ProGAN and fed style information in each layer of the generator to control visual features at different scales, using an adaptive instance normalization layer (AdaIN)\cite{Huang2017ArbitraryST}. StyleGANv2 \cite{Karras2020AnalyzingAI} redesigned several aspects in the architecture like weight demodulation, path length regularization and progressive growing to improve upon StyleGAN. In all these models, the input to the generator is usually a random vector that follows a normal distribution and thus these models offer very little controllability in the generation process. Conditional GANs (cGANs) by contrast synthesize images based on a user-specified condition. Examples for conditions are class-labels\cite{Mescheder2018WhichTM, miyato2018cgans, brock2018large}, text\cite{Reed2016GenerativeAT, xu2018attngan, hong2018inferring} or other images\cite{isola2017image, wang2018high, park2019semantic, zhu2017unpaired, huang2018multimodal, tang2019local}.

{\bf Supervised Semantic Image Synthesis} is an image generation task where the condition is a semantic mask. The task was first introduced by Pix2pix\cite{isola2017image}. The semantic map was color-coded and fed to an encoder-decoder architecture. A PatchGAN discriminator classifies overlapping patches in the generated images as real or fake. Chen et al. \cite{chen2017photographic} proposed to use cascaded refinement networks and perceptual losses\cite{Johnson2016Perceptual, Gatys2016} with a pretrained VGG network\cite{Simonyan15} for the task. Pix2PixHD\cite{wang2018high} further improved the quality of generated results by employing feature matching losses to stabilize GAN training, a multiscale discriminator and a more sophisticated generator architecture. But the breakthrough came with SPADE\cite{park2019semantic} which realised the inadequacy of using normalization layers with semantic labels. To remedy the issue, Park et al.\cite{park2019semantic} proposed to condition the modulation parameters of the normalization layers the semantic layout. Moreover, the parameters vary spatially. Other choices have also been proposed to remedy the issue. For isntance, CLADE\cite{tan2020rethinking} proposed to use class-adaptive modulation parameters instead. CC-FPSE\cite{liu2019learning} employed spatially-varying convolutional weights instead of the spatially-varying normalization layers. SEAN\cite{Zhu2020SEANIS} used the SPADE layer but redesigned the network to add more controllability so to edit the style of each semantic region individually. Similar to advances in the generator architecture, various improvements in the discriminator architecture have been proposed, even though perceptual losses was a standard in all these frameworks. OASIS\cite{schonfeld2021you} were the first to utilize a Unet\cite{Ronneberger2015UNetCN}-based discriminator to improve the semantic image synthesis task. This "segmentation" discriminator, previously used to improve semantic segmentation\cite{souly2017semi} or unconditional image generation\cite{schonfeld2020u}, tries to classify each pixel of real images into its semantic class and generated images as fake.

{\bf Unpaired Image-to-Image translations} is a conditional image generation task where it is either impossible or expensive to collect paired data. There has been two main approaches to solve this problem: using a cycle consistency loss or imposing a relationship preservation constraint. Cycle consistency aims to find correspondences between the input and output, by learning the inverse mapping and reconstructing the input\cite{zhu2017unpaired, yi2017dualgan}. The CycleGAN\cite{zhu2017unpaired} framework first introduced this approach and consisted of two generators (forward and inverse mapping) and two discriminators (one for each dataset). Research in this area has leveraged cycle consistency losses to allow for many-to-many mappings\cite{choi2017stargan}, multimodal mapping between two domains\cite{lee2018diverse, liu2017unsupervised, huang2018multimodal, almahairi2018augmented} and an improved generation quality\cite{gokaslan2018improving, liang2018generative, tang2019attentiongan, wu2019transgaga, zhang2019harmonic}. In many of these frameworks, the image data wass assumed to be generated from a content and a style latent variables. MUNIT\cite{huang2018multimodal} mapped an image from Domain A to Domain B by combining its content code and a sampled style code from Domain B. Cycle losses were applied not on images but rather on the latent codes. However, the problem with cycle losses, is that they assume that a mapping from one domain to another is a bijection which is often a restrictive assumption. On the other hand, some works\cite{taigman2017unsupervised, shrivastava2017learning, bousmalis2017unsupervised, benaim2017one, amodio2019travelgan, fu2019geometry} have approached unpaired I2I translation by imposing a relationship preservation constraint. If $x_{1}$ and $x_{2}$ are 2 images in Domain A with a certain relationship $\mathcal{R}$, their mappings in Domain B $G(x_{1})$ and $G(x_{2})$ should have the same $\mathcal{R}$. The relationship preservation constraint doesn't have to happen only on an image level, it can also occur between patches of the same image\cite{zhang2019harmonic, park2020cut}. In some works, the relationship constraint is a predefined distance loss (content losses\cite{taigman2017unsupervised, shrivastava2017learning, bousmalis2017unsupervised} or geometric constraints\cite{fu2019geometry}), in others it is based on a contrastive loss like in \cite{amodio2019travelgan, park2020cut}. Our work can be considered as both a cycle-consistency loss and a relationship preservation constraint on a pixel-level. 

{\bf Frequency-based Approaches in Deep Learning} have gained more attention in recent years. The bias of convolutional neural networks (CNNs) towards low-frequency has been studied in several works\cite{Chen2021SSDGANMT, Durall2020WatchYU, Dzanic2020FourierSD}. More specifically, in \cite{Chen2021SSDGANMT} it has been observed that the discriminator is missing high frequency information due to the downsampling operations in the network architecture while in \cite{Durall2020WatchYU}, it was found out that upsampling layers in the generator cannot reconstruct the spectral distribution of the real data. As a remedy to these issues, high frequency representations of images like 2D-Fourier transform\cite{bracewell1986fourier} and Haar-Wavelet transform\cite{Daubechies1990TheWT, doi:10.1137/1.9781611970104} have been used with neural networks either as components in the architecture\cite{Gao2016AHW, Liu2019MultiLevelWC, Williams2018WaveletPF, Yoo2019PhotorealisticST, Liu2020WaveletBasedDN, Kang2017ADC} in several generative applications like image super-resolution, image denoising or style-transfer. Other works have completely based their generative networks on high-frequency representation. For instance, \cite{Liu2019AttributeAwareFA} proposed a discriminator that observes the wavelet decomposition of generated and real images while \cite{Zhang2019SuperresolutionRA} proposed a generator that works entirely in the wavelet domain. Some works used the wavelet representation in both the generator and discriminator.  MW-GAN\cite{Wang2020MultilevelWG} proposed a multi-level wavelet generator and a discriminator that evaluates the images in the spatial and wavelet domains. WaveletSRGAN\cite{Huang2019WaveletDG} presented a wavelet generator, a pixel-level discriminator and a wavelet-level discriminator for super-resolution task. SWAGAN\cite{Gal2021SWAGANAS} was the first unconditional GAN to use a wavelet architecture in the generator and discriminator with progressive growing like StyleGANv2\cite{Karras2020AnalyzingAI} and has shown promising results. We notice that the wavelet representation has been consistently preferred over the Fourier representation in deep learning models, due to the fact that the wavelet decomposition offers simultaneous information in both the space and spectral domains making it more suitable to CNNs. Wavelet decomposition also offers a multi-resolution analysis and has a faster computation time than the Fourier Transform. However, none of the previously mentioned works used wavelet representations in an unsupervised setting: the wavelet decomposition of the groundtruth is always available for reconstruction. Also our framework is different because our input consists of label maps whose wavelet transform has little or no semantic value.         
\vspace{-0.8cm}

\section{Preliminaries}
\label{sec:3}
In this section, we briefly review some fundamental concepts about semantic image synthesis and wavelet transform.
\vspace{-0.3cm}
\subsection{The SPADE Baseline}
\label{sec:31}
We briefly review the SPADE\cite{park2019semantic} baseline. SPADE is a GAN which consists of two components: (1) a generator with a decoder like structure which cascades several ResNet blocks\cite{He2016DeepRL} with upsampling layers in between, and (2) a multiscale Patch-discriminator. The input to the generator is a random vector sampled from a multivariate Gaussian distribution while the semantic map is fed to the SPADE layer in each ResBlock after being downsampled to the corresponding resolution. Normalization layers are replaced by the SPatially Adaptive DEnormalization layers (SPADE), where the features $f$ coming from convolutional layers are first normalized per channel and modulated with spatially variant learned parameters from semantic maps. More concretely, the semantic maps pass through a few convolutional layers to produce two tensors: $\gamma$ and $\beta$. After normalization, the output $o$ of SPADE is:
\begin{equation}
\begin{aligned}
\label{spade_eq}
o_{n,c,i,j} &= \gamma_{c,i,j}(\mathbf{m}) \frac{f_{b,c,i,j}-\mu_{c}}{\sigma_{c}} + \beta_{c,i,j}(\mathbf{m})\\
\mu_{c} &= \frac{1}{BHW}\sum_{b,i,j} f_{b,c,i,j}\\
\sigma_{c} &= \sqrt{\frac{1}{BHW} \sum_{b,i,j} ((f_{b,c,i,j})^{2}-(\mu_{c})^{2})}
\end{aligned}
\end{equation}
where $\mathbf{m}$ is the semantic layout, $B$, $H$ and $W$ are the batchsize, height and width of the features respectively and $b \in B, c \in C, i \in H, j \in W$. The losses of SPADE are similar to Pix2PixHD\cite{wang2018high} and consist of a GAN loss and a feature matching loss\cite{Alexey2016, Gatys2016, Johnson2016Perceptual} for each discriminator, and a perceptual loss based on a pretrained VGG-network on Imagenet\cite{deng2009imagenet}.   
\vspace{-0.5cm}
\subsection{The Wavelet Transform}
\label{sec:32}
Our work is strongly based on the Wavelet Transform, which passes the image through a series of low-pass and high-pass filters to generate Haar-Wavelet coefficients. The Haar-Wavelet is a family of functions that form an orthonormal basis which can represent discrete signals. One-level Wavelet decomposition generates 4 subbands of lower resolution: an LL frequency subband, which is a blurred version of the image, and 3 high frequencies subbands: LH,HL and HH which represent higher frequencies in horizontal, vertical and diagonal directions respectively. The Wavelet-Transform can be applied recursively to the LL sub-band to generate 4 more subbands at a smaller resolution, thus offering a multi-frequency multi-resolution analysis of the image. In deep learning, there has been 2 main ways to exploit wavelets: either by employing a multi-level decomposition and a reconstruction loss\cite{Huang2019WaveletDG}, or by using only one-level decomposition at a time and progressively growing the network like in \cite{Gal2021SWAGANAS}.   

In this work, we seek to synthesize an image from a semantic map in an unsupervised way. Since there is no direct feedback signal from the groundtruth, the network can favour the generation of big classes (streets, buildings) over small ones (pedestrians, traffic signs) to minimize the GAN objective. Providing the network with a high-frequency representation is advantageous for the unpaired setting because small objects in the image domain have coefficients with bigger magnitude in the wavelet domain. By exploiting this property in the proposed framework we seek to accomplish two objectives: (1) to generate more fine-details and refine the texture of bigger semantic classes and (2) to foster the generation of smaller semantic classes.   
\vspace{-0.5cm}
\section{Problem Definition}
\label{sec:4}
In this section, a formal definition of the SIS task is presented along with some common problems found in the previous unpaired baselines; all of which lays the groundwork for the proposed model in then next section.  

In SIS, we seek to synthesize an RGB image $\textbf{x}$ from a semantic mask $\textbf{m}$ with $C$ labels in an unsupervised way. Let $i$ denote the pixel position. The images are normalized between 0 and 1, and $\textbf{m}$ is one-hot encoded. The problem can be broken down to two tasks:
\begin{itemize}
    \item Class appearance matching: how will the discriminator know the correspondence between the semantic class in the segmentation map and its appearance (texture) in the image without supervision? and how to learn this mapping in a multimodal way? 
    \item Semantic alignment: how can the generator preserve the content/geometrical structure of the segmentation map without ignoring small classes ? 
\end{itemize}
The reason why unpaired GANs are suboptimal is that they color-code the segmentation map $\textbf{m}$ and feed to the network as an RGB image with values between 0 and 1. This way they learn a mapping between color information $(\textbf{m}_{i,j} \in [0, 1]  \longrightarrow \textbf{x}_{i,j} \in [0, 1])$ instead of learning a mapping between a semantic label and color information $( \textbf{m}_{i,j} \in \mathbb{L}=\{1,2,..., C\}  \longrightarrow \textbf{x}_{i,j} \in [0, 1])$. 

Additionally, the 2 main paradigms of unpaired GANs presented in Section \ref{sec:2}, cycle-consistency and relationship preservation, suffer from some issues when applied in SIS. Cycle-consistency Mean Absolute Error (MAE) or Mean Squared Error (MSE) losses can preserve alignment between the segmentation map and the image but might lose semantic information (for example, buildings are generated instead of trees or sky). Furthermore, in SIS, the reverse cycle (RGB to image to RGB) is redundant because it might not always produce segmentation maps, but rather copies the same RGB image with the texture or style of semantic layouts. On the other hand, relationship preservation constraints offer better multimodality (reflected in the FID score) but still suffer from the same problem (good spatial alignment with loss of semantic information). The relationship preservation is usually a constraint imposed between the input label and the generated image in the form of a predefined distance\cite{benaim2017one} or a contrastive loss on the features of an encoder network\cite{park2020cut}. For instance, CUT\cite{park2020cut} maximizes the mutual information between features extracted from corresponding patches in input and output. However, it has been shown\cite{Geirhos2019ImageNettrainedCA} that CNNs are biased toward the texture of the image rather than the shape. This way, we see that color information originating from the color-coding of the classes gets leaked into the features of the encoder and affects the contrastive loss. Finally, a common problem in both approaches is that the normalization layers washout the semantic information\cite{park2019semantic} and convolutional layers are biased towards low-frequency. 

\begin{figure*}
\captionsetup[subfigure]{position=bottom, labelformat=parens, skip=2pt}

\begin{subfigure}{0.3\textwidth}
\begin{minipage}{.48\textwidth}
\includegraphics[width= 1.0\textwidth, height=  1.0\textwidth]{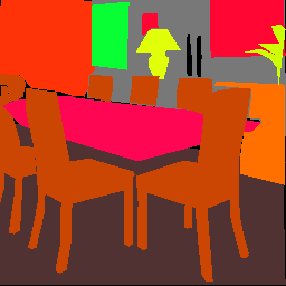}
\end{minipage}
\hfill
\begin{minipage}{0.48\textwidth}
\includegraphics[width= 1.0\textwidth, height=  1.0\textwidth]{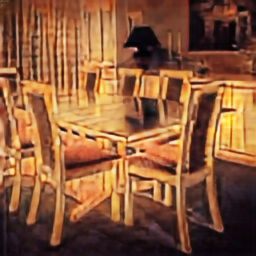}
\end{minipage}
\subcaption{Unrealistic color synthesis}
\end{subfigure}
\qquad
\begin{subfigure}{0.3\textwidth}
\begin{minipage}{.48\textwidth}
\includegraphics[width= 1.0\textwidth, height=  1.0\textwidth]{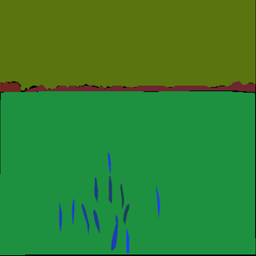}
\end{minipage}
\hfill
\begin{minipage}{0.48\textwidth}
\includegraphics[width= 1.0\textwidth, height=  1.0\textwidth]{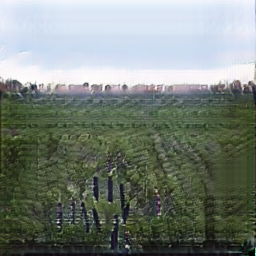}
\end{minipage}
\subcaption{Unrealistic texture synthesis}
\end{subfigure}
\qquad
\begin{subfigure}{0.3\textwidth}
\begin{minipage}{.48\textwidth}
\includegraphics[width= 1.0\textwidth, height=  1.0\textwidth]{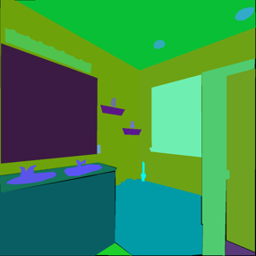}
\end{minipage}
\hfill
\begin{minipage}{0.48\textwidth}
\includegraphics[width= 1.0\textwidth, height=  1.0\textwidth]{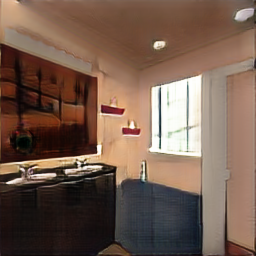}
\end{minipage}
\subcaption{Textureless synthesis}
\end{subfigure}
\vfill
\begin{subfigure}{0.48\textwidth}
\begin{minipage}{.48\textwidth}
\includegraphics[width= 1.0\textwidth, height=  0.5\textwidth]{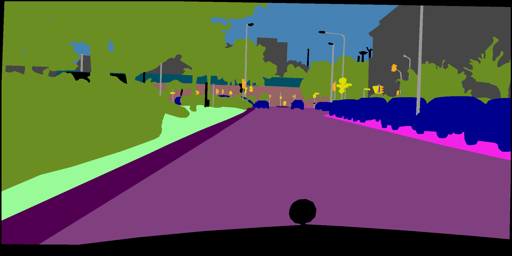}
\end{minipage}
\hfill
\begin{minipage}{0.48\textwidth}
\includegraphics[width= 1.0\textwidth, height=  0.5\textwidth]{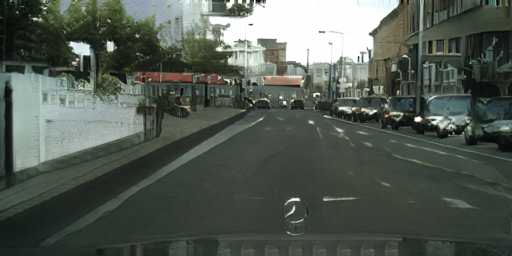}
\end{minipage}
\subcaption{Class mix}
\end{subfigure}
\qquad
\begin{subfigure}{0.48\textwidth}
\begin{minipage}{.48\textwidth}
\includegraphics[width= 1.0\textwidth, height=  0.5\textwidth]{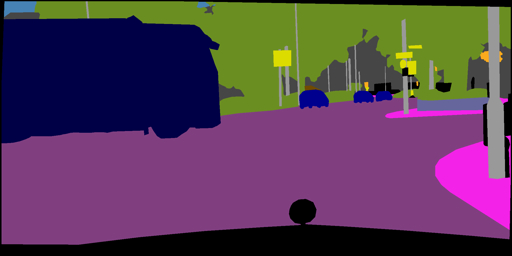}
\end{minipage}
\hfill
\begin{minipage}{0.48\textwidth}
\includegraphics[width= 1.0\textwidth, height=  0.5\textwidth]{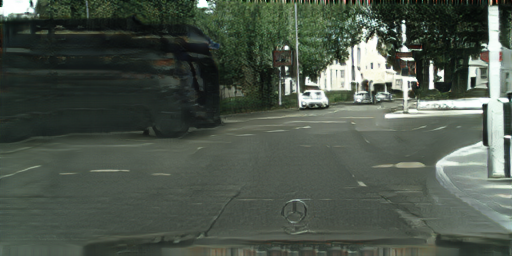}
\end{minipage}
\subcaption{Textureless synthesis of rare classes (Bus)}
\end{subfigure}
\caption{Common Challenges in SIS. The shown images were generated from CUT\cite{park2020cut} and CycleGAN\cite{zhu2017unpaired} on Cityscapes and ADE20K. The cycle consistency penalty in CycleGAN leads to a faulty semantic correspondence in Cityscapes (d) and an unrealistic color generation in ADE20K (a). CUT suffers from either a textureless synthesis (c) or an unrealistic texture synthesis (b). (e) is a common problem in the supervisd and unsupervised settings.}
\label{fig:problems}
\end{figure*}

The most common problems that occur in unsupervised semantic image synthesis can be summarized in the following list and visualized in Figure \ref{fig:problems}:
\begin{itemize}
    \item Class Mixing: class A appears instead of class B
    \item Unrealistic color synthesis
    \item Noisy texture synthesis
    \item Textureless objects(problems identified and solved by SPADE\cite{park2019semantic} in the supervised setting)
    \item Loss of fine details in the object or unrealistic appearance: this problem also occurs in the supervised setting and is related to the low-frequency nature of convolutional layers.
    \item Textureless synthesis for rare classes in the dataset 
\end{itemize}

The first 2 problems occur more often in the unsupervised setting, because the discriminator doesn't have a direct feedback that allows for class recognition. The rest is common in both supervised and unsupervised settings. We assume that the greater the number of classes is in a dataset, the more these problems appear because the classes become harder to distinguish when they have more similar color-codes. Moreover, the number of rare classes increases in a large and diverse dataset. This will be later shown to hold true in the Experiments (section \ref{sec:7}). 
\vspace{-0.3cm}
\section{Proposed Framework}
\label{sec:5}
In this section, we propose a novel framework USIS for unpaired semantic image synthesis, which builds upon the fragmented benefits of the cycle-consistency approach and the relationship preservation approach. We first introduce the proposed unsupervised paradigm. Then, we explain how the self-supervised segmentation loss helps preserve the semantic alignment and enhance the ability of the generator to match the appearance of real data. Finally, we analyze how the design of the discriminator influences the unsupervised learning.

\subsection{SIS Unsupervised Paradigm}
\label{sec:51}
The proposed framework consists of a UNet segmentor which is trained cooperatively with the generator by the means of a self-supervised segmentation loss; and a whole image wavelet-based discriminator which is trained adversarially to capture the color and texture distribution of all semantic classes.

Our framework consists of three parts: (1) a wavelet SPADE Generator $\mathcal{G}$, (2) a wavelet Discriminator $\mathcal{D}$ and (3) a Unet segmentation network $\mathcal{S}$. The generator generates an RGB image $y$ from the semantic map $m$ (one-hot encoded), the discriminator makes a decision whether the generated image is real or fake while the segmentation network tries to segment the generated image back to the mask $m$. We note that $\mathcal{S}$ only observes the generated images unlike the discriminator which sees real and generated images. Thus $\Dis$ competes with $\Gen$ to encourage the generation of photorealistic images (class appearance matching) while $\Seg$ and $\Gen$ cooperate to achieve semantic alignment in the form of a class-balanced self-supervised segmentation loss with the input mask. Thus the combined losses of our framework are:

\begin{equation}
    \label{eq:losses}
    \begin{aligned}
    \mathcal{L}_{Gen} &= \lambda_{seg}\mathcal{L}_{seg}(\textbf{m}, \Seg(\Gen(\textbf{m}))) + \mathcal{L}_{adv_{G}}(\Dis(\Gen(\textbf{m})))\\
    \mathcal{L}_{UNet} &= \mathcal{L}_{seg}(\textbf{m}, \Seg(\Gen(\textbf{m})))\\
    \mathcal{L}_{Dis} &= \mathcal{L}_{adv_{D}}(\Dis(\textbf{x}), \Dis(\Gen(\textbf{m})))
    \end{aligned}
\end{equation}
where $\mathcal{L}_{seg}$ is a class-balanced\cite{schonfeld2021you} self-supervised segmentation loss and is expressed as:
\begin{equation}
    \begin{aligned}
    - \mathbb{E}_{\textbf{m} } \left[ \sum_{c=1}^{C}\alpha_{c}\sum_{i,j}^{H \times W}\textbf{m}_{c,i,j} \log(\mathcal{S}(\mathcal{G}(\textbf{m}))_{c,i,j}) \right]    \end{aligned}
\end{equation}
The class-balancing weights $\alpha_{c}$ are proportional to the inverse of the per-pixel class-frequency.
\begin{ceqn}
    \begin{align}
    \alpha_{c} = \frac{H \times W}{ \sum_{i,j}^{H \times W} \mathbb{E}_{\textbf{m}} \left[ \mathbbm{1}[\textbf{m}_{c,i,j}] \right] } 
    \end{align}
\end{ceqn}
The class balancing makes sure the smaller-classes have a strong contribution in comparison to bigger classes in the segmentation categorical cross entropy loss.

We adopt the non-saturating version\cite{Goodfellow2014GenerativeAN} of the GAN logistic loss $\mathcal{L}_{adv}$, where the discriminator tries to maximize the probability of classifying the images $\textbf{x}$ as real and the generated images $\Gen(\textbf{m})$ as fake; and the generator tries to maximize the probability that the discriminator classifies $\Gen(\textbf{m})$ as real.
\begin{equation}
    \begin{aligned}
    \mathcal{L}_{adv_{D}} &= -\mathbb{E}_{\textbf{x}}\left[ \log(\Dis(\textbf{x})) \right]  -\mathbb{E}_{\textbf{m}}\left[ \log(1-\Dis(\Gen(\textbf{m}))) \right]\\
    \mathcal{L}_{adv_{G}} &= -\mathbb{E}_{\textbf{m}}\left[ \log(\Dis(\Gen(\textbf{m}))) \right]
    \end{aligned}
\end{equation}
We adopt the same regularization scheme as in StyleGANv2\cite{Karras2020AnalyzingAI} to stabilize the training. An $\mathcal{R}1$ regularization\cite{Drucker1992ImprovingGP, Mescheder2018WhichTM, Ross2018ImprovingTA} is applied to the discriminator every 16 minibatches. We design the discriminator to output a single score for the whole image, which stands in contrast to other conditional GANs\cite{wang2018high, zhu2017unpaired, fu2019geometry, benaim2017one, park2020cut, park2019semantic} that use patch discriminators. 
\begin{figure}
\centering
\captionsetup[subfigure]{indention=2pt}
\begin{subfigure}{\linewidth}
\centering
\includegraphics[width=\linewidth]{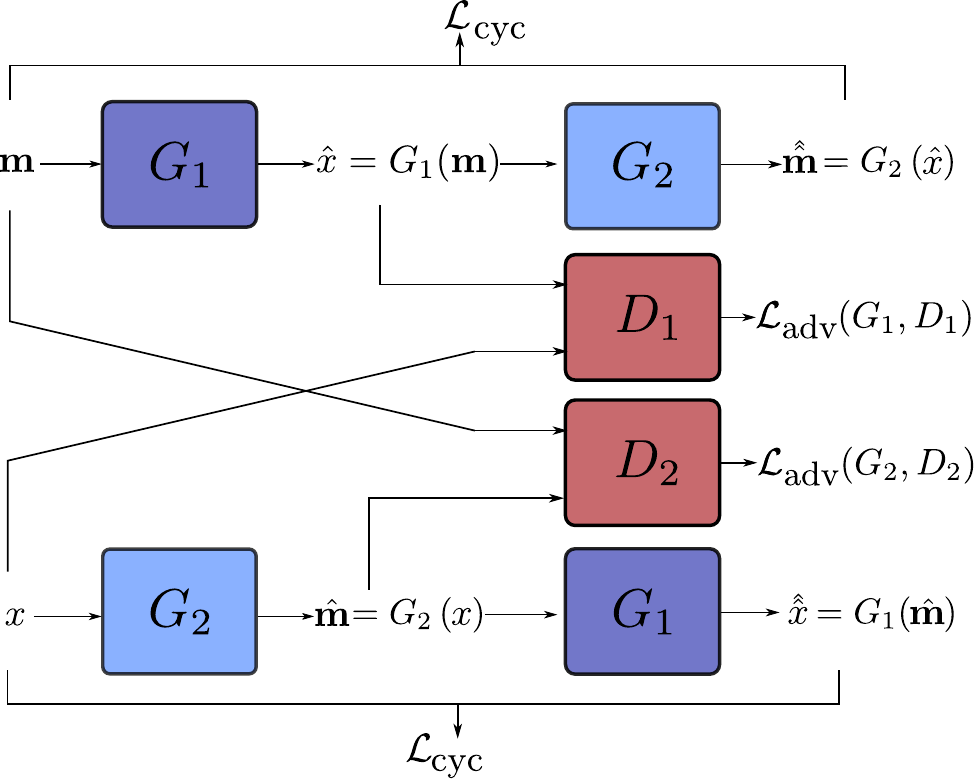}
\caption{Cycle losses in Cycle-GAN\cite{zhu2017toward}}
\label{fig:paradigm:a}
\end{subfigure}
\vspace{3mm}
\begin{subfigure}{\linewidth}
\centering
\includegraphics[width=\linewidth]{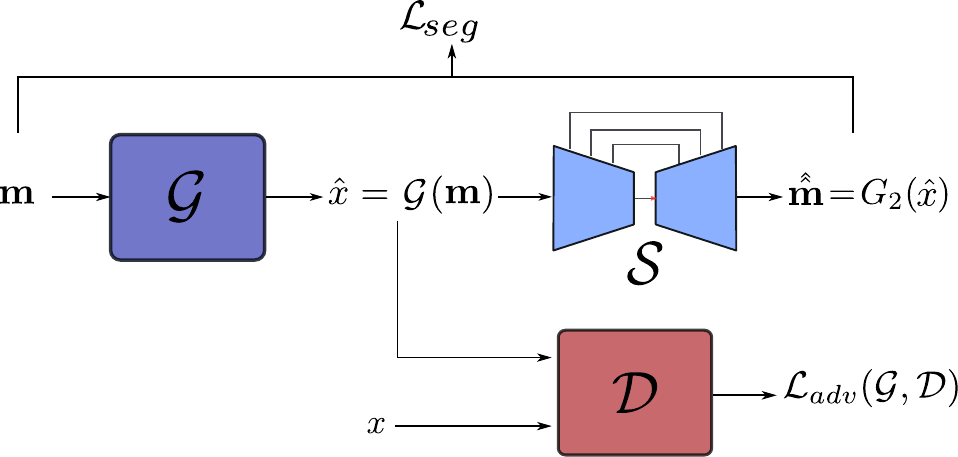}
\caption{Self-supervised segmentation loss}
\label{fig:paradigm:b}
\end{subfigure}
\caption{Cycle losses and self-supervised segmentation loss in USIS. A generated image should appear realistic to the discriminator $\mathcal{D}$ while generating classes that could be distinguished by $\mathcal{S}$.}
\label{fig:paradigm}
\vspace{-0.4cm}
\end{figure}

 The proposed paradigm has two main motivations. First, the self-supervised segmentation loss heavily punishes the inseparability between regions belonging to different semantic labels and prevents the class-mixing problem. In the beginning of the training, it is easier for the generator to produce realistic images by matching the appearance of big classes while ignoring small ones. The self-supervised loss pushes the generator to synthesize small classes and it achieves a better semantic alignment, to counteract the tendency of the generator to satisfy the GAN objective by finding a trivial solution (like matching one or two big classes to make the image appear realistic). It stands out that the self-supervised segmentation loss is different than the cycle losses in CycleGAN\cite{zhu2017unpaired}, MUNIT\cite{huang2018multimodal},  and DRIT\cite{lee2018diverse}. We do not seek to generate segmentation maps or do the inverse mapping. There is neither a discriminator for semantic maps nor a reverse cycle (Real image $\longrightarrow$ Segmentation map $\longrightarrow$ Real image). 

The second motivation is that the self-supervised segmentation loss can also be seen as a relationship preservation constraint. In the previous works on unpaired GANs, the relationship was either defined between different images of one domain\cite{amodio2019travelgan, benaim2017one} , transformations of the same image\cite{fu2019geometry} or patches of the same image\cite{park2020cut}. Instead of contrasting different output patches against each other like CUT\cite{park2020cut}, we contrast different pixels against each other with the help of the self-supervised segmentation loss. Our assumption is that pixels belonging to the same class-label should have similar features in the generator while pixels belonging to different class-labels should have dissimilar features. This is encouraged by classifying the generated pixels back to their labels. 

The preservation of a semantic relationship between different pixels of the image improves the generation capability of the network. In a previous work by Collins\cite{Collins2020EditingIS}, a spherical k-means clustering has been conducted on the deep features of unconditional generative models like Progressive GANs and StyleGAN and has revealed that the feature clusters of a good generative model spatially span semantic objects. The proposed unsupervised paradigm encourages feature clustering explicitly: intra-class feature similarity is enabled by the SPADE layers, while the self-supervised loss enforces the inter-class feature separability inside the generator, to ensure a higher generation quality and diversity. 
\vspace{-0.5cm} 

\subsection{Discriminator Design}
\label{sec:52}
The discriminator is an essential part of the framework because it is responsible for capturing the data statistics. Most importantly, it prevents the generator from learning trivial mappings (like identity mapping) that minimize the self-supervised segmentation loss; and it is the part responsible for discovering the appearance and texture of different classes. 

An important design feature in the discriminator is its visual receptive field. Previous unsupervised models were mostly dependent on patch discriminators, which classify overlapping patches of size $N \times N$ pixels in the original image. The motivation for this design choice was to model the image as a Markov random field assuming that pixels separated by more than a patch size are independent. Patch GANs would thus capture high frequency content in the image, like its texture. However, we argue that the PatchGAN paradigm is not optimal for the purpose of unsupervised SIS because the discriminator is incapable of sufficiently penalizing individual confined objects with unrealistic texture when it has only a localized view of the image. This is due to its intrinsic design of averaging out the scores of all individual image patches which dilutes its sensitivity to local errors. As a result, patchGANs in the context of SIS tend to match only the color distribution of real images while neglecting the texture distribution.

To counteract the drawbacks of patchGANs, we propose the utilization of whole-image discrimination. The whole discriminator assigns a bigger penalty to images with small unrealistic objects even if the remainder of the image has photorealistic textures. This situation happens particularly in the beginning of the training. However, when the training progresses, the whole-image discrimination keeps providing the generator with a strong feedback signal to keep generating finer and smaller classes (humans, poles, traffic signs..).

Wavelet discrimination has been a valuable tool in supervised conditional image generation\cite{Zhang2019SuperresolutionRA, Wang2020MultilevelWG, Huang2019WaveletDG} and unconditional image synthesis\cite{Gal2021SWAGANAS}. In this work, we illustrate its usefulness in the unsupervised setting. The motivation for a frequency-based discrimination is that class appearance is multimodal and objects belonging to the same class may vary a lot across the dataset making it hard for the discriminator in an unsupervised setting to capture all variations of an object in regard to its scale, style, texture, pose and illumination. The smaller and finer an object is, the harder it is to render it in a photorealistic way without a direct supervision signal. To further enhance the capability of the discriminator network in rendering photorealistic small objects, the whole image discriminator architecture is also extended with the use of wavelet-based representations. More specifically, we incorporate the SWAGAN discriminator architecture, which was previously proposed in \cite{Gal2021SWAGANAS} to enhance the texture of generated images. Notably, we repurpose this architectural design to be used for unsupervised class appearance matching because it is more suited for discriminating high-frequency content while focusing on the whole image, in contrast to patch GANs. By allowing the discriminator to process the Discrete Wavelet Transform (DWT) of the image, the higher frequencies are not entirely lost in the downsampling layers of the discriminator and consequently the smaller classes can now have a bigger contribution in the adversarial loss function.
\begin{figure}
\centering
\includegraphics[width=\linewidth]{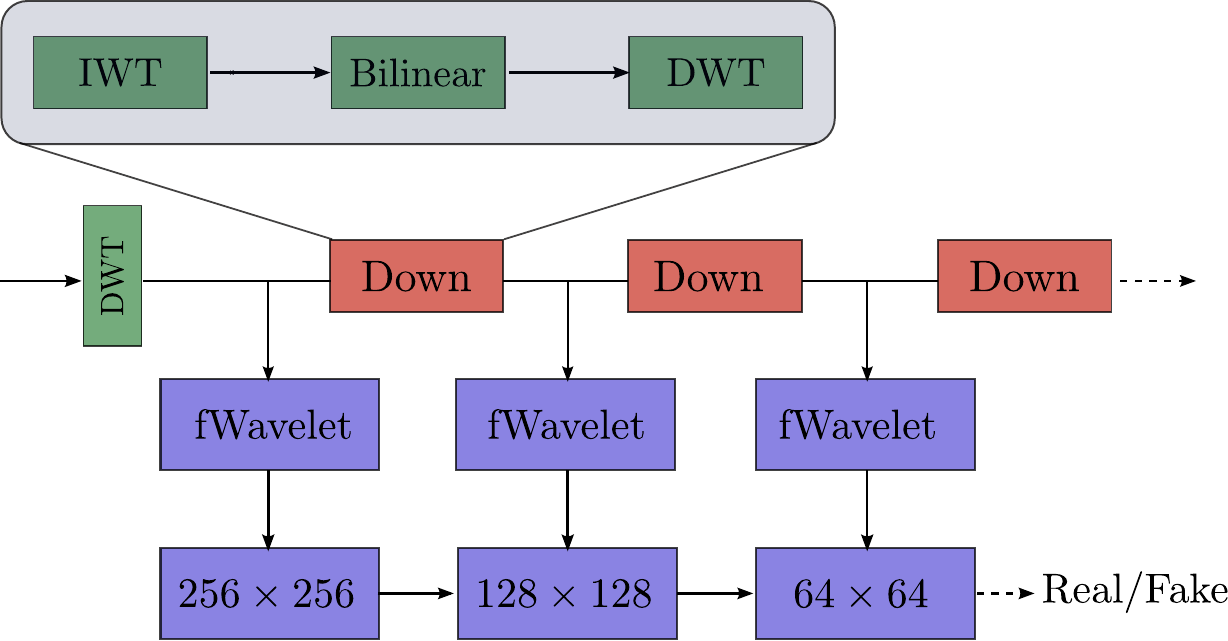}
\caption{Wavelet Discriminator architecture. fWavelet denotes from Wavelet and refers to normal convolution layers that transform the image from wavelet domain to a higher dimensional feature space, in order to be processed further by the network. Bilinear refers to bilinear downsampling of the image. IWT and DWT refer to Inverse Wavelet Transform and Discrete Wavelet Transform.}
\label{fig:disc}
\vspace{-0.4cm}
\end{figure}

The SWAGAN discriminator\cite{Gal2021SWAGANAS} differs from other previously proposed wavelet discriminators\cite{Liu2019AttributeAwareFA, Wang2020MultilevelWG, Huang2019WaveletDG}. It doesn't just use an n-level wavelet decomposition as input to the network, but rather downsamples the image multiple times in the pixel domain, performs a DWT on each resolution then maps it to high-dimensional features using a convolutional block (fWavelets). As can be seen in Figure\ref{fig:disc}, features from different resolutions are mapped together using skip connections. This architecture offers two advantages: first, it performs all n-level wavelet decomposition, one at a time in the network aggregating multiscale features instead of just performing 3 or 4 level wavelet decomposition at the input of the network. Second, this architecture was inspired from StyleGANv2 and designed to replace progressive growing\cite{Karras2018ProgressiveGO} while retaining its advantage: to initially focus on low-resolution features then produce sharper details. While we use the discriminator without modifying its architecture, its motivation and role are different from SWAGAN: in unconditional image generation, objects of different scale can be generated but suffer from a loss of high-frequency details. In our case, the generator in the unsupervised setting cannot even synthesize some of the classes in the dataset without the wavelet discriminator. Typically, these are classes that occupy a small scale.  
\vspace{-0.5cm}

\section{Experiments and Discussion}
\label{sec:7}
We conduct our experiments on 3 datasets: Cityscapes\cite{cordts2016cityscapes}, COCO-stuff\cite{caesar2018coco} and ADE20K\cite{zhou2017scene}. Cityscapes contains street scenes in German cities with pixel-level annotations of 19 classes. It is widely used for vision tasks in autonomous driving and contains 3000 training images and 500 test images. ADE20K and COCO-stuff are more challenging datasets because they offer a high diversity of indoor and outdoor scenes; and they have a lot of semantic classes. COCO-stuff has 182 classes while ADE20K has 150 classes. These 3 datasets are the standard benchmark in the supervised image synthesis task. In contrast, in all of the previous works on unpaired GANs, the semantic image synthesis experiments were only performed on Cityscapes. In what follows, we start by providing more details about the training setup and evaluation metrics. Then, we show the results of our ablation study on Cityscapes to illustrate the role of different parts of the proposed model. Next, we discuss the performance of USIS against the state-of-the-art unpaired models and some of the supervised frameworks. Finally, we showcase the performance of our model in a practical use case: we perform the translation between labels extracted from a modern computer game (GTA-V) and images captured in real-time (Cityscapes).   
\vspace{-0.4cm}
\subsection{Training Details and Evaluation Metrics}
We follow BigGAN\cite{brock2018large} and OASIS\cite{schonfeld2021you} and perform our experiments using an exponential moving average of the generator weights with 0.9999 decay. The image resolution is $256 \times 256$ for COCO and ADE20K, and we use two settings for Cityscapes: $256 \times 256$ and $256 \times 512$. We use a batchsize of 8 on one Titan-RTX GPU for Cityscapes and a batchsize of 32 for ADE20K and COCO on 4 Titan-RTX GPUS. The optimizer in all our experiments is ADAM\cite{Kingma2015AdamAM} with momentums $\beta_{1}=0$, $\beta_{2}=0.999$ and a constant learning rate of 0.0001. In Eq \ref{eq:losses}, the segmentation loss coefficient $\lambda_{seg}$ is set to 1.0.

The standard evaluation metrics for this task are utilized to measure both the quality and diversity of generated images. Specifically, we show the Frechet Inception Distance or FID\cite{heuselttur2017}, to assess both quality and diversity. We also follow SPADE\cite{park2019semantic}: we run pretrained semantic segmentation models on the generated images and report the mean Intersection-over-Union (mIoU) to evaluate the semantic alignment. We employ DRN-D-105\cite{yu2017dilated} (pretrained on multiple scales) for Cityscapes, DeepLabV2\cite{chen2014semantic} for COCO-stuff and UperNet101\cite{xiao2018unified} for ADE20K. The reported mIoU is not only a measure of the semantic alignment but also a measure of the quality of the generated images, because even if the image is aligned with the mask but some objects have an unrealistic or an out-of-distribution texture, a pretrained segmentation network will attribute the wrong class to the object. This is mainly due to the bias of the segmentation networks towards the texture or pixel-statistics of the input image.  
\vspace{-0.4cm}
\begin{table*}[t]

  \centering
    \resizebox{\textwidth}{!}{\begin{tabular}{l @{\hskip 4mm} ccccccc @{\hskip 4mm} cc @{\hskip 4mm} cc}
    \toprule
    \multirow{2}{*}{\textbf{Method}} & \multicolumn{7}{c}{\textbf{Implementation Details}}&\multicolumn{2}{c}{\textbf{ 256 $\times$ 256}} & \multicolumn{2}{c}{\textbf{ 512$\times$256}} \\ 
    \cmidrule(r){2-8} \cmidrule(r){9-10} \cmidrule(r){11-12}
    &\multirow{2}{*}{\textbf{SS}} & \multicolumn{3}{c}{\textbf{Discriminator}} & \multicolumn{3}{c}{\textbf{Generator}} & \multirow{2}{*}{\textbf{FID}} & \multirow{2}{*}{\textbf{mIoU}} & \multirow{2}{*}{\textbf{FID}} & \multirow{2}{*}{\textbf{mIoU}} \\
    \cmidrule(r){3-5}\cmidrule(r){6-8}
    && \textbf{Patch} & \textbf{Whole} & \textbf{Wavelet} & \textbf{CNN} & \textbf{OASIS} & \textbf{Wavelet} \\
    \midrule
    CUT\cite{park2020cut} & & & \cmark & & \cmark & & & 56.4 & 24.7 & 57.3 & 29.8\\
    Config 0  & & & \cmark & &&\cmark & & \multicolumn{4}{c}{No Convergence} \\
    Config A  & \cmark & \cmark & & & &\cmark && 110.39  & 24.76 & 128.67 & 31.41 \\
    Config B  & \cmark & &\cmark & & &\cmark && 47.09 & 32.05 & 55.57 & 35.17 \\
    Config C  & \cmark & & &\cmark & &\cmark && \textbf{45.18} & 37.48 & 52.19 & \textbf{42.8} \\
    Config D  & \cmark & & &\cmark && &\cmark& 45.62 & \textbf{37.74} & \textbf{50.52} & 40.27\\
    \bottomrule
    \end{tabular}}
  \caption{\textbf{Ablations.} Quantitative results of different experiments with different settings to highlight the importance of our contributions. Wavelet generator is an OASIS generator with an IWT in the last layer.}
  \label{table:ablation}
  \vspace{-2mm}

\end{table*}
\begin{table*}[t]
    \centering
    \renewcommand\arraystretch{1.3}
    \begin{adjustbox}{width=1\textwidth}
    \begin{tabular}{l|lllllllllllllllllll|l}
    \hline
                {\small Discriminator} & \rotatebox{90}{Road} & \rotatebox{90}{SW} & \rotatebox{90}{Build} & \rotatebox{90}{Wall} & \rotatebox{90}{Fence} & \rotatebox{90}{Pole} & \rotatebox{90}{TL} & \rotatebox{90}{TS} & \rotatebox{90}{Veg} & \rotatebox{90}{Terrain} & \rotatebox{90}{Sky}  & \rotatebox{90}{Person} & \rotatebox{90}{Rider} & \rotatebox{90}{Car}  & \rotatebox{90}{Truck} & \rotatebox{90}{Bus}  & \rotatebox{90}{Train} & \rotatebox{90}{MC} & \rotatebox{90}{Bike} & mIoU \\ \hline
                
\hline
 
Patch   & 84.9 & 53.1 & 59.0 & 0.2 & 3.5 & 36.3 & 15.2 & 35.5 & 56.0 & 33.5 & 77.4 & 44.5 & 24.4 & 18.7 & 0.5 & 0.0 & 0.0 & 15.9 & 38.0 & 31.41 \\
Whole   & 93.2 & 58.6 & 65.1 & 13.1 & 12.4 & 31.3 & 3.3 & 22.3 & 65.3 & 49.1 & 76.2 & 35.5 & 16.1 & 84.2 & 10.3 & 2.2 & 0.0 & 1.4 & 28.6 & 35.17 \\
Wavelet & 92.7 & 56.3 & 80.7 & 20.6 & 27.8 & 33.9 & 21.3 & 28.9 & 79.6 & 57.7 & 77.6 & 51.1 & 34.4 & 81.1 & 15.5 & 11.1 & 0.01 & 0.02 & 43.2 & 42.80  \\ 
\hline
OASIS\cite{schonfeld2021you} & 97.1 & 80.0 & 85.8 & 70.0 & 65.3 & 40.8 & 46.1 & 57.4 & 85.6 & 70.2 & 91.6 & 63.6 & 49.9 & 88.8 & 78.4 & 78.4 & 66.4 & 47.9 & 60.7 & 69.7 \\

\hline
\end{tabular}
\end{adjustbox}
\caption{IoU per class on Cityscapes dataset.}
\label{tab:miou}
\end{table*}{}
\subsection{Ablation Study}
{\textbf{Main Ablation}} In Table \ref{table:ablation}, we perform an ablation study on Cityscapes to analyze the effect of the different components on the generation capability of the model. For fair comparison, all models were trained with the same OASIS generator (which consists of the SPADE generator with added 3D noise tensor) and a batchsize of 8. We perform our experiments on 2 resolutions: $256 \times 256$ and $512 \times 256$. The smaller resolution is the standard used in unpaired models while the larger one has been the standard in paired models.

First, we notice that the adversarial training alone is not enough for model convergence. Our finding is in line with other unpaired frameworks, which either use cycle losses or relationship preservation losses for convergence. In contrast, we use the self-supervised segmentation loss (SS) and find that it is essential for convergence in all of the following configurations. The introduction of the whole discriminator in Config B boosts the generation capability beyond the state-of-the-art (CUT\cite{park2020cut}) and its effect is mostly visible in the FID score. Not only does the addition of the whole discriminator produce images with an overall more realistic texture than Config A, but also the images are more diverse because the absence of the averaging effect on the output, previously present in the patch discriminator, has enabled for a starker discrimination. 

To further look into the role of the whole discriminator in the image synthesis, we include CUT\cite{park2020cut} in the ablation study because the same whole StyleGANv2\cite{Karras2020AnalyzingAI} discriminator is used with a contrastive loss. The comparison between CUT\cite{park2020cut} and our model in Config B reveals that the improvement should not attributed to the whole discrimination alone but rather it is the combination of the self-supervised segmentation loss and whole discriminator that leads to a higher generation capability (in fact, according to CUT\cite{park2020cut}, the StyleGANv2 discriminator has had only a marginal effect on the quality of their generated images). Another advantage of the SS-loss can be witnessed by comparing the mIoU scores of CUT\cite{park2020cut} and Config A: even though CUT has a higher FID and produces more realistic images than the configuration with the patch discriminator, the latter has still a higher mIoU scores. Thanks to the SS-loss, the generator is able to synthesize objects that are discernible enough.

We experiment with the wavelet discriminator in Config C results and notice a two-fold improvement in the FID and mIoU. However, in contrast to Config B, the FID improvement is marginal while the mIoU improvement is significant in both resolutions; which translates to a better semantic alignment and a higher image quality. Finally, in Config D, we investigate the effect of adding a wavelet decomposition in the generator as well: the architecture of the OASIS generator is kept unchanged and an Inverse Wavelet Transform (IWT) operation is added after the output of the generator. This slight change means that the generator learns to produces wavelet coefficients in contrast to pixels in the previous configurations and that the high features inside the generator are learned inside the frequency domain. While almost inconsequential on the performance in the low-resolution setting, in the high-resolution setting, the addition leads to a slight boost in the FID and a slight drop in the mIoU scores.

\begin{figure*}
    \begin{subfigure}{0.6\textwidth}
    \begin{tikzpicture}
\begin{axis}[%
width=3.5in,
height=2.5in,
at={(0in,0in)},
scale only axis,
xticklabel style = {font=\large},
xmin=0,
xmax=200000,
xlabel style={font=\color{white!15!black}},
xlabel={\large Iteration},
every outer y axis line/.append style={black},
every y tick label/.append style={font=\large \color{black}},
every y tick/.append style={black},
every minor tick,
yticklabel style={%
	/pgf/number format/.cd,
	fixed,
	fixed zerofill,
	precision=3,
},
xtick={0,25000,50000,75000,100000,125000,150000,175000,200000,250000},
xticklabels = {{},{25},{50},{75},{100},{125},{150},{175},{},{}},
ymin=0,
ymax=360,
ytick={0, 40, 80, 120, 160, 200, 240, 280, 320, 360},
yticklabels = {{},{40},{80},{120},{160},{200},{240},{280},{320},{}},
ylabel style={font=\color{black}},
ylabel={\large \textcolor{black}{FID}},
axis background/.style={fill=white},
axis x line*=bottom,
axis y line*=left,
xmajorgrids,
ymajorgrids,
legend style={legend cell align=left, align=left, draw=none}
]

\addplot [color=blue, line width=1.3pt, solid]
table[row sep=crcr]{
2500.0 290.15726754547404 \\
5000.0 311.6595563544605 \\
7500.0 302.27463049973767 \\
10000.0 257.3508930552245 \\
12500.0 267.3391546128326 \\
15000.0 259.2980754705016 \\
17500.0 242.7526966081575 \\
20000.0 240.2481400686246 \\
22500.0 219.25864225409305 \\
25000.0 197.55185866531014 \\
27500.0 197.0269006004691 \\
30000.0 177.1990680729285 \\
32500.0 168.01455073154557 \\
35000.0 179.22916522956598 \\
37500.0 168.26050118932415 \\
40000.0 178.10374077487484 \\
42500.0 173.4936158602914 \\
45000.0 158.03757032295445 \\
47500.0 156.98154150367117 \\
50000.0 156.21739441705375 \\
52500.0 143.6057491167763 \\
55000.0 177.90336586822605 \\
57500.0 161.93798319165444 \\
60000.0 165.45180505034625 \\
62500.0 177.07943011609302 \\
65000.0 156.95082555945757 \\
67500.0 158.71088191310443 \\
70000.0 165.39101596514405 \\
72500.0 176.3674402580932 \\
75000.0 191.0464980312112 \\
77500.0 162.48627046953214 \\
80000.0 190.4486752035054 \\
82500.0 160.08207661426007 \\
85000.0 181.18091022841435 \\
87500.0 175.20050846839666 \\
90000.0 172.67547833611883 \\
92500.0 173.8805039613299 \\
95000.0 174.3624225676949 \\
97500.0 153.8914640767038 \\
100000.0 170.99549377000665 \\
102500.0 160.4623056119134 \\
105000.0 167.49195574752054 \\
107500.0 159.56193576984504 \\
110000.0 177.44653071489626 \\
112500.0 195.07209695144243 \\
115000.0 198.60710268438623 \\
117500.0 183.79961146205414 \\
120000.0 220.01329174467125 \\
122500.0 183.49727976654572 \\
125000.0 177.39557252699706 \\
127500.0 194.54838244245897 \\
130000.0 190.45101853956538 \\
132500.0 173.37367814286804 \\
135000.0 196.95306186921806 \\
137500.0 197.1826457227711 \\
140000.0 161.7538091767858 \\
142500.0 190.07318794649018 \\
145000.0 204.1493370899113 \\
147500.0 189.4648372313088 \\
150000.0 169.79910845892994 \\
152500.0 166.86678016724514 \\
155000.0 176.72319895671757 \\
157500.0 204.41014571963328 \\
160000.0 187.58910949800526 \\
162500.0 185.38442411247092 \\
165000.0 180.63605362112665 \\
167500.0 260.91142020600876 \\
170000.0 204.93833203904373 \\
172500.0 212.1151385594784 \\
175000.0 210.5845938810284 \\
177500.0 223.89068965613453 \\
180000.0 210.2619195366886 \\
182500.0 200.7352244463686 \\
185000.0 198.28488079165447 \\
187500.0 199.70946518784964 \\
190000.0 210.48464534431903 \\
192500.0 190.23069773996906 \\
195000.0 191.15773936361677 \\
197500.0 157.89672706459262 \\
200000.0 175.74652208836577 \\
};
\label{lab:patch}

\addplot [color=green, line width=1.3pt, solid]
table[row sep=crcr]{
2500 287.004 \\
5000 254.581 \\
7500 129.507 \\
10000 114.693 \\
12500 96.900 \\
15000 85.113 \\
17500 82.006 \\
20000 79.249 \\
22500 79.447 \\
25000 77.741 \\
27500 78.358 \\
30000 80.826 \\
32500 76.636 \\
35000 72.084 \\
37500 71.827 \\
40000 73.529 \\
42500 70.442 \\
45000 73.420 \\
47500 73.257 \\
50000 69.704 \\
52500 69.319 \\
55000 71.061 \\
57500 72.657 \\
60000 71.664 \\
62500 64.708 \\
65000 67.696 \\
67500 70.291 \\
70000 66.596 \\
72500 65.203 \\
75000 67.389 \\
77500 70.177 \\
80000 68.210 \\
82500 67.860 \\
85000 65.179 \\
87500 67.061 \\
90000 67.931 \\
92500 67.032 \\
95000 64.353 \\
97500 61.420 \\
100000 62.508 \\
102500 61.377 \\
105000 64.338 \\
107500 62.796 \\
110000 62.357 \\
112500 62.157 \\
115000 65.880 \\
117500 61.969 \\
120000 67.842 \\
122500 61.895 \\
125000 60.905 \\
127500 57.348 \\
130000 62.878 \\
132500 63.007 \\
135000 60.787 \\
137500 62.958 \\
140000 60.471 \\
142500 61.507 \\
145000 60.518 \\
147500 57.640 \\
150000 61.375 \\
152500 57.019 \\
155000 59.607 \\
157500 60.355 \\
160000 58.230 \\
162500 63.032 \\
165000 62.395 \\
167500 55.586 \\
170000 56.342 \\
172500 57.019 \\
175000 62.259 \\
177500 64.855 \\
180000 58.615 \\
182500 63.263 \\
185000 64.077 \\
187500 61.088 \\
190000 62.546 \\
192500 58.348 \\
195000 57.351 \\
197500 62.914 \\
198199 63.914 \\
};
\label{lab:whole}

\addplot [color=red, line width=1.3pt, solid]
table[row sep=crcr]{
2500 236.678 \\
5000 217.202 \\
7500 187.185 \\
10000 127.985 \\
12500 96.463 \\
15000 88.557 \\
17500 72.096 \\
20000 74.265 \\
22500 67.051 \\
25000 65.913 \\
27500 65.014 \\
30000 63.587 \\
32500 62.899 \\
35000 60.907 \\
37500 61.734 \\
40000 58.588 \\
42500 59.624 \\
45000 58.408 \\
47500 60.412 \\
50000 58.139 \\
52500 56.781 \\
55000 60.226 \\
57500 56.780 \\
60000 57.081 \\
62500 57.078 \\
65000 55.544 \\
67500 56.588 \\
70000 57.851 \\
72500 56.226 \\
75000 57.516 \\
77500 58.312 \\
80000 57.265 \\
82500 57.643 \\
85000 55.368 \\
87500 56.880 \\
90000 57.821 \\
92500 55.780 \\
95000 52.729 \\
97500 57.193 \\
100000 54.689 \\
102500 54.477 \\
105000 55.498 \\
107500 56.331 \\
110000 54.354 \\
112500 53.604 \\
115000 52.879 \\
117500 54.155 \\
120000 53.827 \\
122500 52.772 \\
125000 53.326 \\
127500 54.754 \\
130000 52.191 \\
132500 54.567 \\
135000 56.562 \\
137500 53.184 \\
140000 54.838 \\
142500 54.370 \\
145000 56.525 \\
147500 55.637 \\
150000 53.033 \\
152500 53.368 \\
155000 54.155 \\
157500 54.128 \\
160000 54.600 \\
162500 53.070 \\
165000 52.335 \\
167500 54.703 \\
170000 53.408 \\
172500 53.474 \\
175000 55.638 \\
177500 54.620 \\
180000 57.788 \\
182500 57.325 \\
185000 54.581 \\
187500 56.130 \\
190000 54.886 \\
192500 53.291 \\
195000 54.215 \\
197500 56.826 \\
198199 57.650 \\
};
\label{lab:wavelet}

\coordinate (legendPos) at (axis description cs:0.975,0.855);
\end{axis}

\node[anchor=east,text width=2.4cm,align=left,fill=white,draw=black!80,line width=0.1mm,font=\fontsize{12}{0}\selectfont] at (legendPos) {\ref{lab:patch} \large Patch \\ \vspace{1mm} \ref{lab:whole} \large Whole \\ \vspace{1mm} \ref{lab:wavelet} \large Wavelet};
\end{tikzpicture}%
    \end{subfigure}
    \qquad
    \begin{subfigure}{0.4\textwidth}
    \centering
    \begin{subfigure}{0.9\textwidth}
    \includegraphics[width=\textwidth, height=0.4\textwidth]{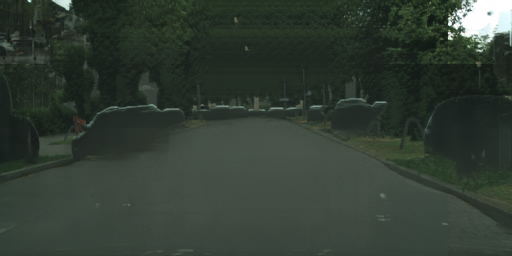}
    \caption{Patch}
    \end{subfigure}
    \vfill
    \begin{subfigure}{0.9\textwidth}
    \includegraphics[width=\textwidth, height= 0.4\textwidth]{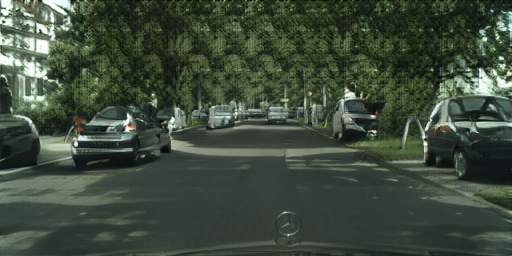}
    \caption{Whole}
    \end{subfigure}   
    \vfill
    \begin{subfigure}{0.9\textwidth}
    \includegraphics[width=\textwidth, height= 0.4\textwidth]{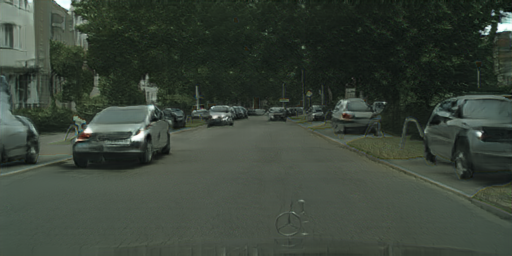}
    \caption{Wavelet}
    \end{subfigure}
    
    \end{subfigure}
    \caption{Quantitative and qualitative ablation results on Cityscapes dataset.}
    \label{fig:plots}
\end{figure*}

{\textbf{Ablation study on the discriminator type.}} In Figure \ref{fig:plots}, we showcase the isolated effect of the discriminator type on the model training and the image quality. Whole discrimination adds more stability to the training compared to the patch discriminator. On the other hand, wavelet discrimination is slightly slower than whole discrimination but reaches a lower FID and exhibits the same stability during training. In terms of visual quality, the patch discriminator has the worst performance of the three types. Although it is able to match the colors of the bigger classes in Figure \ref{fig:plots} (road, tree), the cars are barely visible and have the same color as the road. Although the car boundaries are discernible (thanks to the self-supervised loss), the texture is almost non-existent. On the other hand, the whole discriminator can clearly generate cars with sharper details (wheels, car lights) but doesn't generate trees with a realistic texture, due to the loss of higher frequencies in the downsampling layers of the generator. The wavelet discriminator solves this issue and combines the advantages of patch and whole image discriminators to yield a higher quality.The effect is visible in both large (street, building and trees) and small classes (cars).    

\begin{table*}[t]

	 %

	\setlength{\tabcolsep}{0.2em}
	\renewcommand{\arraystretch}{0.95}
	\centering
	
	\begin{tabular}{c|c|cc|cc|cc}
    	\multirow{2}{*}{\normalsize{} Method } & \multirow{2}{*}{\normalsize{} Supervised }& \multicolumn{2}{c|}{\normalsize{} Cityscapes} &  \multicolumn{2}{c|}{\normalsize{} ADE20K} & \multicolumn{2}{c}{\normalsize{} COCO-stuff}  \tabularnewline
    	&  & \normalsize{} FID$\downarrow$  & \normalsize{} mIoU$\uparrow$ &  \normalsize{} FID$\downarrow$  & \normalsize{} mIoU$\uparrow$ &  \normalsize{} FID$\downarrow$  & \normalsize{} mIoU$\uparrow$ \tabularnewline
    	
    	\hline 
    	
    	{\normalsize{} CycleGAN\cite{zhu2017unpaired} } &  \footnotesize{} \xmark  & \normalsize{87.2} & \normalsize{24.5}  &  \normalsize{96.3} & \normalsize{5.4} &  \normalsize{104.7} & \normalsize{2.08} \tabularnewline

    	{\normalsize{} MUNIT\cite{huang2018multimodal}  } &  \footnotesize{} \xmark  & \normalsize{84} & \normalsize{8.2} &   \normalsize{n/a} & \normalsize{n/a}  &  \normalsize{n/a} & \normalsize{n/a} \tabularnewline

    	{\normalsize{} DRIT\cite{lee2018diverse}  } & \footnotesize{} \xmark  & \normalsize{164} & \normalsize{9.5} &  \normalsize{132.2} & \normalsize{0.016} &  \normalsize{135.5} & \normalsize{0.008} \tabularnewline

    	 {\normalsize{} DistanceGAN\cite{benaim2017one}  } &  \footnotesize{}  \xmark  & \normalsize{78} & \normalsize{17.6} &  \normalsize{80} & \normalsize{0.035} &  \normalsize{92.4} & \normalsize{0.014 } \ \tabularnewline

    	{\normalsize{} GCGAN\cite{fu2019geometry} } &  \footnotesize{} \xmark  & \normalsize{80} & \normalsize{8.4} &   \normalsize{92} & \normalsize{0.07} &  \normalsize{99.8} & \normalsize{0.019} \tabularnewline

    	\normalsize{} CUT\cite{park2020cut} &  \footnotesize{} \xmark  & 
    	\normalsize{57.3} & \normalsize{29.8} &  \normalsize{79.1.1}  &\normalsize{6.9} &  \normalsize{85.6} & \normalsize{2.21}  \tabularnewline
    	\arrayrulecolor{lightgray}
    	
    	 \hline
    
    	\normalsize{} U-SIS	&  \footnotesize{} \xmark  & \normalsize{\textbf{52.2}} & \normalsize{\textbf{42.8}} & \normalsize{\textbf{32.4}} & \normalsize{\textbf{16.87}} & \normalsize{\textbf{27.3}} & \normalsize{\textbf{13.16}} \tabularnewline
    	
    	\hline
    	
		{\normalsize{} CRN\cite{chen2017photographic} } & \footnotesize{} \cmark  & \normalsize{104.7} & \normalsize{52.4} & \normalsize{73.3} & \normalsize{22.4}  &    \normalsize{70.4} & \normalsize{23.7} \tabularnewline

		{\normalsize{} SIMS\cite{qi2018semi}  } &  \footnotesize{} \cmark  & \normalsize{49.7} & \normalsize{47.2}  & \normalsize{n/a} & \normalsize{n/a} &     \normalsize{n/a} & \normalsize{n/a} \tabularnewline

	{\normalsize{} Pix2pixHD\cite{wang2018high}  } &  \footnotesize{} \cmark  &  \normalsize{95.0} & \normalsize{58.3} & \normalsize{81.8} & \normalsize{20.3} &    \normalsize{111.5} & \normalsize{14.6} \tabularnewline
		
		{\normalsize{} SPADE\cite{park2019semantic} } &  \footnotesize{} \cmark  & \normalsize{71.8} & \normalsize{62.3} & \normalsize{33.9} & \normalsize{38.5} &   \normalsize{22.6} & \normalsize{37.4}  \tabularnewline
		
		{\normalsize{} CC-FPSE\cite{liu2019learning} } &  \footnotesize{} \cmark  & \normalsize{54.3} & \normalsize{65.5} & \normalsize{31.7} & \normalsize{43.7} &   \normalsize{19.2} & \normalsize{41.6}  \tabularnewline
		
		{\normalsize{} OASIS\cite{schonfeld2021you} } &  \footnotesize{} \cmark  & \normalsize{\textcolor{darkred}{47.7}} & \normalsize{\textcolor{darkred}{69.3}} & \normalsize{\textcolor{darkred}{28.3}} & \normalsize{\textcolor{darkred}{48.8}} &   \normalsize{\textcolor{darkred}{17.0}} & \normalsize{\textcolor{darkred}{44.1}}  \tabularnewline
		
		\arrayrulecolor{lightgray}

		\end{tabular}
		\caption{Comparison against SOTA methods on 3 datasets. Bold denotes the best performance for unsupervised models while red denotes the state-of-the-art for supervised models and is considered as the upper bound. All models have been evaluated using the officially published codes. GCGAN with the vertical flip consistency was evaluated on Cityscapes because of the rectangular resolution of the images in the dataset while GCGAN with rotation consistency was evaluated on ADE20K and Cocostuff.}
	\label{table:comp_with_sota}
\end{table*}

In order to reliably measure the quality of image synthesis, and identify the strengths and weaknesses of the proposed framework, we showcase the IoU for each class in the generated Cityscapes images. The IoU was obtained by applying a pre-trained DRN-D-105\cite{yu2017dilated} on Cityscapes, as previously discussed. For the purpose of this study, Cityscapes presents itself as the most suitable dataset because it has a limited amount of classes and the experiments were run on the higher resolution ($256 \times 512$), making it easier to visualize the effects of the different discriminator architectures. We present our results in Table \ref{tab:miou}. The improvement brought by the whole discriminator is visible in large and medium classes. Road, sidewalks, buildings and terrain for instance are generated with a better texture. The largest improvement was exhibited by the class "car", whose IoU jumped from 18.7 to 84.2. The downside of the whole discriminator is that smaller classes (person, Rider, Traffic Lights and Bikes) have suffered from a slight drop in the IoU scores. However, the overall improvement overshadows the negative side effects and lead to an overall better mIoU and FID, which correlates to a better visual perception. The wavelet discriminator keeps the same improvements brought by the whole discriminator and even corrects its shortcomings, by generating the small classes. Compared to the patch discriminator, the wavelet discriminator can generate classes that were almost absent in the images generated by patch discrimination (like Wall, Truck and Bus). Finally, we have included OASIS\cite{schonfeld2021you} in the comparison as an upper bound to our model, in order to identify improvement opportunities. Most notably, the small classes still need improvement but it's the rare classes that have suffered the most in unsupervised training. Classes like Truck, Bus, Train and Motorcycle are not present in a lot of training images so they are assigned a low weight in the discriminator loss. In contrast, the classes that have seen the most improvement are almost present in every scene in Cityscapes; such as Buildings, Cars and even the Person class, which counts as a small class but is frequently seen in the dataset. This pattern is also present in other datasets like ADE20K and Cocostuff, which will be discussed in the next subsection.
\begin{figure*}
    \captionsetup[subfloat]{position=top, labelformat=empty, skip=0pt}
    \centering
    \begin{minipage}{0.7\textwidth}
    \subfloat[Label]{\includegraphics[width=  0.195\textwidth, height=  0.155\textwidth]{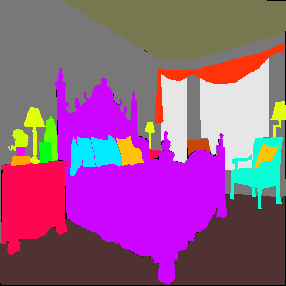}} \hfill
    \subfloat[CycleGAN\cite{zhu2017unpaired}]{\includegraphics[width=  0.195\textwidth, height=  0.155\textwidth]{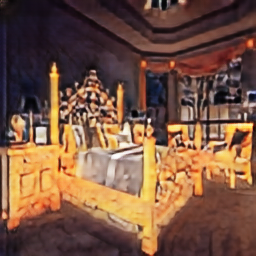}} \hfill
    \subfloat[CUT\cite{park2020cut}]{\includegraphics[width=  0.195\textwidth, height=  0.155\textwidth]{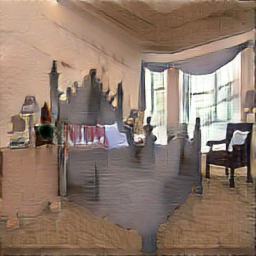}} \hfill
    \subfloat[USIS]{\includegraphics[width=  0.195\textwidth, height=  0.155\textwidth]{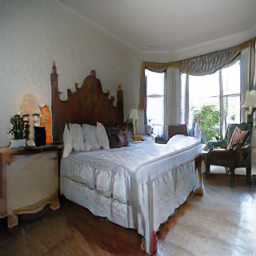}} \hfill
    \subfloat[Groundtruth]{\includegraphics[width=  0.195\textwidth, height=  0.155\textwidth]{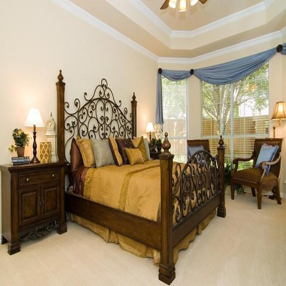}}
    \vfill
    
    \subfloat[]{\includegraphics[width=  0.195\textwidth, height=  0.155\textwidth]{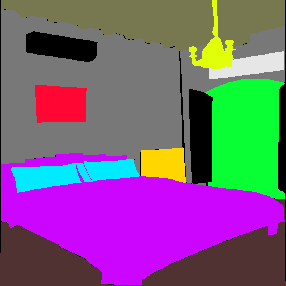}} \hfill
    \subfloat[]{\includegraphics[width=  0.195\textwidth, height=  0.155\textwidth]{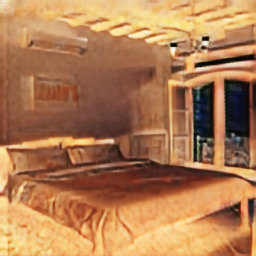}} \hfill
    \subfloat[]{\includegraphics[width=  0.195\textwidth, height=  0.155\textwidth]{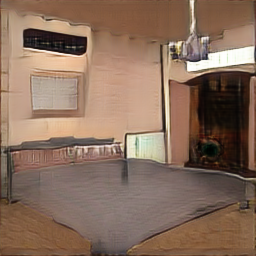}} \hfill
    \subfloat[]{\includegraphics[width=  0.195\textwidth, height=  0.155\textwidth]{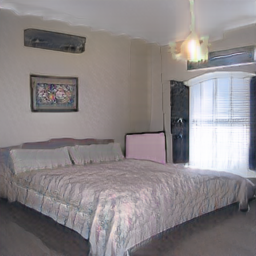}} \hfill
    \subfloat[]{\includegraphics[width=  0.195\textwidth, height=  0.155\textwidth]{figures/ade20k/gt_2.jpg}}
    \vfill
    
    \subfloat[]{\includegraphics[width=  0.195\textwidth, height=  0.155\textwidth]{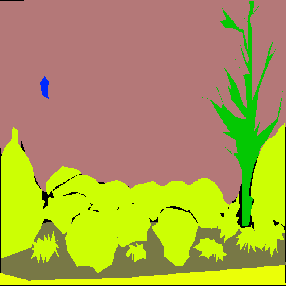}} \hfill
    \subfloat[]{\includegraphics[width=  0.195\textwidth, height=  0.155\textwidth]{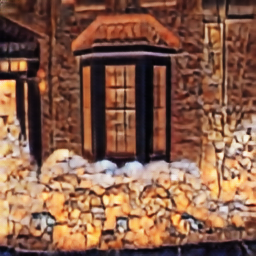}} \hfill
    \subfloat[]{\includegraphics[width=  0.195\textwidth, height=  0.155\textwidth]{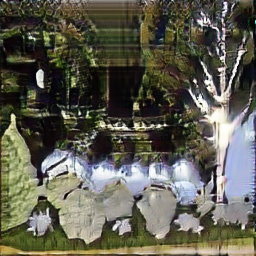}} \hfill
    \subfloat[]{\includegraphics[width=  0.195\textwidth, height=  0.155\textwidth]{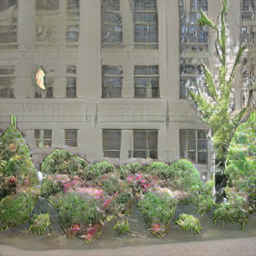}} \hfill
    \subfloat[]{\includegraphics[width=  0.195\textwidth, height=  0.155\textwidth]{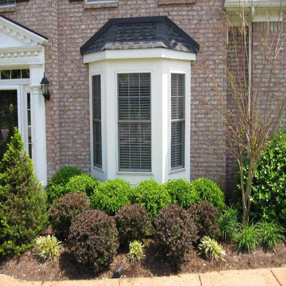}}
    \vfill
    
    \subfloat[]{\includegraphics[width=  0.195\textwidth, height=  0.155\textwidth]{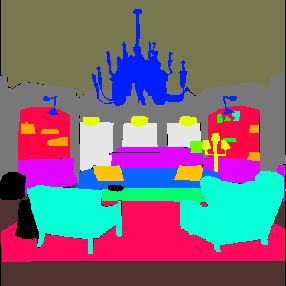}} \hfill
    \subfloat[]{\includegraphics[width=  0.195\textwidth, height=  0.155\textwidth]{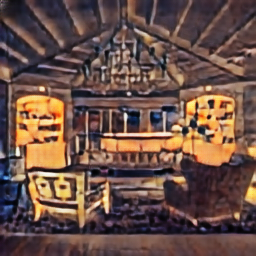}} \hfill
    \subfloat[]{\includegraphics[width=  0.195\textwidth, height=  0.155\textwidth]{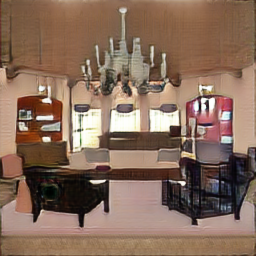}} \hfill
    \subfloat[]{\includegraphics[width=  0.195\textwidth, height=  0.155\textwidth]{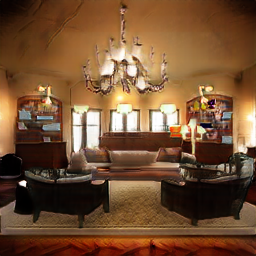}} \hfill
    \subfloat[]{\includegraphics[width=  0.195\textwidth, height=  0.155\textwidth]{figures/ade20k/gt_9.jpg}}
    \vfill
    
    \subfloat[]{\includegraphics[width=  0.195\textwidth, height=  0.155\textwidth]{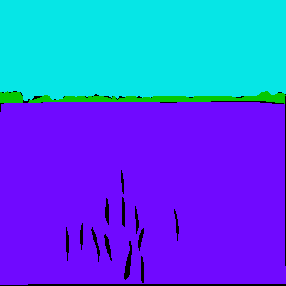}} \hfill
    \subfloat[]{\includegraphics[width=  0.195\textwidth, height=  0.155\textwidth]{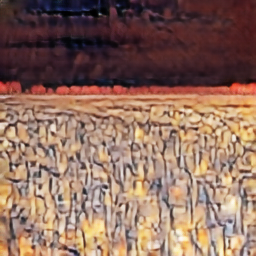}} \hfill
    \subfloat[]{\includegraphics[width=  0.195\textwidth, height=  0.155\textwidth]{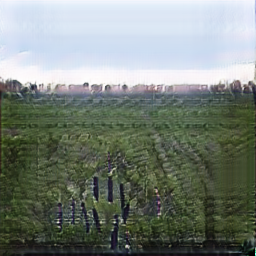}} \hfill
    \subfloat[]{\includegraphics[width=  0.195\textwidth, height=  0.155\textwidth]{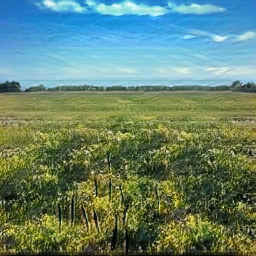}} \hfill
    \subfloat[]{\includegraphics[width=  0.195\textwidth, height=  0.155\textwidth]{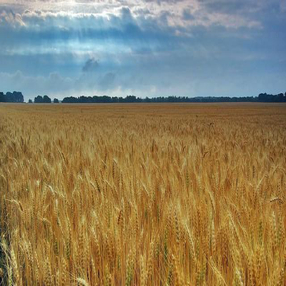}}
    \vfill

    \end{minipage}
    
   \caption{Qualitative Comparison against state-of-the-art on ADE20K Dataset}
   \label{fig:ade20k_paper_results}
\end{figure*}

\subsection{Main Results}
In Table \ref{table:comp_with_sota}, the performance of USIS is compared against the state-of-the-art models in unpaired image-to-image translation on all 3 datasets. The results showcase the effectiveness of the proposed framework in the task of unpaired image synthesis, with respect to both FID and mIoU. We incorporate supervised baselines in the table as an upper bound to the unsupervised model and to illustrate that the proposed USIS is a first step to bridge the existing performance gap. 

First, it is visible that many unpaired frameworks have better FID scores than paired frameworks, especially the baselines that appeared before SPADE\cite{park2019semantic}. FID is influenced by both image quality and diversity; and since many supervised baselines were not trained to learn multimodal generation, their FID score is affected. However, they still exhibit a better mIoU, thanks to the supervised losses which establish clear correspondences between input and output. In this case, the higher mIoU score can be interpreted to correlate with a superior visual quality in more classes. 

Second, we find that CycleGAN and CUT perform consistently good on the 3 datasets relative to the other baselines. However, we observe in Figure \ref{fig:ade20k_paper_results} that images were generated with either unrealistic color (in the case of CycleGAN) or unrealistic texture (CUT). Other baselines can more or less approximate the color and texture distributions of real images but often fail to output semantically meaningful objects. For instance, DistanceGAN has a better FID than CycleGAN but it doesn't generate objects with discernible boundaries. In contrast, CycleGAN produces more visible objects (higher mIoU scores) but with unrealsitic color and texture, due to the restrictions of the cycle losses. CUT consistently performs better than other baselines on the 2 metrics. \par
\begin{figure*}
    \captionsetup[subfloat]{position=top, labelformat=empty, skip=0pt}
    \centering
    \begin{minipage}{0.8\textwidth}
    \subfloat[Label]{\includegraphics[width=  0.33\textwidth, height=  0.155\textwidth]{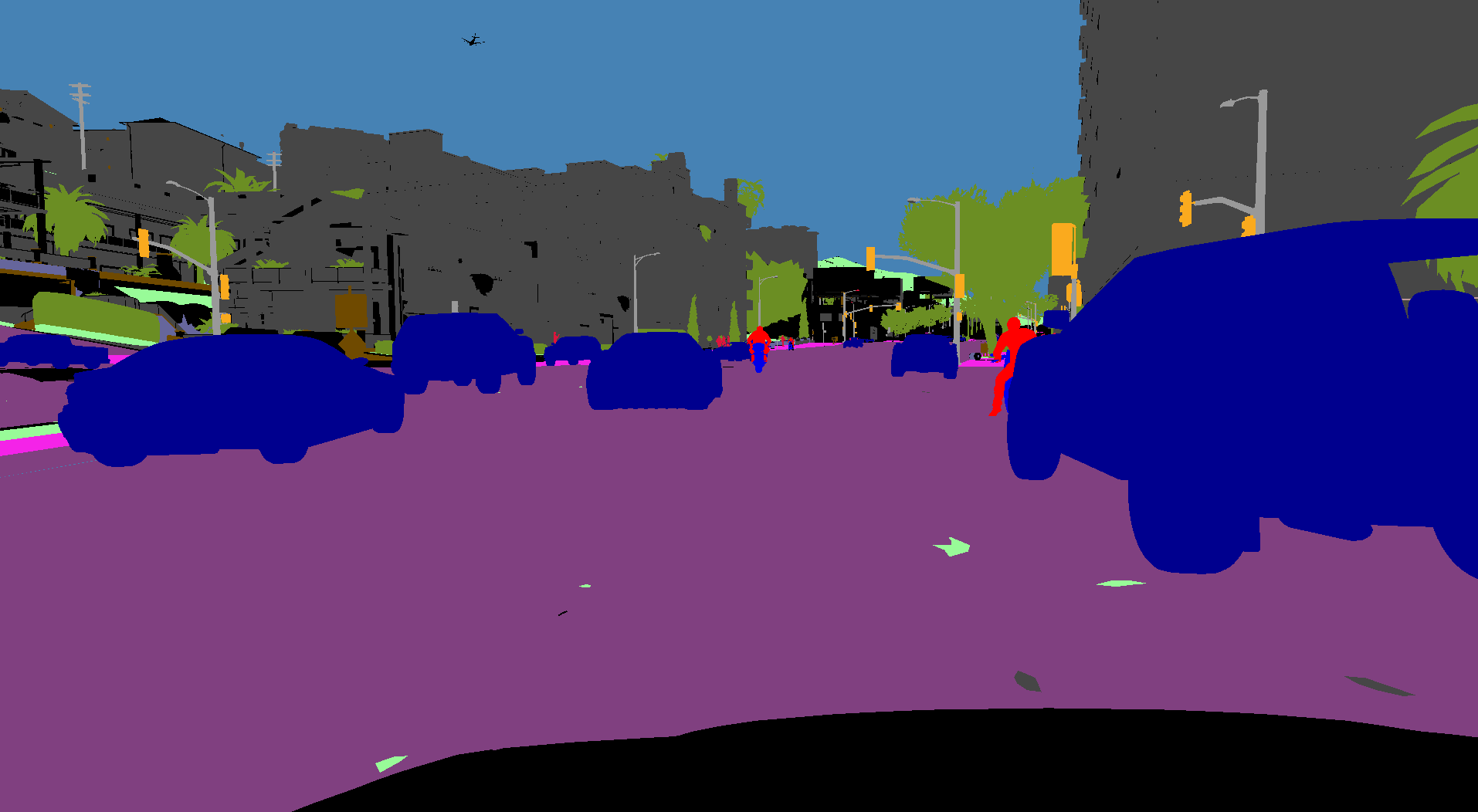}} \hfill
    \subfloat[USIS]{\includegraphics[width=  0.33\textwidth, height=  0.155\textwidth]{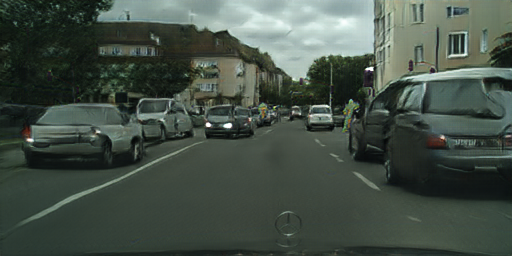}} \hfill
    \subfloat[Groundtruth]{\includegraphics[width=  0.33\textwidth, height=  0.155\textwidth]{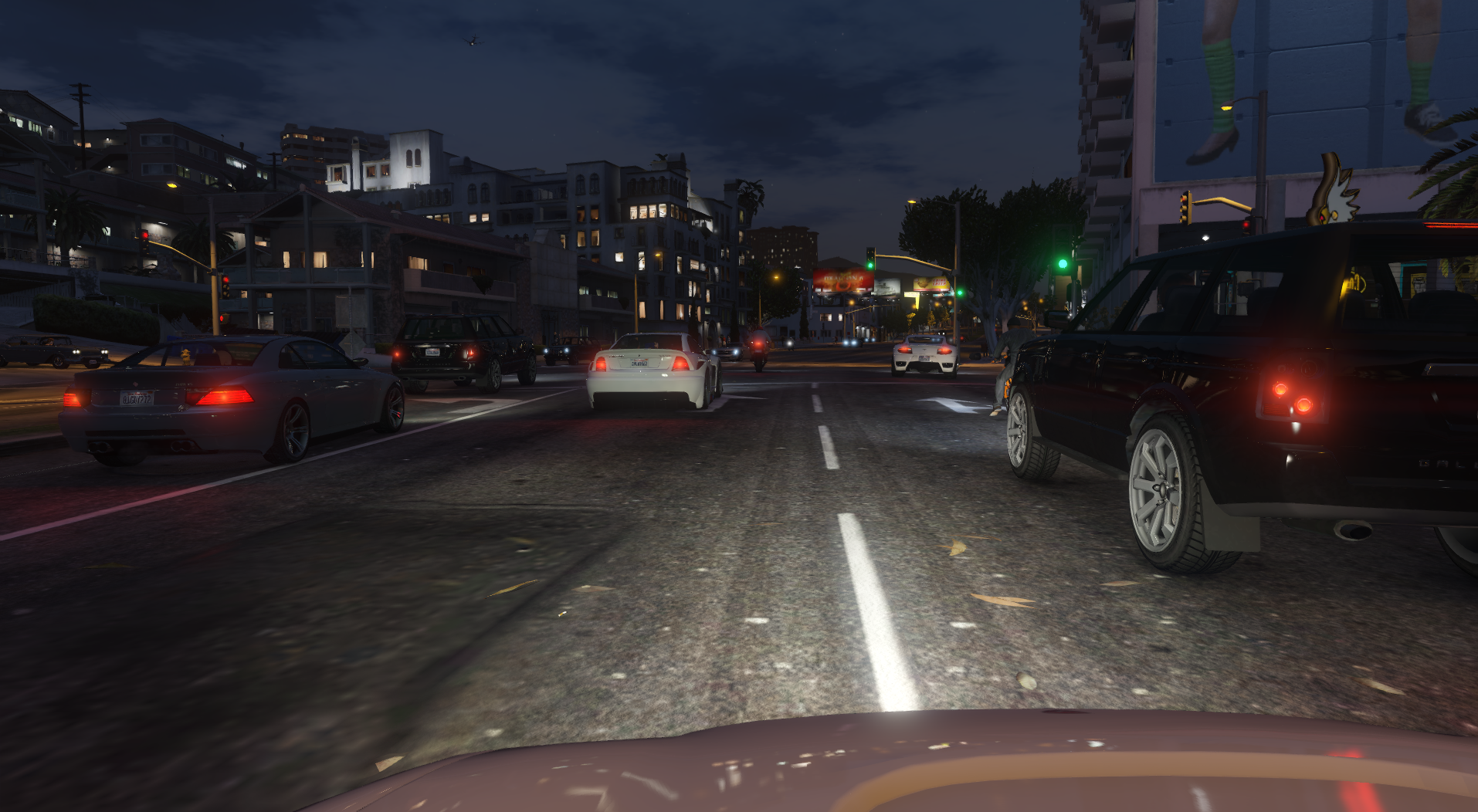}}
    \vspace{-0.2cm}
    
    \subfloat[]{\includegraphics[width=  0.33\textwidth, height=  0.155\textwidth]{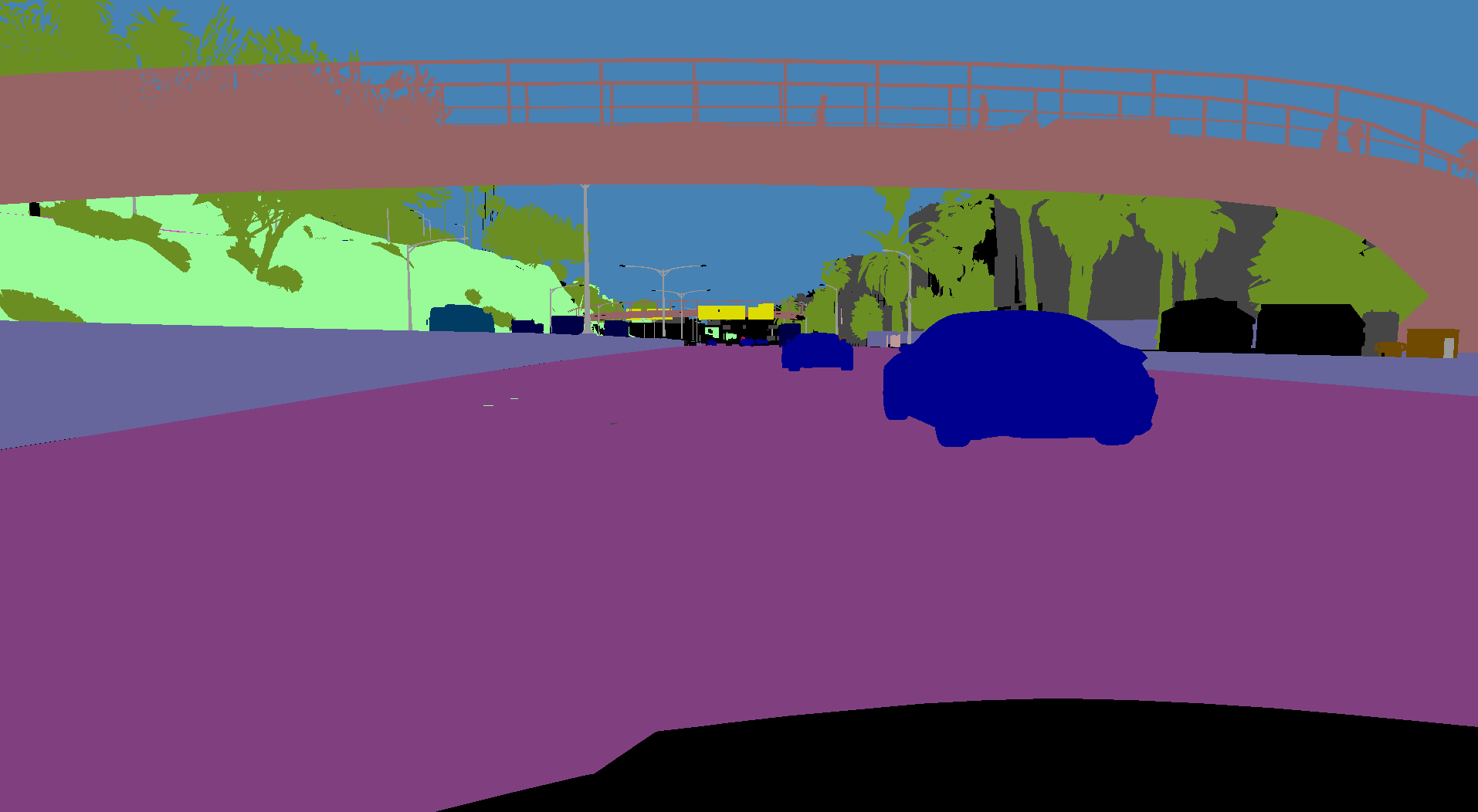}} \hfill
    \subfloat[]{\includegraphics[width=  0.33\textwidth, height=  0.155\textwidth]{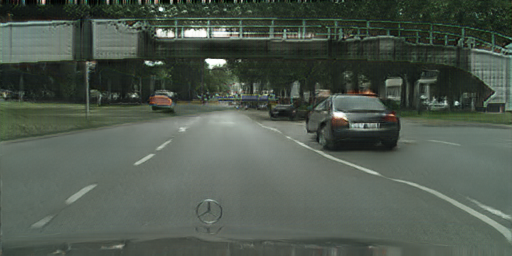}} \hfill
    \subfloat[]{\includegraphics[width=  0.33\textwidth, height=  0.155\textwidth]{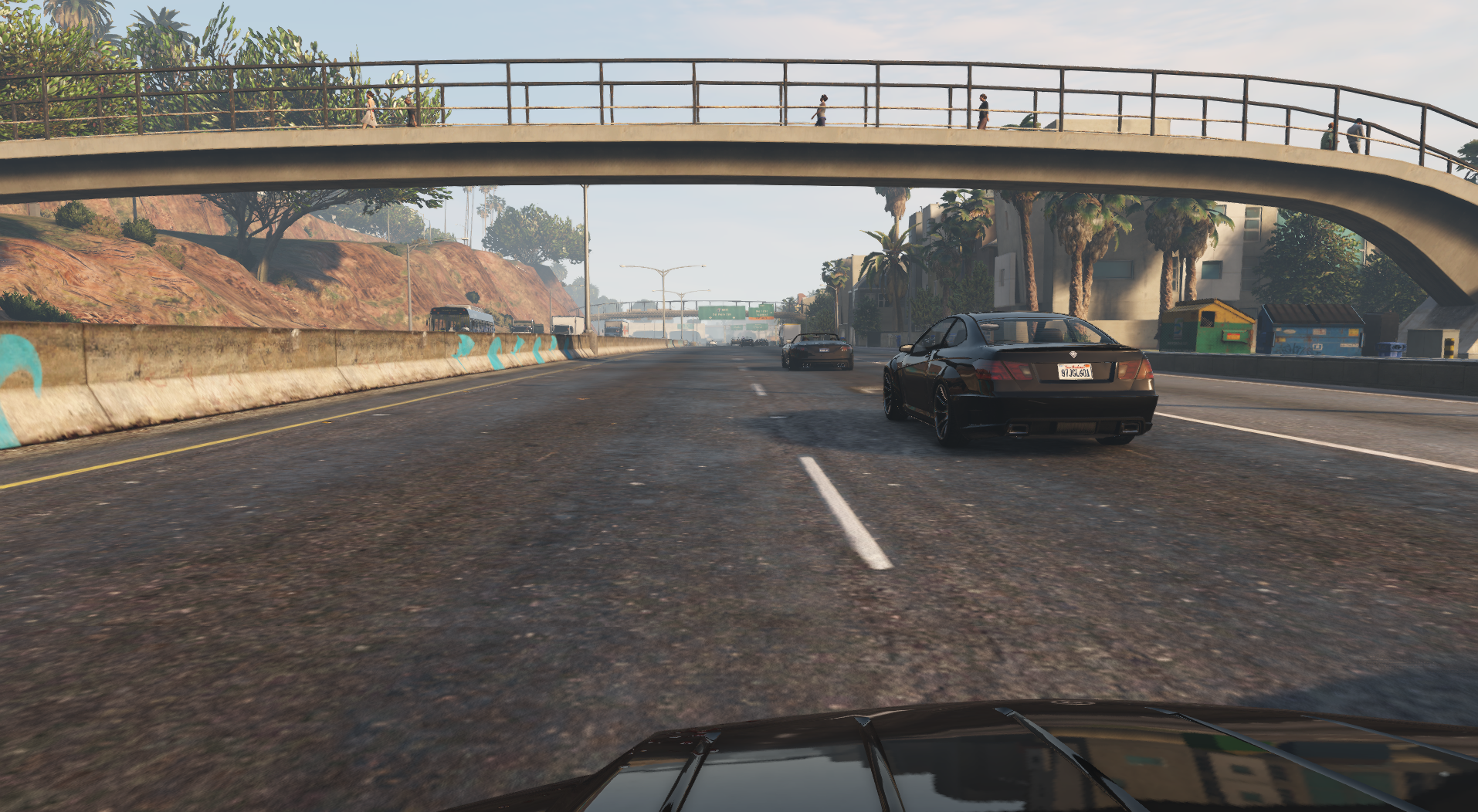}}
    \vspace{-0.2cm}
    
    \subfloat[]{\includegraphics[width=  0.33\textwidth, height=  0.155\textwidth]{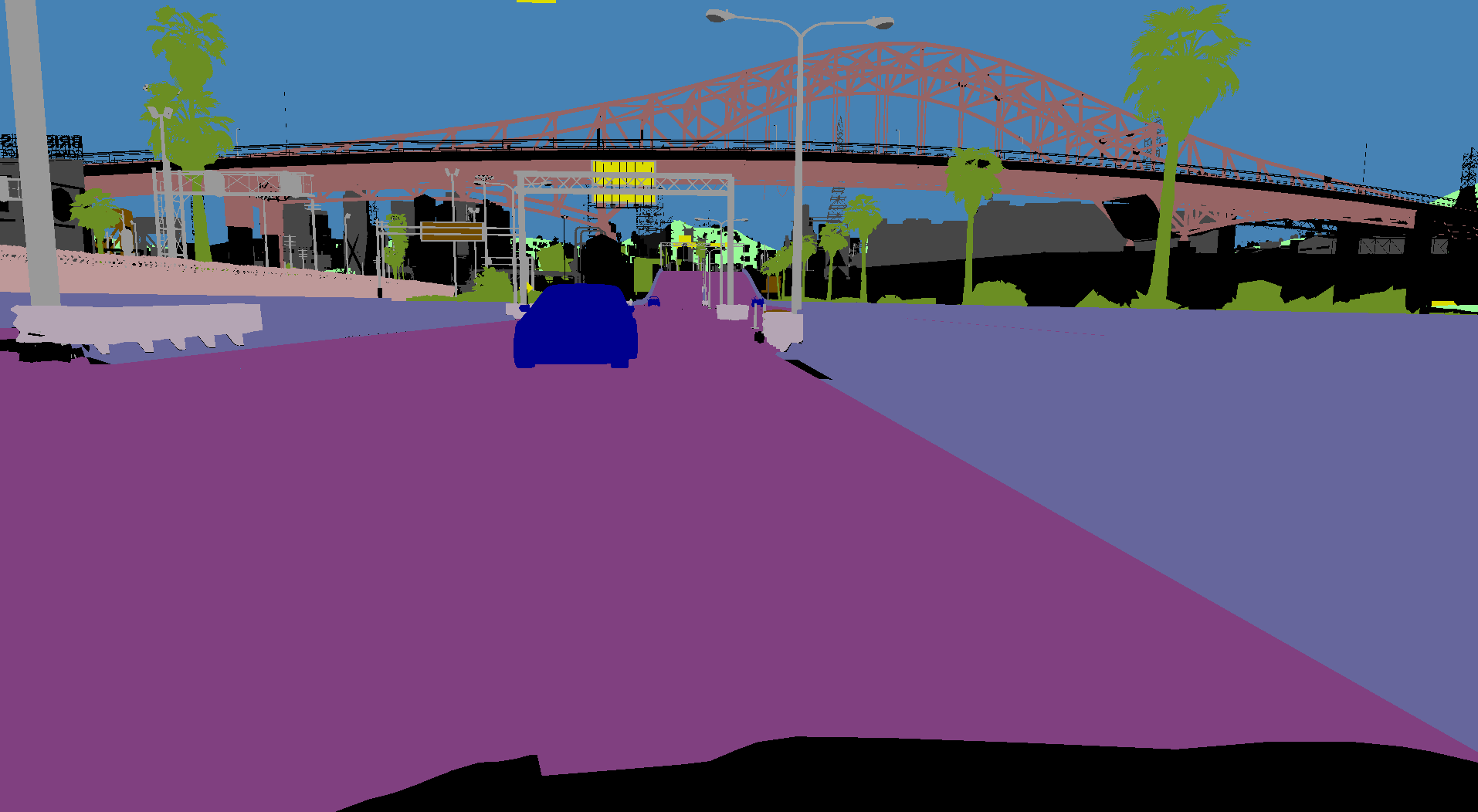}} \hfill
    \subfloat[]{\includegraphics[width=  0.33\textwidth, height=  0.155\textwidth]{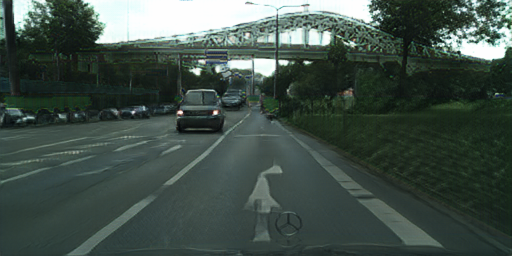}} \hfill
    \subfloat[]{\includegraphics[width=  0.33\textwidth, height=  0.155\textwidth]{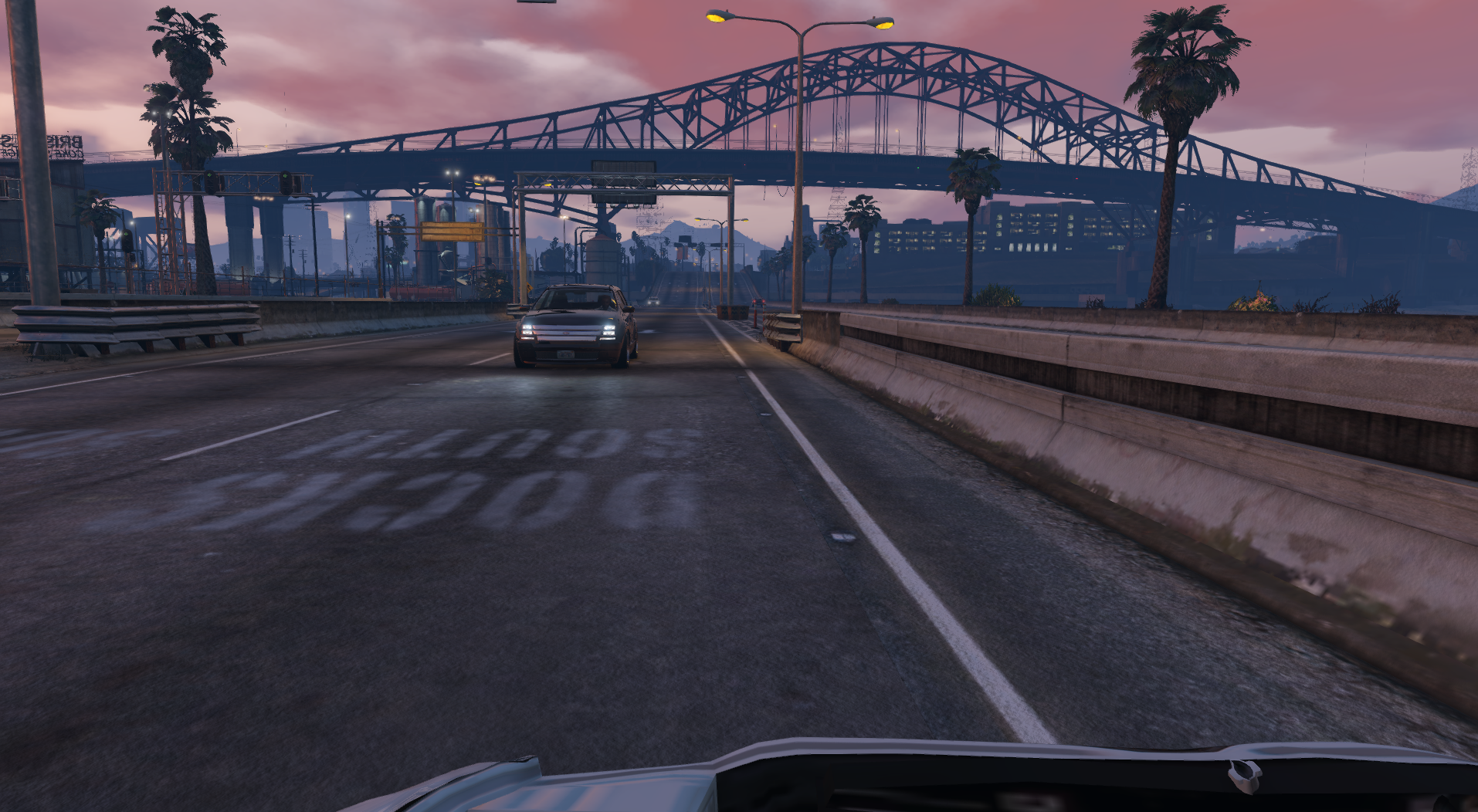}}
    \vspace{-0.2cm}
    
    \subfloat[]{\includegraphics[width=  0.33\textwidth, height=  0.155\textwidth]{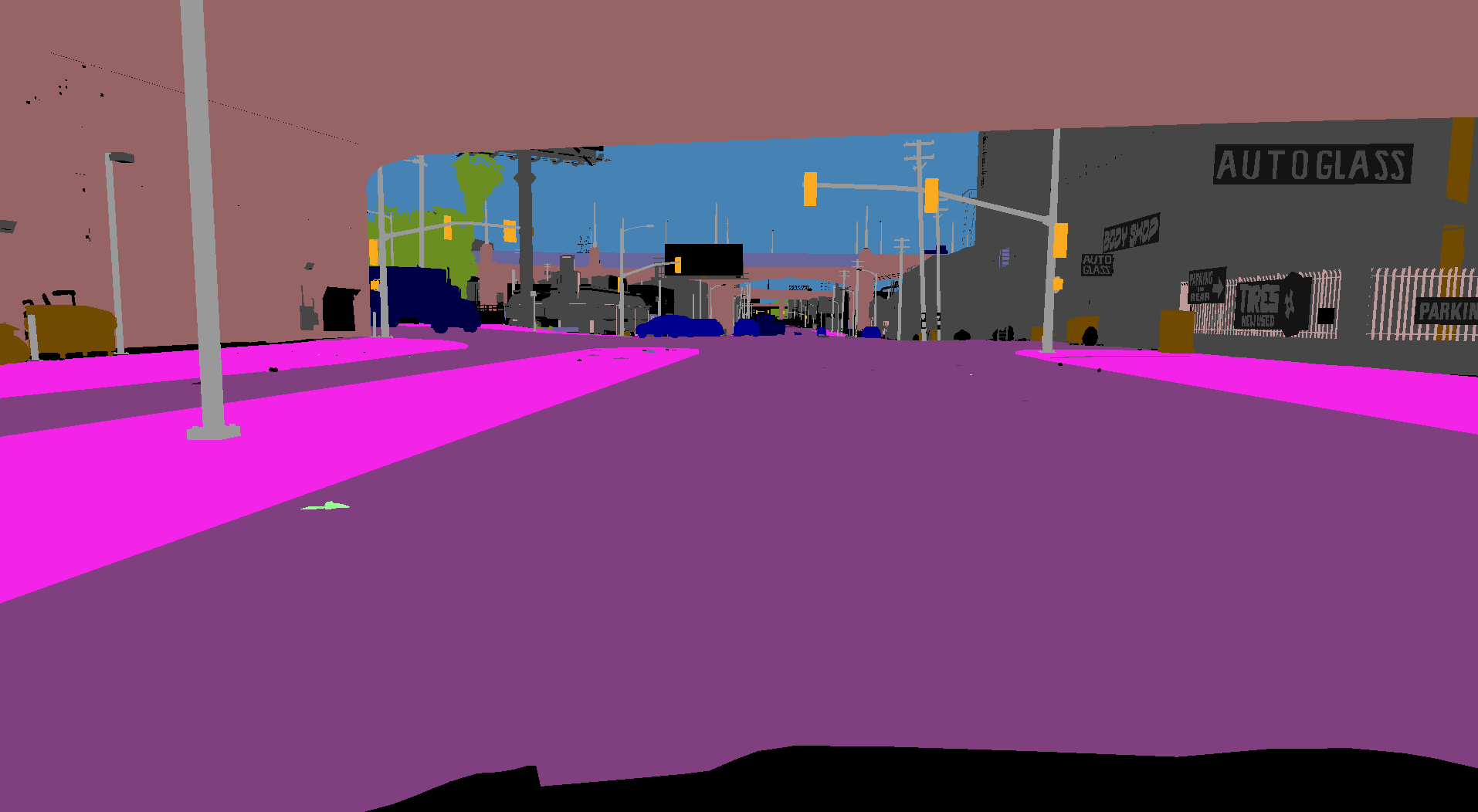}} \hfill
    \subfloat[]{\includegraphics[width=  0.33\textwidth, height=  0.155\textwidth]{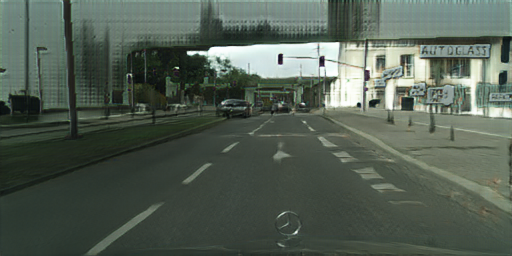}} \hfill
    \subfloat[]{\includegraphics[width=  0.33\textwidth, height=  0.155\textwidth]{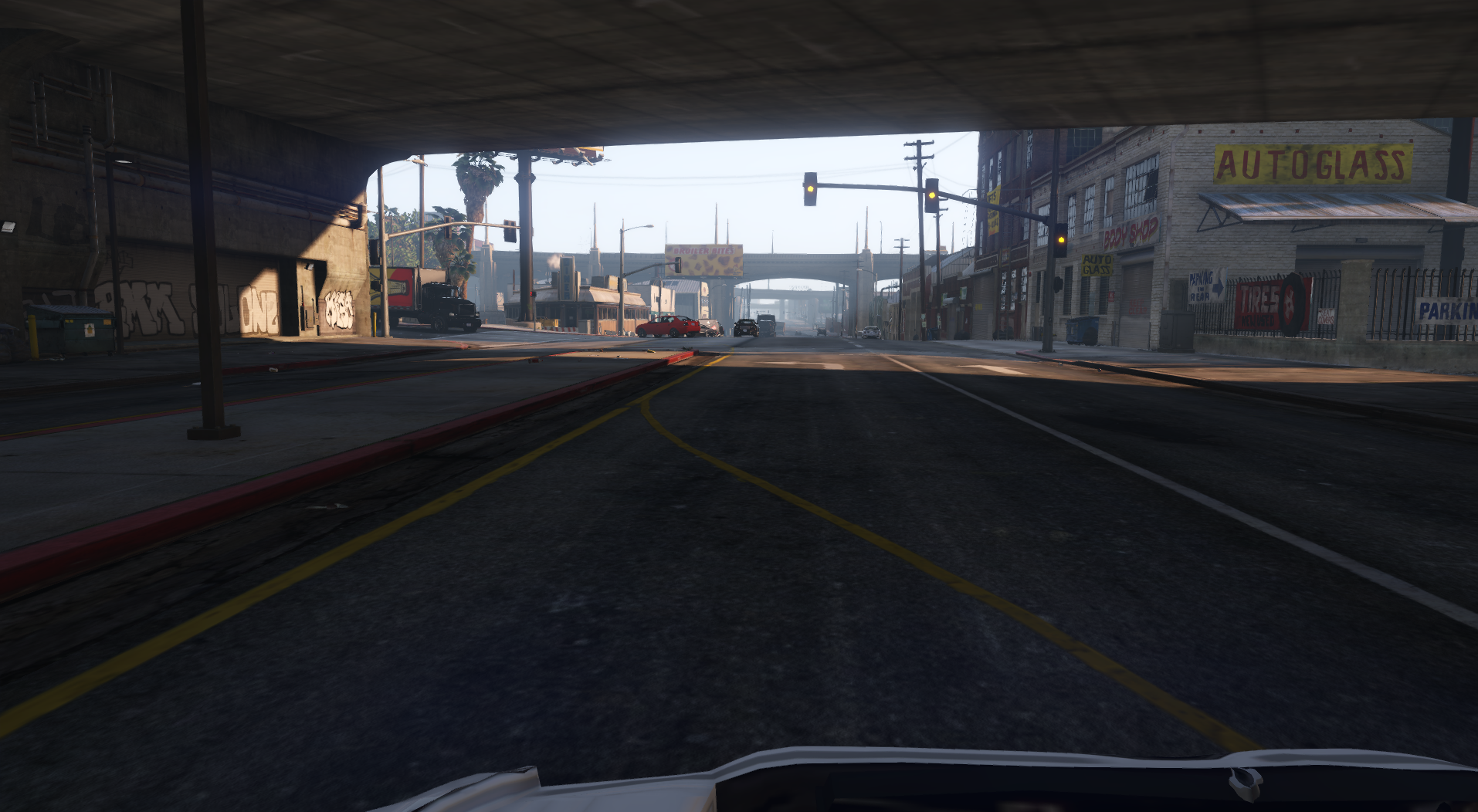}}
    \vspace{-0.2cm}

    \end{minipage}
    
   \caption{Results of USIS when trained on GTA-V labels and Cityscapes images}
   \label{fig:gtav_paper_results}
\end{figure*}

However, we observe a performance gap between Cityscapes on one hand and ADE20K and Cocostuff on the other hand. To our knowledge, there has been no evaluation of unpaired image-to-image translation baselines on ADE20K and Cocostuff, so we used the published codes to train and evaluate the aforementioned baselines on these 2 datasets. We assume that the performance drop on ADE20K and Cocostuff correlates directly to the larger amount of classes ($> 150$) they contain in contrast to Cityscapes (34). This entails 2 consequences: first, the color codes assigned to the classes are much closer in value leading most of the baselines to establish label-to-image correspondences that do not preserve semantics; and second, because of the high diversity in images in both datasets (showing indoor and outdoor scenes), there exists a larger number of "rare" classes that should be evaluated by the discriminator. The low mIoU scores in ADE20K and Cocostuff columns quantifies the suboptimal quality observed visually in Figure \ref{fig:ade20k_paper_results}. The proposed model USIS was able to generate more photorealistic classes in many scenes, both indoor and outdoor. We encourage interested readers to refer to the Appendix \ref{Appendix} for more visual results on the 3 datasets. 

\vspace{-0.5cm}    
\subsection{Application on different datasets}
Finally, we conduct an experiment for a practical use case where the labels and the images come from two different datasets: GTA-V\cite{Richter_2016_ECCV} and Cityscapes. GTA-V is a collection of 25,000 frames taken from a modern computer game and for which dense pixel-level annotations have been generated in an automatic way. In Figure \ref{fig:gtav_paper_results}, we show some synthesized Cityscapes like images from GTA-V labels. The task is more challenging because there exists a domain gap between GTA-V labels and Cityscapes labels: on one hand, some elements are present in GTA-V labels but almost absent in Cityscapes (like bridges, tunnels or some steel structures from construction sites); on the other hand, GTA-V is modeled after USA cities while Cityscapes was acquired in European cities leading to a different scene composition. Nevertheless, we have included some challenging scenes in Figure \ref{fig:gtav_paper_results} in order to test the performance of the proposed model in such a setting. The results show a texture that is still similar to Cityscapes albeit with more semantic misalignment at the borders of challenging objects. The new objects themselves (bridge, tunnel) are rendered in a reasonable way although they have not been observed by the discriminator during the training. We argue that a simple data augmentation for Cityscapes images (which is easy to acquire) would be enough to enhance the performance. No data augmentation for the labels is needed.    
\vspace{-0.5cm}
\section{Conclusion}
\label{sec:8}
\vspace{-0.2cm}
We propose a framework, USIS, for semantic image synthesis in an unpaired setting. It deploys a SPADE generator along with a Unet and an unconditional wavelet-based whole image discriminator. The UNet fosters class separability and content preservation while the discriminator matches the color and texture distribution of real images. The effectiveness of the proposed framework in the semantic image synthesis was shown on 3 challenging datasets: Cityscapes, ADE20K and Cocostuff. An ablation study was performed to analyze the role of the different components of the unsupervised paradigm. Finally, we tested USIS on (GTA-V label)-to-(Cityscapes image) translation to validate its performance in a more challenging setting, as there exists a domain gap between the input labels and the groundtruth images.

USIS is a first step towards bridging the performance gap between paired and unpaired settings. Unsupervised image synthesis paves the way for the semi-supervised setting which is a promising setup for its practical use cases and the fast improvements it can bring to SIS models while reducing the needed amount of labeled data. We also plan to address the problem of rare classes synthesis in both supervised and unsupervised frameworks.
\vspace{-0.3cm}
\begin{acknowledgements}
The research leading to these results is funded by the German Federal Ministry for Economic Affairs and Energy within the project "KI Delta Learning – Development of methods and tools for the efficient expansion and transformation of existing AI  modules  of  autonomous  vehicles  to  new  domains."  The authors would like to thank the consortium for the successful cooperation.
\end{acknowledgements}
\newpage

\bibliographystyle{spmpsci}      
\bibliography{references.bib}   
\newpage
\onecolumn
\begin{appendices}
\section{\texorpdfstring{\centering}{} Additional Results}
\label{Appendix}
\begin{figure*}[h]
    \captionsetup[subfloat]{position=top, labelformat=empty, skip=0pt}
    \centering
    \begin{minipage}{0.7\textwidth}
    
    \subfloat[Label]{\includegraphics[width=  0.195\textwidth, height=  0.155\textwidth]{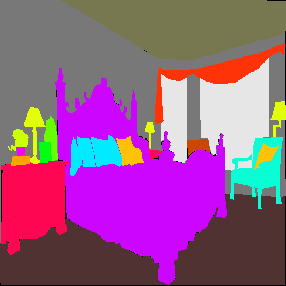}} \hfill
    \subfloat[Sample 1]{\includegraphics[width=  0.195\textwidth, height=  0.155\textwidth]{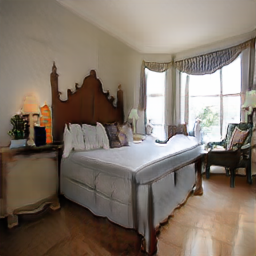}} \hfill
    \subfloat[Sample 2]{\includegraphics[width=  0.195\textwidth, height=  0.155\textwidth]{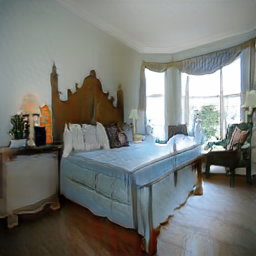}} \hfill
    \subfloat[Sample 3]{\includegraphics[width=  0.195\textwidth, height=  0.155\textwidth]{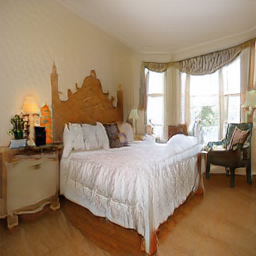}} \hfill
    \subfloat[Sample 4]{\includegraphics[width=  0.195\textwidth, height=  0.155\textwidth]{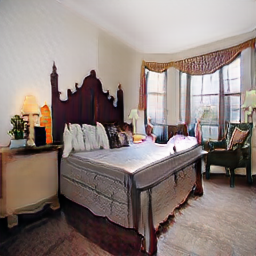}}
    \vfill
    
    \subfloat[]{\includegraphics[width=  0.195\textwidth, height=  0.155\textwidth]{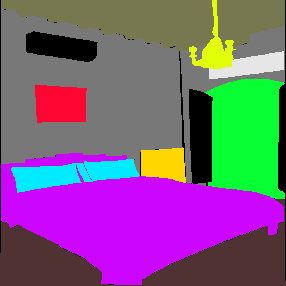}} \hfill
    \subfloat[]{\includegraphics[width=  0.195\textwidth, height=  0.155\textwidth]{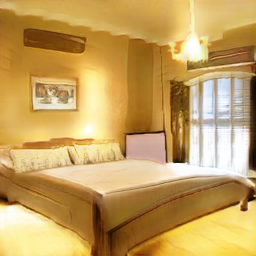}} \hfill
    \subfloat[]{\includegraphics[width=  0.195\textwidth, height=  0.155\textwidth]{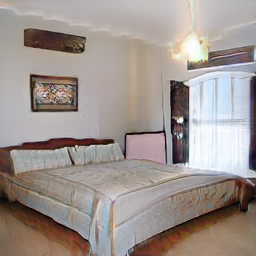}} \hfill
    \subfloat[]{\includegraphics[width=  0.195\textwidth, height=  0.155\textwidth]{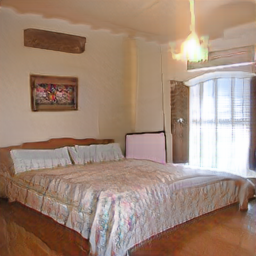}} \hfill
    \subfloat[]{\includegraphics[width=  0.195\textwidth, height=  0.155\textwidth]{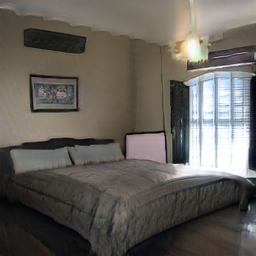}}
    \vfill
    \subfloat[]{\includegraphics[width=  0.195\textwidth, height=  0.155\textwidth]{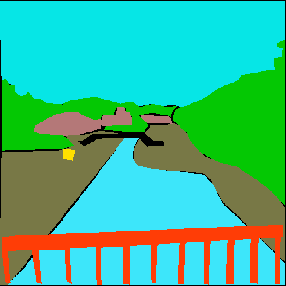}} \hfill
    \subfloat[]{\includegraphics[width=  0.195\textwidth, height=  0.155\textwidth]{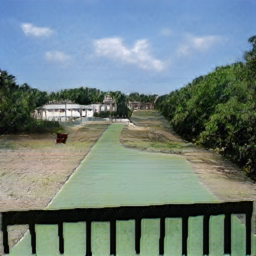}} \hfill
    \subfloat[]{\includegraphics[width=  0.195\textwidth, height=  0.155\textwidth]{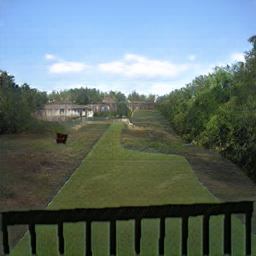}} \hfill
    \subfloat[]{\includegraphics[width=  0.195\textwidth, height=  0.155\textwidth]{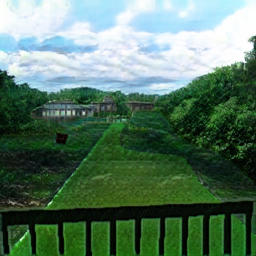}} \hfill
    \subfloat[]{\includegraphics[width=  0.195\textwidth, height=  0.155\textwidth]{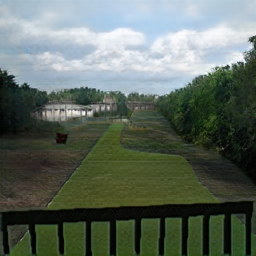}}
    \vfill

    \subfloat[]{\includegraphics[width=  0.195\textwidth, height=  0.155\textwidth]{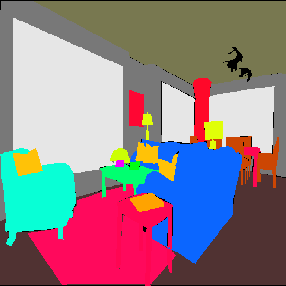}} \hfill
    \subfloat[]{\includegraphics[width=  0.195\textwidth, height=  0.155\textwidth]{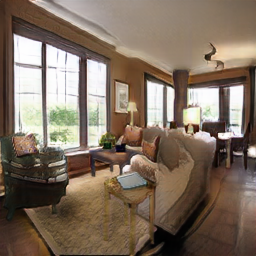}} \hfill
    \subfloat[]{\includegraphics[width=  0.195\textwidth, height=  0.155\textwidth]{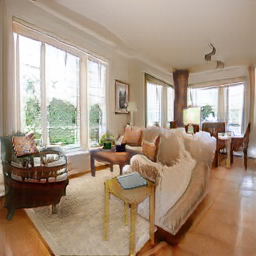}} \hfill
    \subfloat[]{\includegraphics[width=  0.195\textwidth, height=  0.155\textwidth]{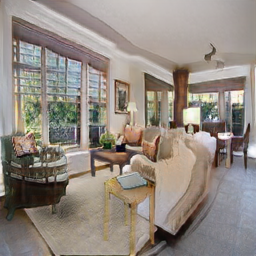}} \hfill
    \subfloat[]{\includegraphics[width=  0.195\textwidth, height=  0.155\textwidth]{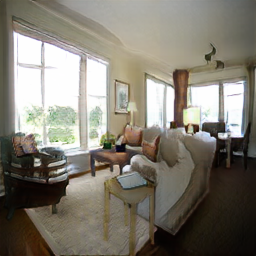}}
    \vfill
    
    \subfloat[]{\includegraphics[width=  0.195\textwidth, height=  0.155\textwidth]{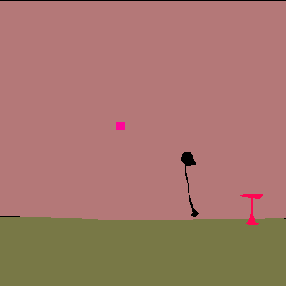}} \hfill
    \subfloat[]{\includegraphics[width=  0.195\textwidth, height=  0.155\textwidth]{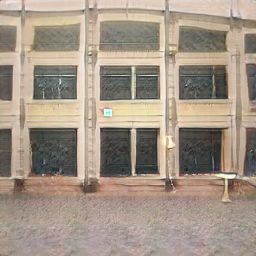}} \hfill
    \subfloat[]{\includegraphics[width=  0.195\textwidth, height=  0.155\textwidth]{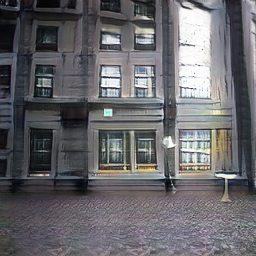}} \hfill
    \subfloat[]{\includegraphics[width=  0.195\textwidth, height=  0.155\textwidth]{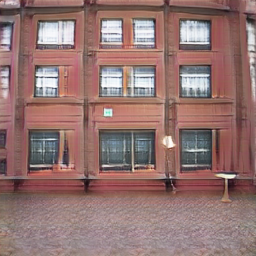}} \hfill
    \subfloat[]{\includegraphics[width=  0.195\textwidth, height=  0.155\textwidth]{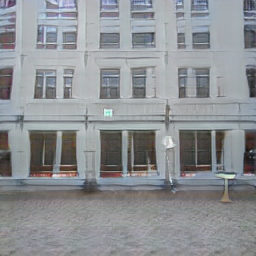}}
    \vfill
    
    \subfloat[]{\includegraphics[width=  0.195\textwidth, height=  0.155\textwidth]{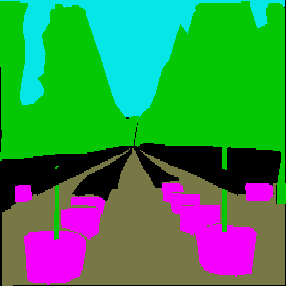}} \hfill
    \subfloat[]{\includegraphics[width=  0.195\textwidth, height=  0.155\textwidth]{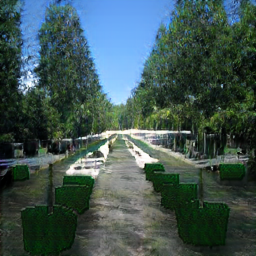}} \hfill
    \subfloat[]{\includegraphics[width=  0.195\textwidth, height=  0.155\textwidth]{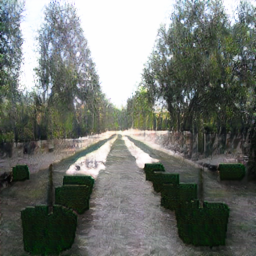}} \hfill
    \subfloat[]{\includegraphics[width=  0.195\textwidth, height=  0.155\textwidth]{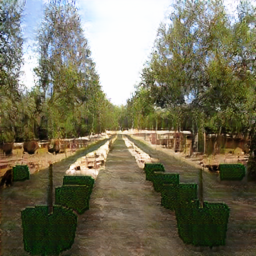}} \hfill
    \subfloat[]{\includegraphics[width=  0.195\textwidth, height=  0.155\textwidth]{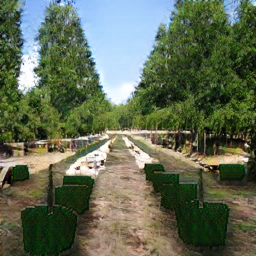}}
    \vfill
  
    \end{minipage}
    
   \caption{Multimodal Generation on ADE20K by resampling from the 3D noise}
   \label{fig:multimodal_Multimodal}
\end{figure*}

\begin{figure*}
    \captionsetup[subfloat]{position=top, labelformat=empty, skip=0pt}
    
    \centering
    \begin{minipage}{.80\textwidth}
    \subfloat[Label]{\includegraphics[width=  0.33\textwidth, height=   0.155\textwidth]{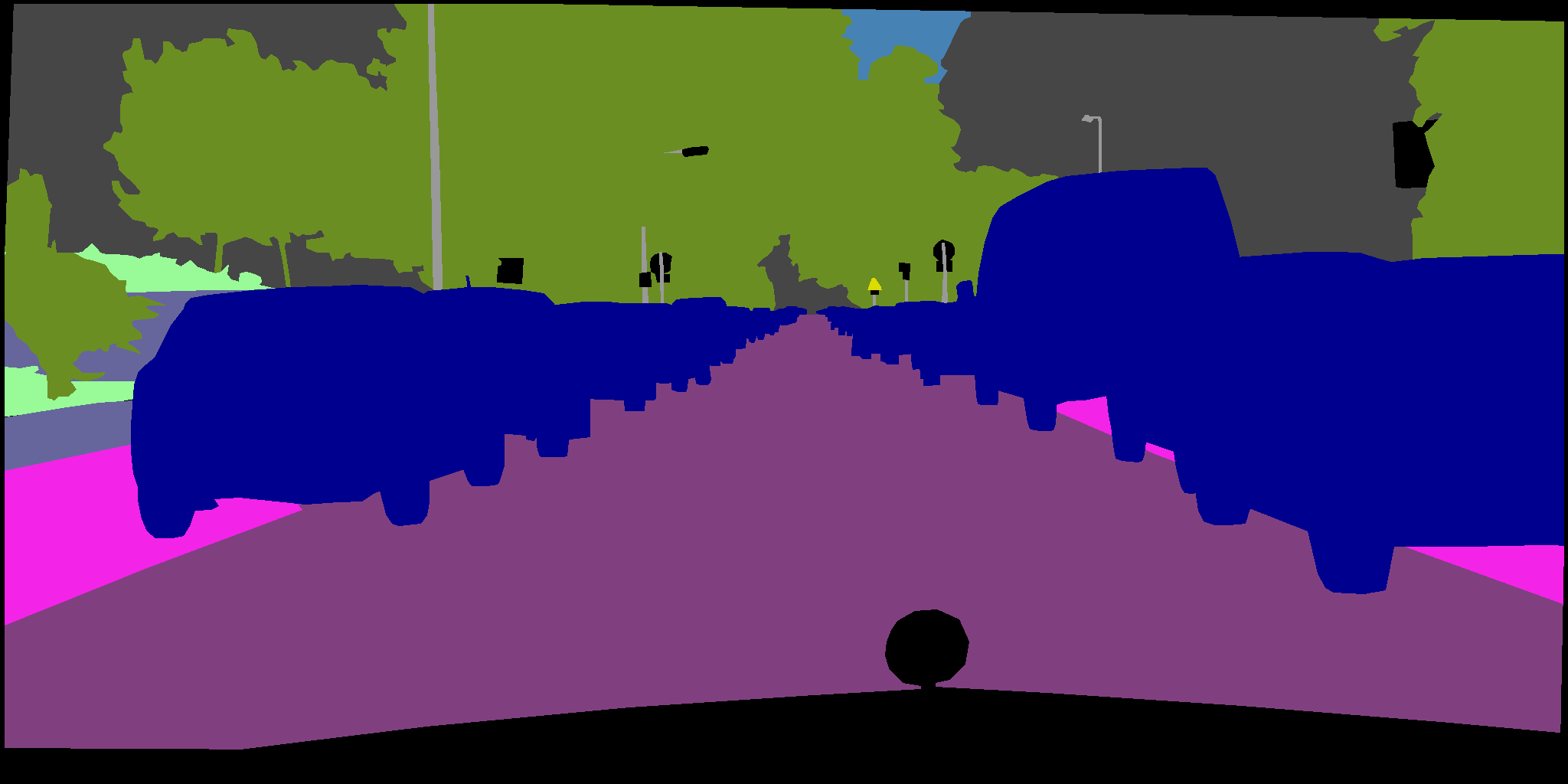}} \hfill
    \subfloat[DistanceGAN\cite{benaim2017one}]{\includegraphics[width=  0.33\textwidth, height=   0.155\textwidth]{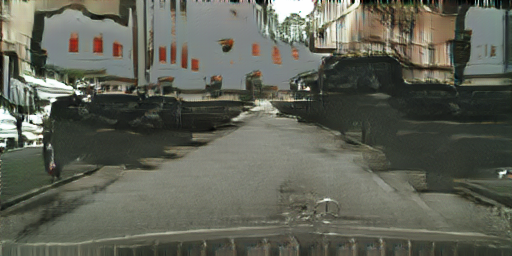}} \hfill
    \subfloat[CycleGAN\cite{zhu2017unpaired}]{\includegraphics[width=  0.33\textwidth, height=   0.155\textwidth]{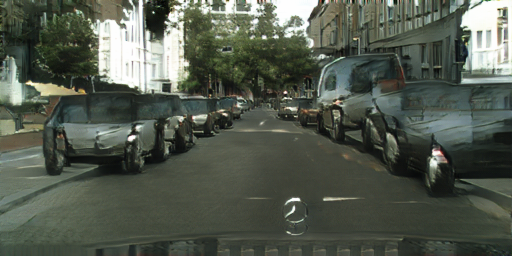}} \vfill
    \subfloat[CUT\cite{park2020cut}]{\includegraphics[width=  0.33\textwidth, height=   0.155\textwidth]{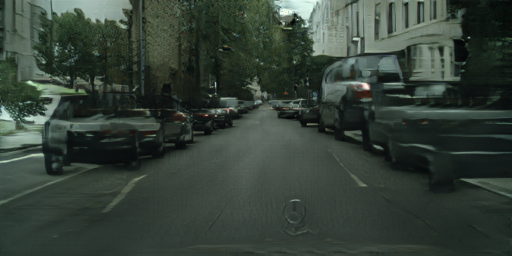}} \hfill
    \subfloat[USIS]{\includegraphics[width=  0.33\textwidth, height=   0.155\textwidth]{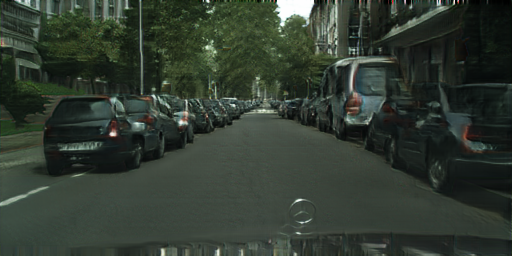}} \hfill
    \subfloat[Groundtruth]{\includegraphics[width=  0.33\textwidth, height=   0.155\textwidth]{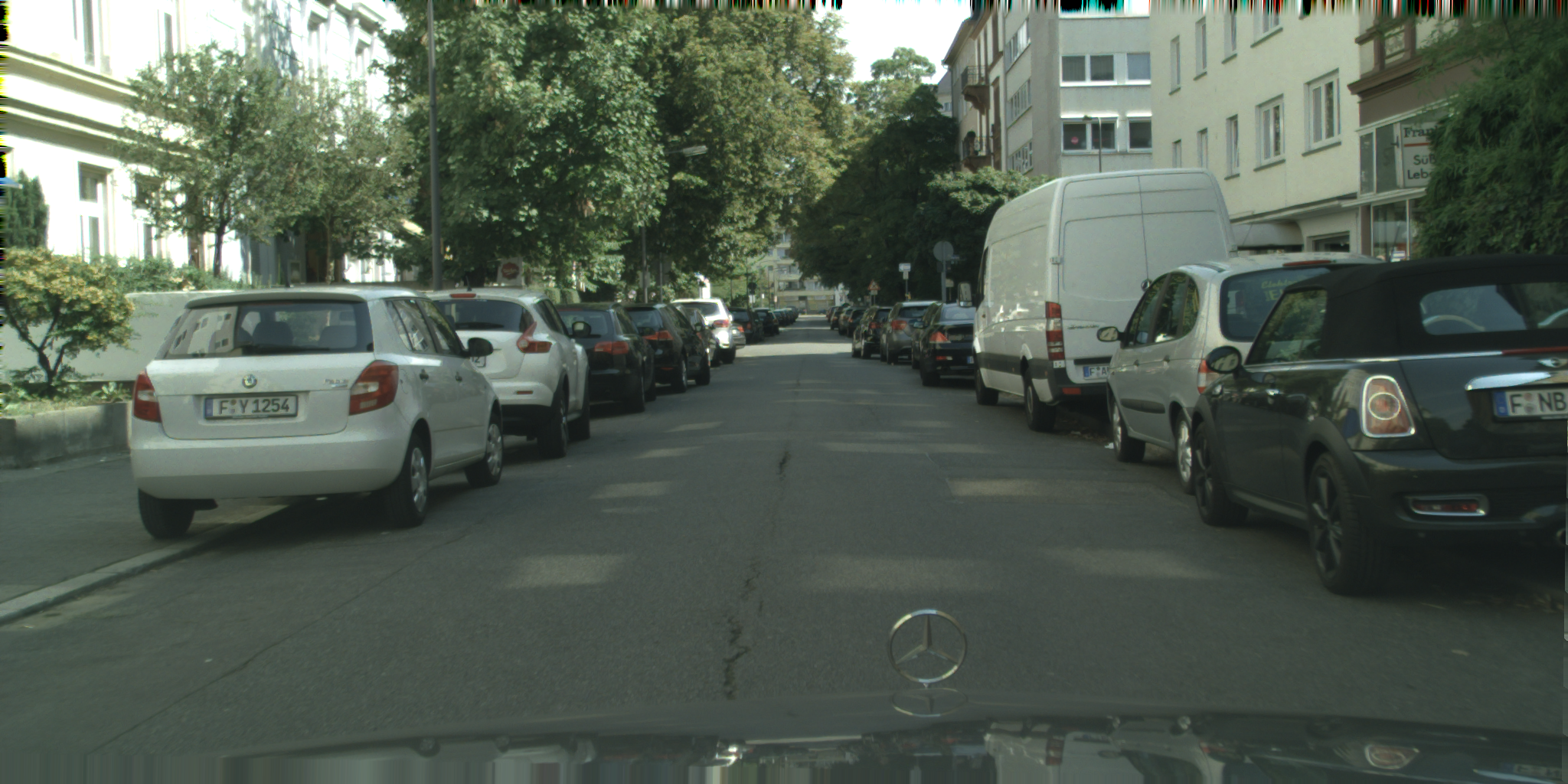}} 
    \end{minipage}
    
    \vspace{0.4cm}
    
    \begin{minipage}{.80\textwidth}
    \subfloat[Label]{\includegraphics[width=  0.33\textwidth, height=   0.155\textwidth]{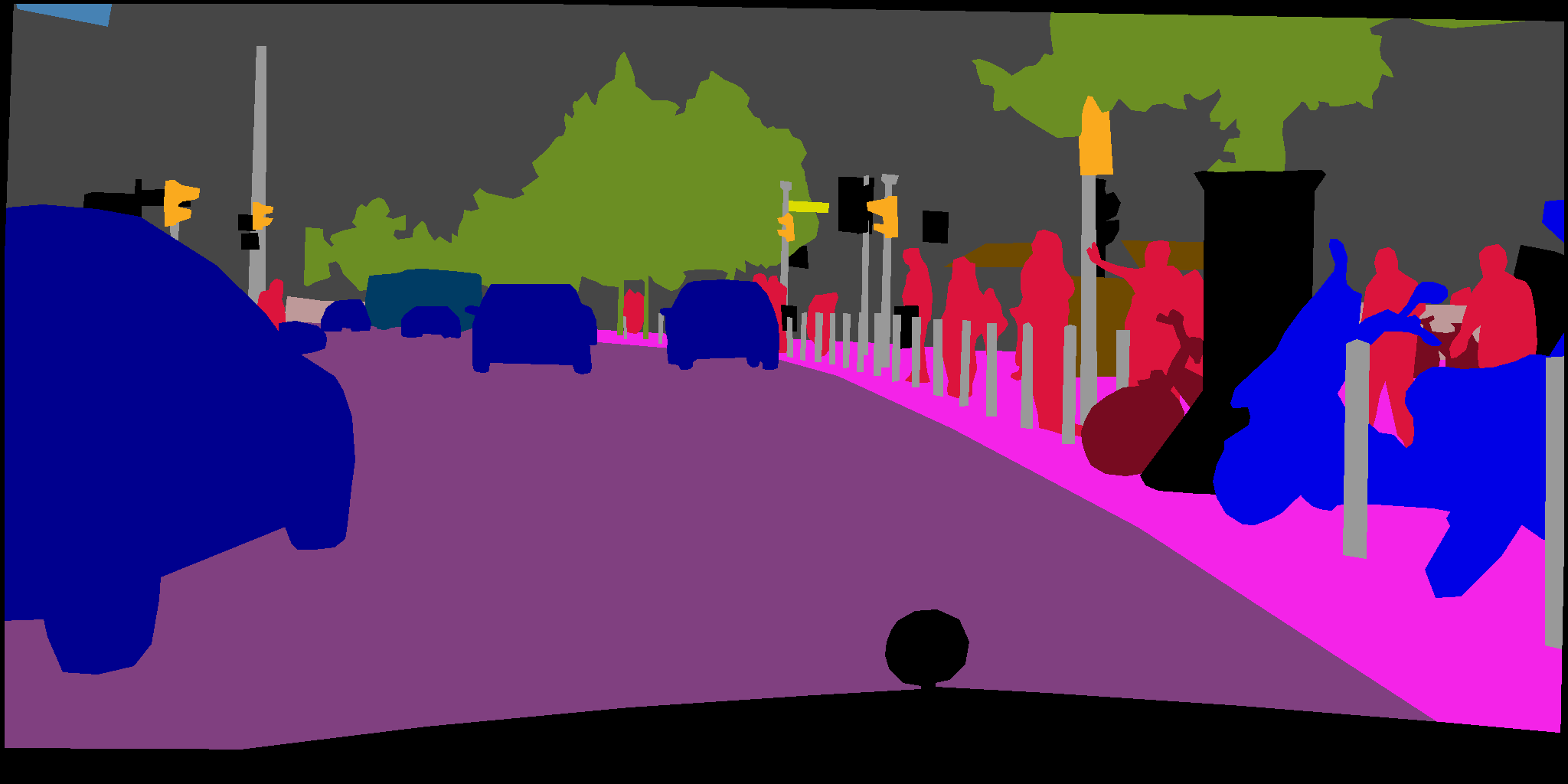}} \hfill
    \subfloat[DistanceGAN\cite{benaim2017one}]{\includegraphics[width=  0.33\textwidth, height=   0.155\textwidth]{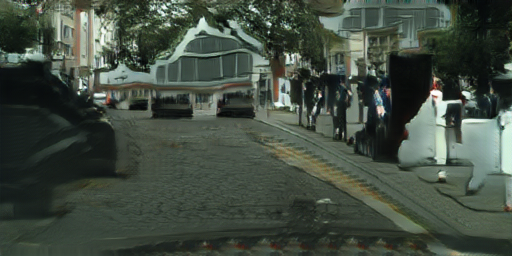}} \hfill
    \subfloat[CycleGAN\cite{zhu2017unpaired}]{\includegraphics[width=  0.33\textwidth, height=   0.155\textwidth]{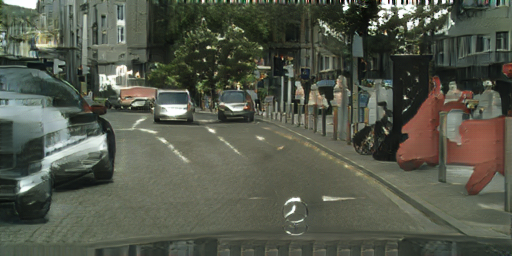}} \vfill
    \subfloat[CUT\cite{park2020cut}]{\includegraphics[width=  0.33\textwidth, height=   0.155\textwidth]{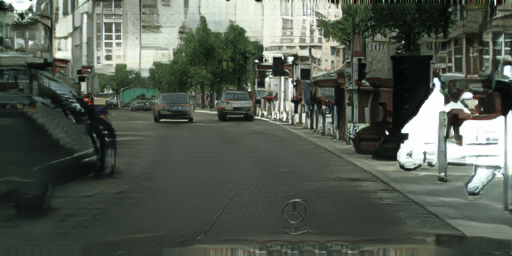}} \hfill
    \subfloat[USIS]{\includegraphics[width=  0.33\textwidth, height=   0.155\textwidth]{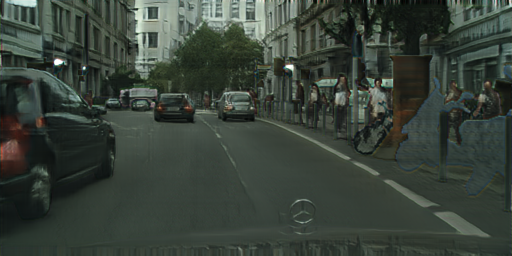}} \hfill
    \subfloat[Groundtruth]{\includegraphics[width=  0.33\textwidth, height=   0.155\textwidth]{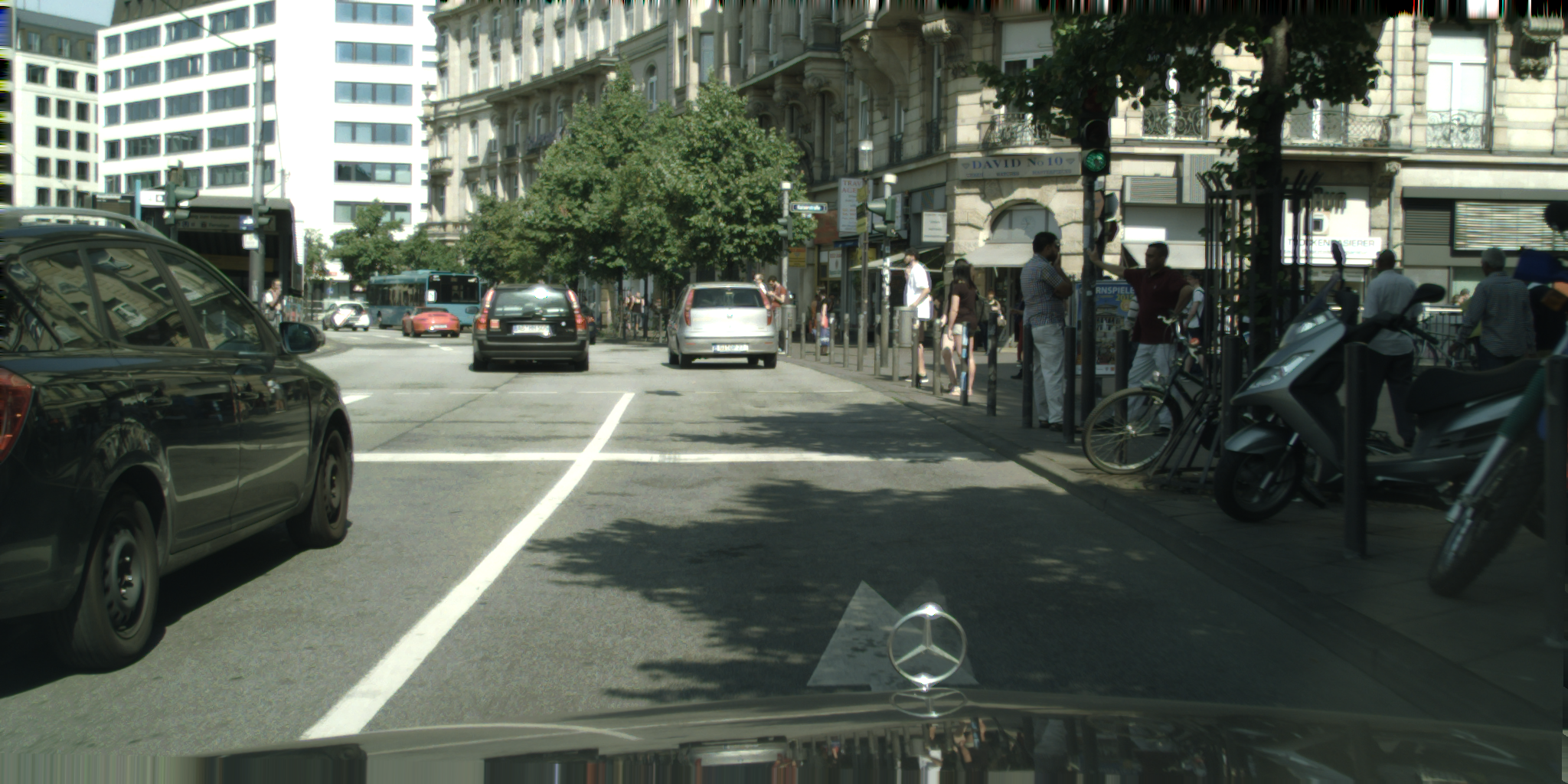}}
    \end{minipage}
    
    \vspace{0.4cm}
    
    \begin{minipage}{.80\textwidth}
    \subfloat[Label]{\includegraphics[width=  0.33\textwidth, height=   0.155\textwidth]{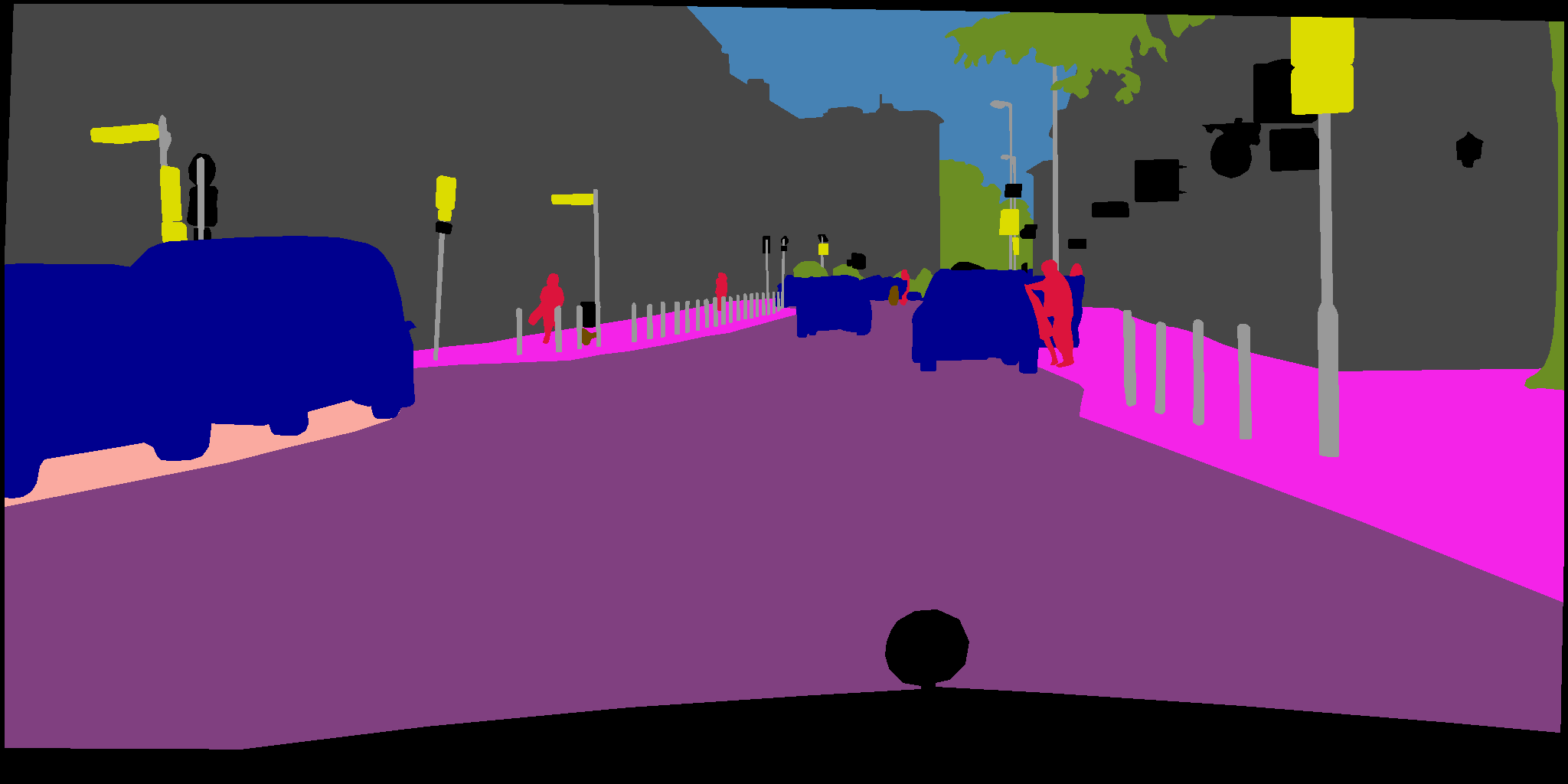}} \hfill
    \subfloat[DistanceGAN\cite{benaim2017one}]{\includegraphics[width=  0.33\textwidth, height=   0.155\textwidth]{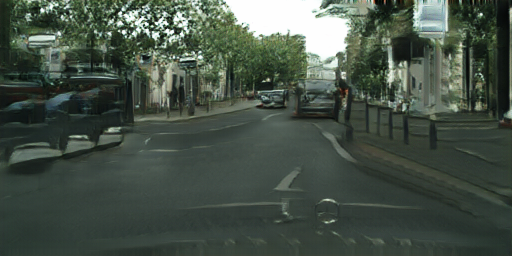}} \hfill
    \subfloat[CycleGAN\cite{zhu2017unpaired}]{\includegraphics[width=  0.33\textwidth, height=   0.155\textwidth]{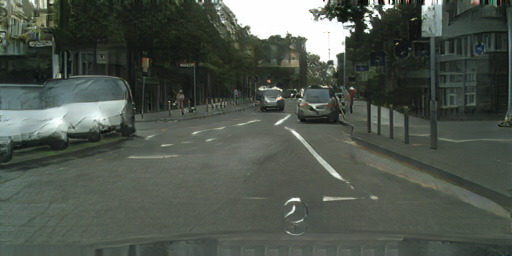}} \vfill
    \subfloat[CUT\cite{park2020cut}]{\includegraphics[width=  0.33\textwidth, height=   0.155\textwidth]{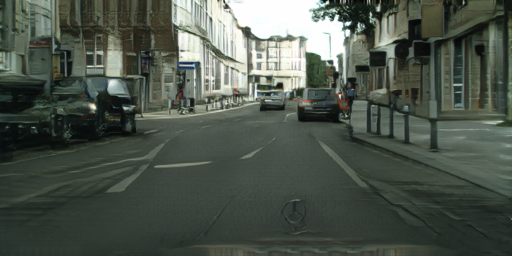}} \hfill
    \subfloat[USIS]{\includegraphics[width=  0.33\textwidth, height=   0.155\textwidth]{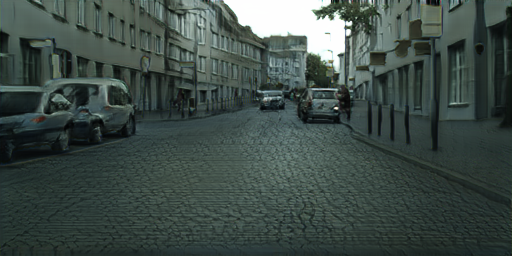}} \hfill
    \subfloat[Groundtruth]{\includegraphics[width=  0.33\textwidth, height=   0.155\textwidth]{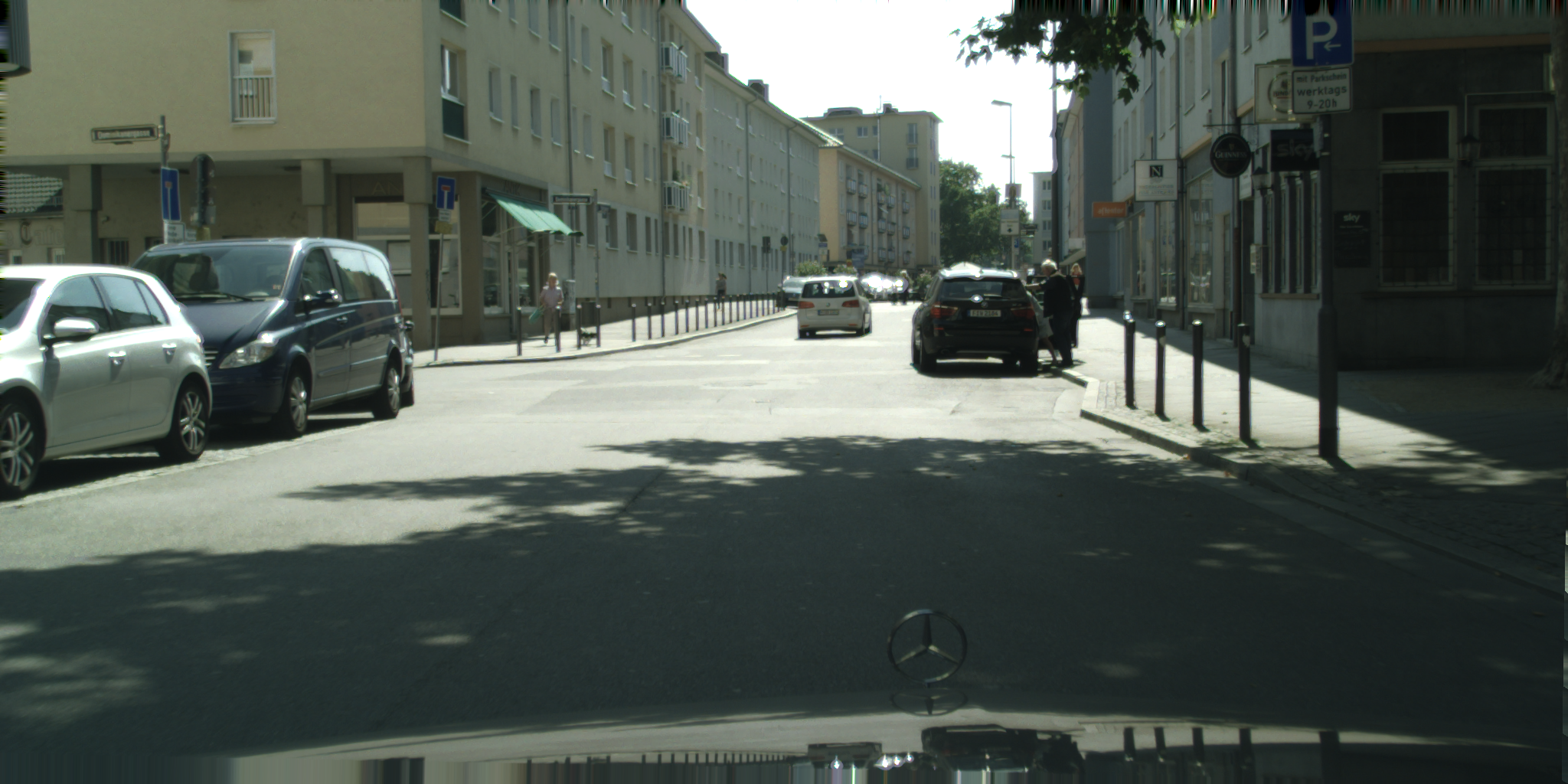}}
    \end{minipage}
    
    \vspace{0.4cm}
    
    \begin{minipage}{.80\textwidth}
    \subfloat[Label]{\includegraphics[width=  0.33\textwidth, height=   0.155\textwidth]{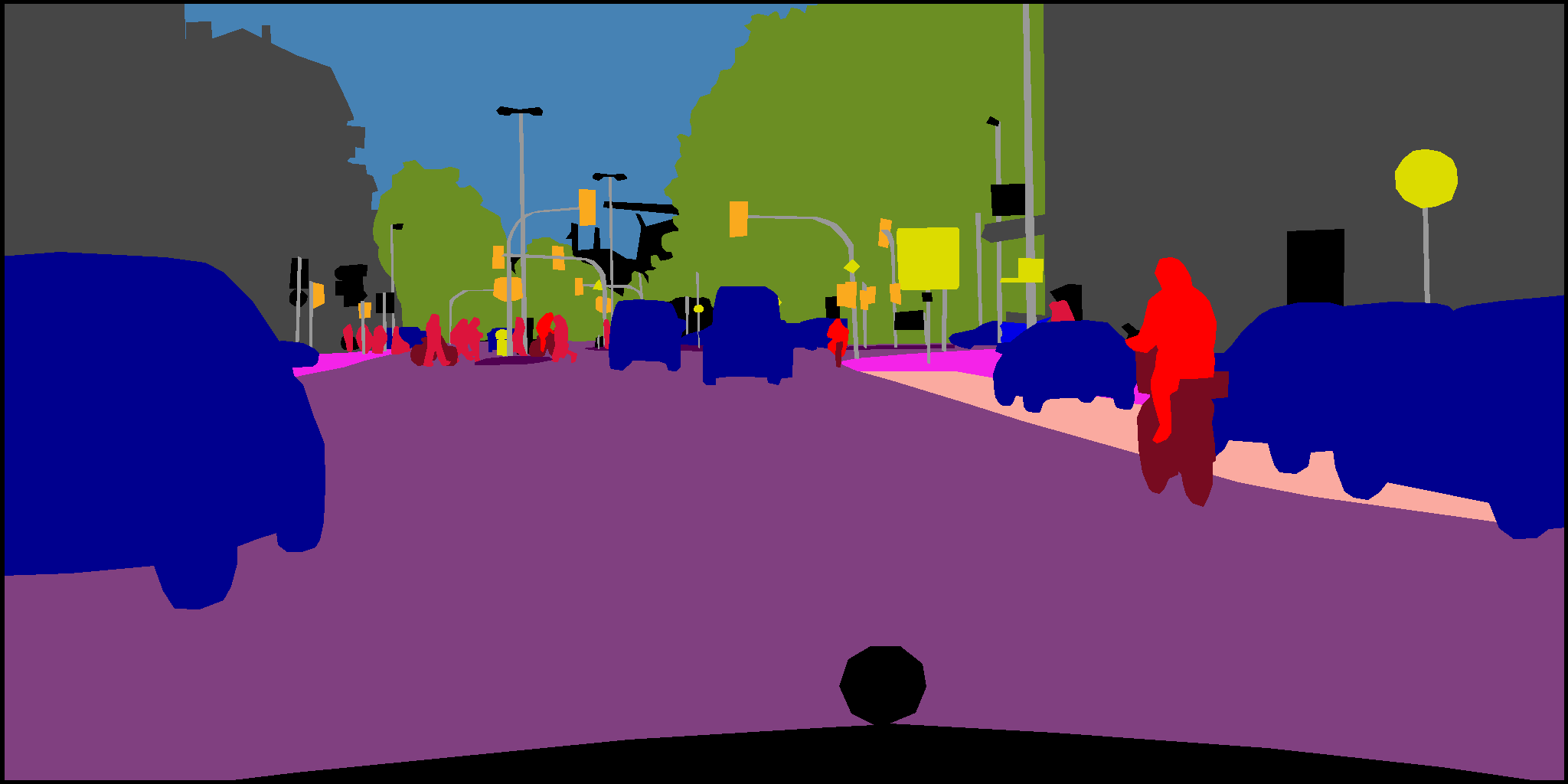}} \hfill
    \subfloat[DistanceGAN\cite{benaim2017one}]{\includegraphics[width=  0.33\textwidth, height=   0.155\textwidth]{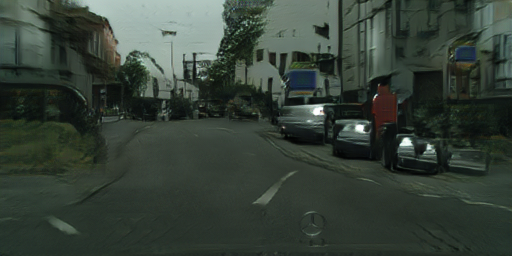}} \hfill
    \subfloat[CycleGAN\cite{zhu2017unpaired}]{\includegraphics[width=  0.33\textwidth, height=   0.155\textwidth]{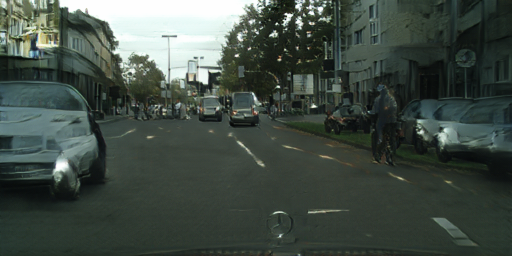}} \vfill
    \subfloat[CUT\cite{park2020cut}]{\includegraphics[width=  0.33\textwidth, height=   0.155\textwidth]{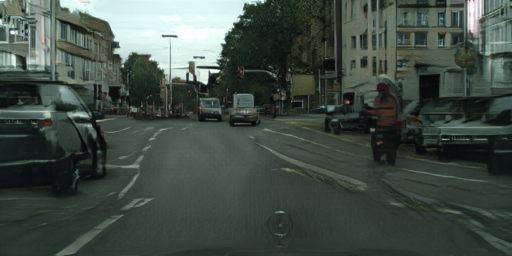}} \hfill
    \subfloat[USIS]{\includegraphics[width=  0.33\textwidth, height=   0.155\textwidth]{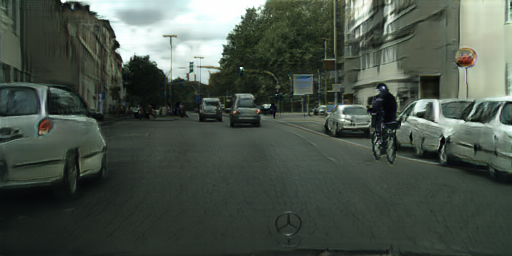}} \hfill
    \subfloat[Groundtruth]{\includegraphics[width=  0.33\textwidth, height=   0.155\textwidth]{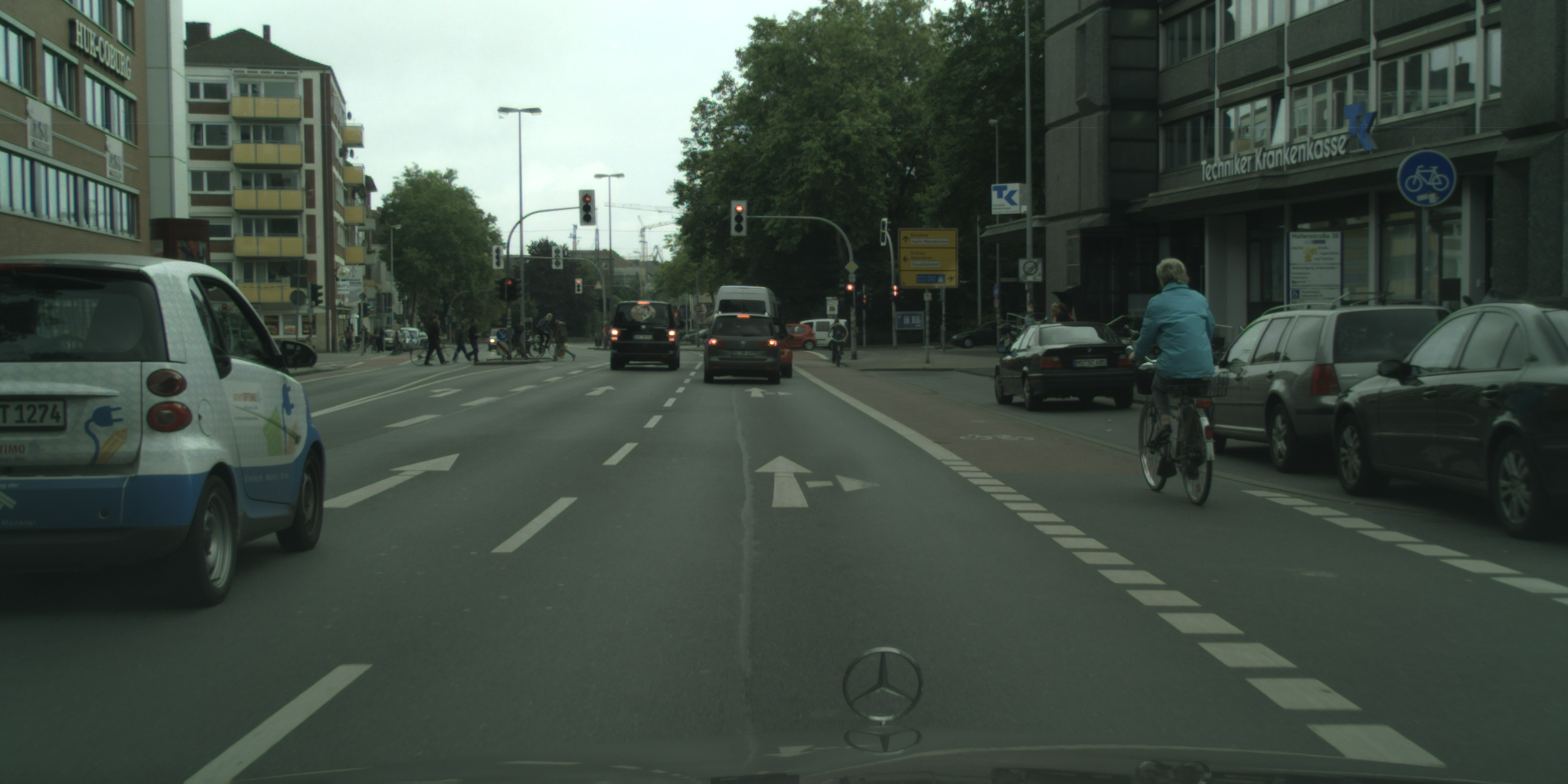}} 
    \end{minipage}
   \caption{Qualitative Comparison on Cityscapes Dataset}
   \label{fig:cityscapes_results}
\end{figure*}

\begin{figure*}
    \captionsetup[subfloat]{position=top, labelformat=empty, skip=0pt}
    
    \centering
    \begin{minipage}{0.7\textwidth}
    \subfloat[Label]{\includegraphics[width=  0.195\textwidth, height=  0.155\textwidth]{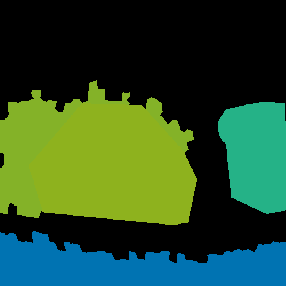}} \hfill
    \subfloat[CycleGAN\cite{zhu2017unpaired}]{\includegraphics[width=  0.195\textwidth, height=  0.155\textwidth]{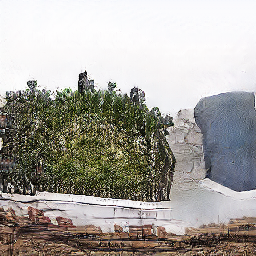}} \hfill
    \subfloat[CUT\cite{park2020cut}]{\includegraphics[width=  0.195\textwidth, height=  0.155\textwidth]{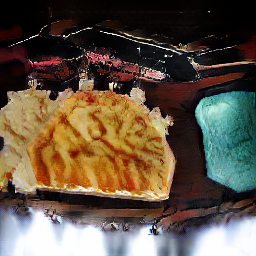}} \hfill
    \subfloat[USIS]{\includegraphics[width=  0.195\textwidth, height=  0.155\textwidth]{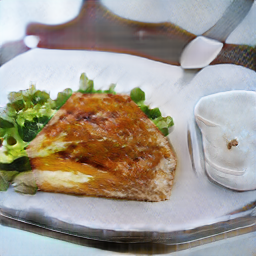}} \hfill
    \subfloat[Groundtruth]{\includegraphics[width=  0.195\textwidth, height=  0.155\textwidth]{figures/cocostuff/gt_1.jpg}}
    \vfill
    
    \subfloat[]{\includegraphics[width=  0.195\textwidth, height=  0.155\textwidth]{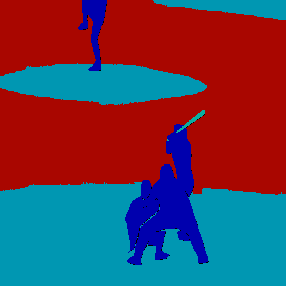}} \hfill
    \subfloat[]{\includegraphics[width=  0.195\textwidth, height=  0.155\textwidth]{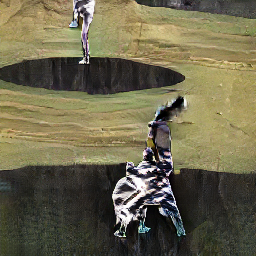}} \hfill
    \subfloat[]{\includegraphics[width=  0.195\textwidth, height=  0.155\textwidth]{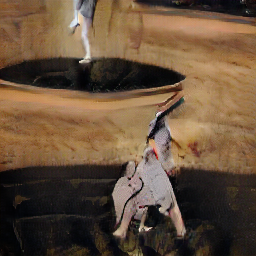}} \hfill
    \subfloat[]{\includegraphics[width=  0.195\textwidth, height=  0.155\textwidth]{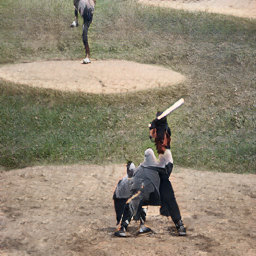}} \hfill
    \subfloat[]{\includegraphics[width=  0.195\textwidth, height=  0.155\textwidth]{figures/cocostuff/gt_2.jpg}}
    \vfill
    
    \subfloat[]{\includegraphics[width=  0.195\textwidth, height=  0.155\textwidth]{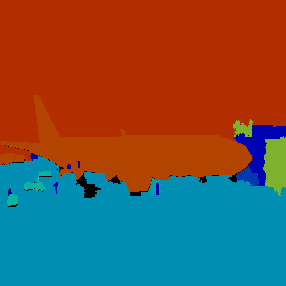}} \hfill
    \subfloat[]{\includegraphics[width=  0.195\textwidth, height=  0.155\textwidth]{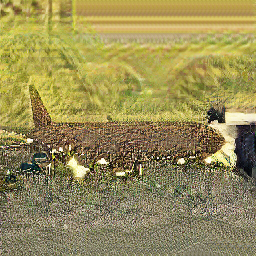}} \hfill
    \subfloat[]{\includegraphics[width=  0.195\textwidth, height=  0.155\textwidth]{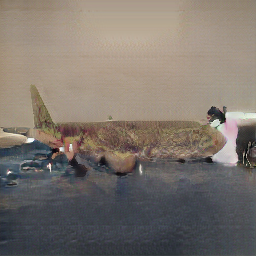}} \hfill
    \subfloat[]{\includegraphics[width=  0.195\textwidth, height=  0.155\textwidth]{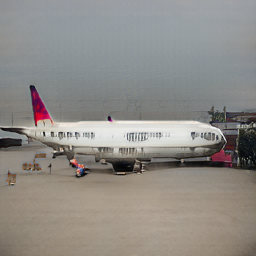}} \hfill
    \subfloat[]{\includegraphics[width=  0.195\textwidth, height=  0.155\textwidth]{figures/cocostuff/gt_3.jpg}}
    \vfill
    
    \subfloat[]{\includegraphics[width=  0.195\textwidth, height=  0.155\textwidth]{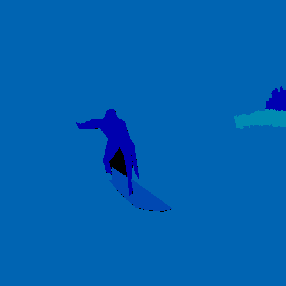}} \hfill
    \subfloat[]{\includegraphics[width=  0.195\textwidth, height=  0.155\textwidth]{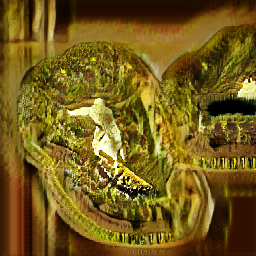}} \hfill
    \subfloat[]{\includegraphics[width=  0.195\textwidth, height=  0.155\textwidth]{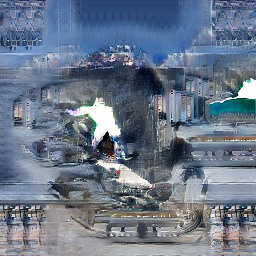}} \hfill
    \subfloat[]{\includegraphics[width=  0.195\textwidth, height=  0.155\textwidth]{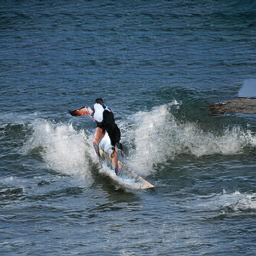}} \hfill
    \subfloat[]{\includegraphics[width=  0.195\textwidth, height=  0.155\textwidth]{figures/cocostuff/gt_4.jpg}}
    \vfill
    
    \subfloat[]{\includegraphics[width=  0.195\textwidth, height=  0.155\textwidth]{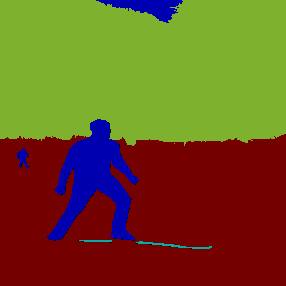}} \hfill
    \subfloat[]{\includegraphics[width=  0.195\textwidth, height=  0.155\textwidth]{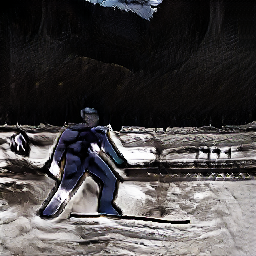}} \hfill
    \subfloat[]{\includegraphics[width=  0.195\textwidth, height=  0.155\textwidth]{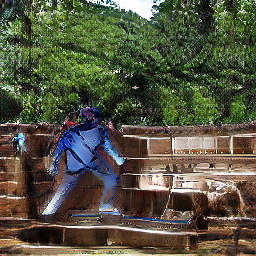}} \hfill
    \subfloat[]{\includegraphics[width=  0.195\textwidth, height=  0.155\textwidth]{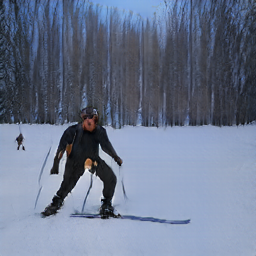}} \hfill
    \subfloat[]{\includegraphics[width=  0.195\textwidth, height=  0.155\textwidth]{figures/cocostuff/gt_5.jpg}}
    \vfill
    
    \subfloat[]{\includegraphics[width=  0.195\textwidth, height=  0.155\textwidth]{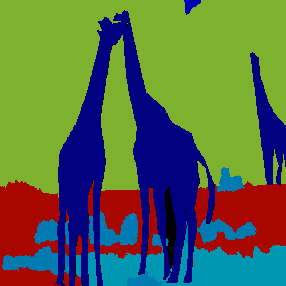}} \hfill
    \subfloat[]{\includegraphics[width=  0.195\textwidth, height=  0.155\textwidth]{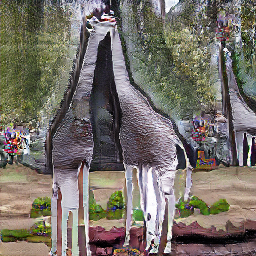}} \hfill
    \subfloat[]{\includegraphics[width=  0.195\textwidth, height=  0.155\textwidth]{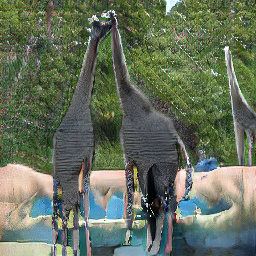}} \hfill
    \subfloat[]{\includegraphics[width=  0.195\textwidth, height=  0.155\textwidth]{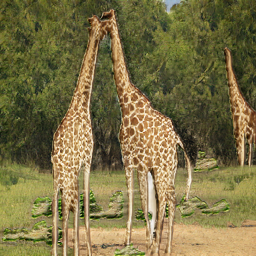}} \hfill
    \subfloat[]{\includegraphics[width=  0.195\textwidth, height=  0.155\textwidth]{figures/cocostuff/gt_6.jpg}}
    \vfill
    
    \subfloat[]{\includegraphics[width=  0.195\textwidth, height=  0.155\textwidth]{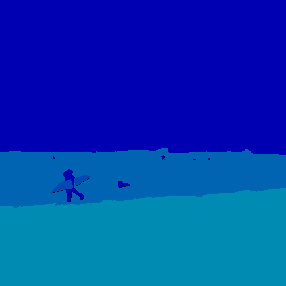}} \hfill
    \subfloat[]{\includegraphics[width=  0.195\textwidth, height=  0.155\textwidth]{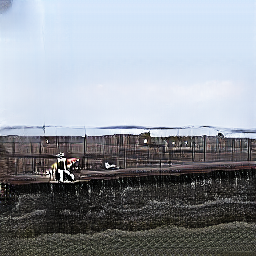}} \hfill
    \subfloat[]{\includegraphics[width=  0.195\textwidth, height=  0.155\textwidth]{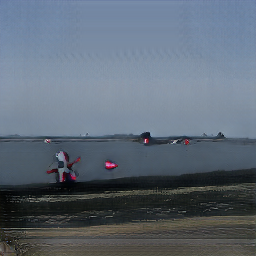}} \hfill
    \subfloat[]{\includegraphics[width=  0.195\textwidth, height=  0.155\textwidth]{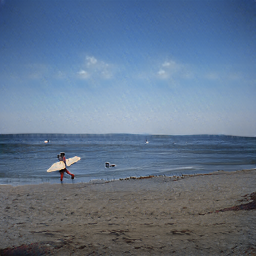}} \hfill
    \subfloat[]{\includegraphics[width=  0.195\textwidth, height=  0.155\textwidth]{figures/cocostuff/gt_7.jpg}}
    \vfill
    
    \subfloat[]{\includegraphics[width=  0.195\textwidth, height=  0.155\textwidth]{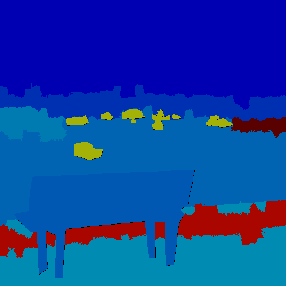}} \hfill
    \subfloat[]{\includegraphics[width=  0.195\textwidth, height=  0.155\textwidth]{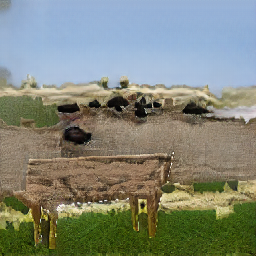}} \hfill
    \subfloat[]{\includegraphics[width=  0.195\textwidth, height=  0.155\textwidth]{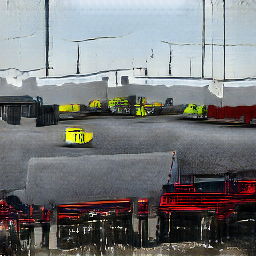}} \hfill
    \subfloat[]{\includegraphics[width=  0.195\textwidth, height=  0.155\textwidth]{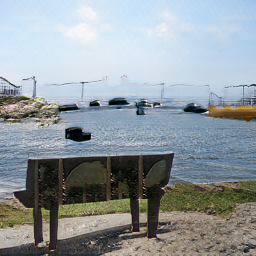}} \hfill
    \subfloat[]{\includegraphics[width=  0.195\textwidth, height=  0.155\textwidth]{figures/cocostuff/gt_8.jpg}}
    \vfill
    
    \subfloat[]{\includegraphics[width=  0.195\textwidth, height=  0.155\textwidth]{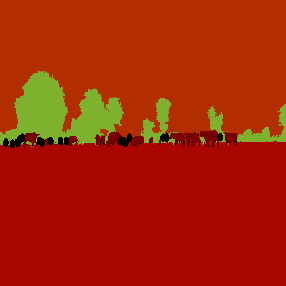}} \hfill
    \subfloat[]{\includegraphics[width=  0.195\textwidth, height=  0.155\textwidth]{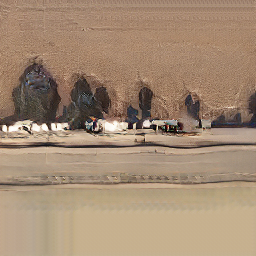}} \hfill
    \subfloat[]{\includegraphics[width=  0.195\textwidth, height=  0.155\textwidth]{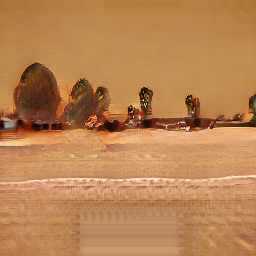}} \hfill
    \subfloat[]{\includegraphics[width=  0.195\textwidth, height=  0.155\textwidth]{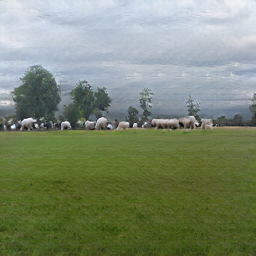}} \hfill
    \subfloat[]{\includegraphics[width=  0.195\textwidth, height=  0.155\textwidth]{figures/cocostuff/gt_9.jpg}}
    \vfill
    
    \subfloat[]{\includegraphics[width=  0.195\textwidth, height=  0.155\textwidth]{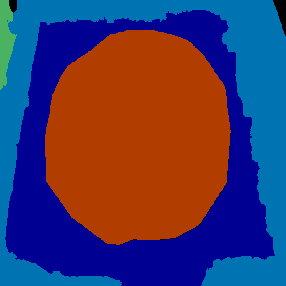}} \hfill
    \subfloat[]{\includegraphics[width=  0.195\textwidth, height=  0.155\textwidth]{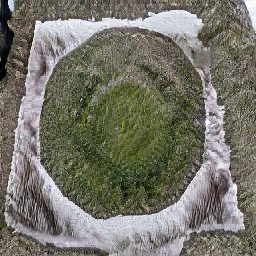}} \hfill
    \subfloat[]{\includegraphics[width=  0.195\textwidth, height=  0.155\textwidth]{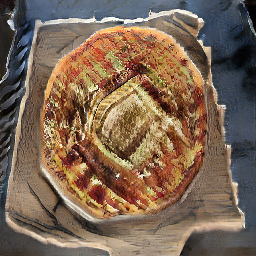}} \hfill
    \subfloat[]{\includegraphics[width=  0.195\textwidth, height=  0.155\textwidth]{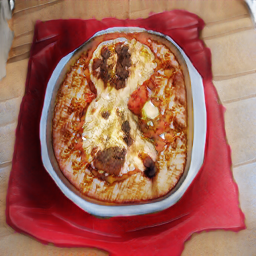}} \hfill
    \subfloat[]{\includegraphics[width=  0.195\textwidth, height=  0.155\textwidth]{figures/cocostuff/gt_10.jpg}}
    \vfill

    \end{minipage}
    
   \caption{Qualitative Comparison on Cocostuff Dataset}
   \label{fig:cocostuff_results}
\end{figure*}

\begin{figure*}
    \captionsetup[subfloat]{position=top, labelformat=empty, skip=0pt}
    \centering
    \begin{minipage}{0.7\textwidth}
    \subfloat[Label]{\includegraphics[width=  0.195\textwidth, height=  0.155\textwidth]{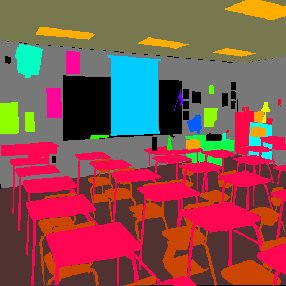}} \hfill
    \subfloat[CycleGAN\cite{zhu2017unpaired}]{\includegraphics[width=  0.195\textwidth, height=  0.155\textwidth]{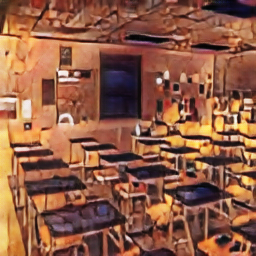}} \hfill
    \subfloat[CUT\cite{park2020cut}]{\includegraphics[width=  0.195\textwidth, height=  0.155\textwidth]{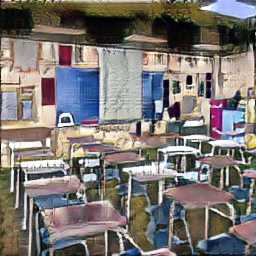}} \hfill
    \subfloat[USIS]{\includegraphics[width=  0.195\textwidth, height=  0.155\textwidth]{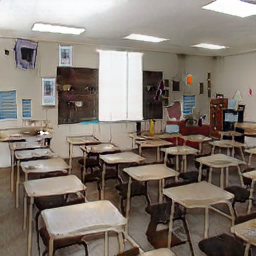}} \hfill
    \subfloat[Groundtruth]{\includegraphics[width=  0.195\textwidth, height=  0.155\textwidth]{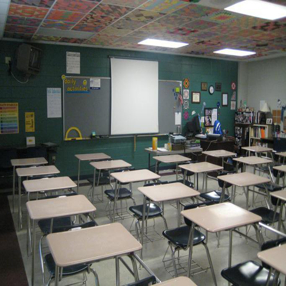}}
    \vfill
    
    \subfloat[]{\includegraphics[width=  0.195\textwidth, height=  0.155\textwidth]{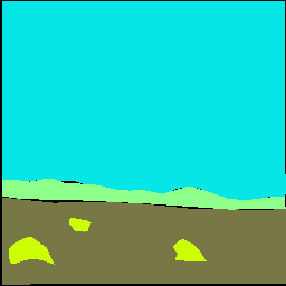}} \hfill
    \subfloat[]{\includegraphics[width=  0.195\textwidth, height=  0.155\textwidth]{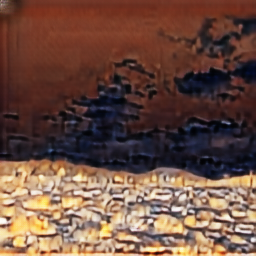}} \hfill
    \subfloat[]{\includegraphics[width=  0.195\textwidth, height=  0.155\textwidth]{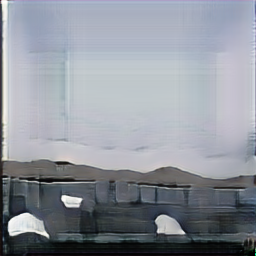}} \hfill
    \subfloat[]{\includegraphics[width=  0.195\textwidth, height=  0.155\textwidth]{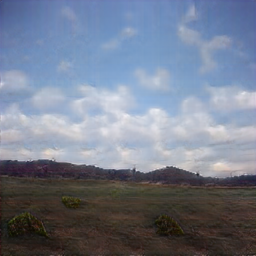}} \hfill
    \subfloat[]{\includegraphics[width=  0.195\textwidth, height=  0.155\textwidth]{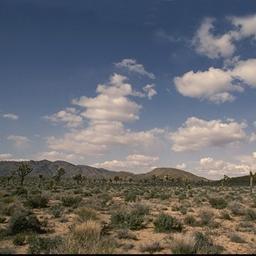}}
    \vfill
    
    \subfloat[]{\includegraphics[width=  0.195\textwidth, height=  0.155\textwidth]{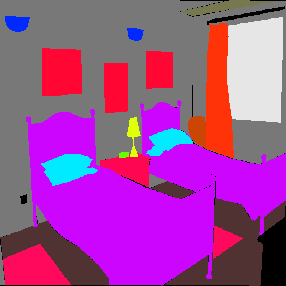}} \hfill
    \subfloat[]{\includegraphics[width=  0.195\textwidth, height=  0.155\textwidth]{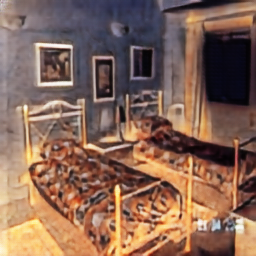}} \hfill
    \subfloat[]{\includegraphics[width=  0.195\textwidth, height=  0.155\textwidth]{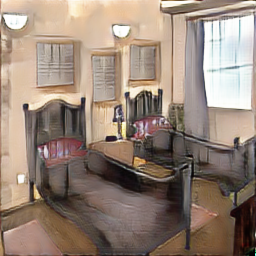}} \hfill
    \subfloat[]{\includegraphics[width=  0.195\textwidth, height=  0.155\textwidth]{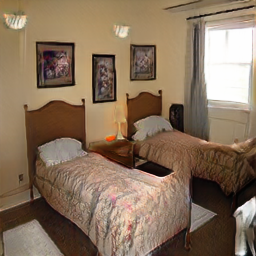}} \hfill
    \subfloat[]{\includegraphics[width=  0.195\textwidth, height=  0.155\textwidth]{figures/ade20k/gt_13.jpg}}
    \vfill
    
    \subfloat[]{\includegraphics[width=  0.195\textwidth, height=  0.155\textwidth]{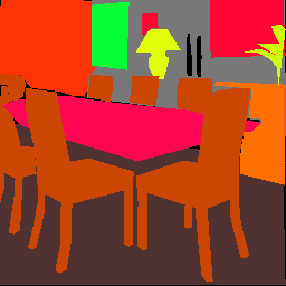}} \hfill
    \subfloat[]{\includegraphics[width=  0.195\textwidth, height=  0.155\textwidth]{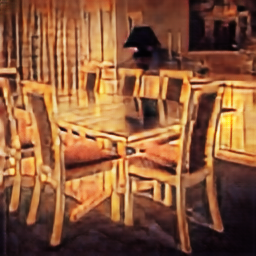}} \hfill
    \subfloat[]{\includegraphics[width=  0.195\textwidth, height=  0.155\textwidth]{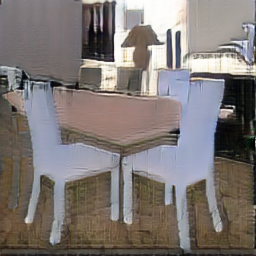}} \hfill
    \subfloat[]{\includegraphics[width=  0.195\textwidth, height=  0.155\textwidth]{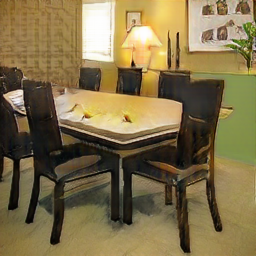}} \hfill
    \subfloat[]{\includegraphics[width=  0.195\textwidth, height=  0.155\textwidth]{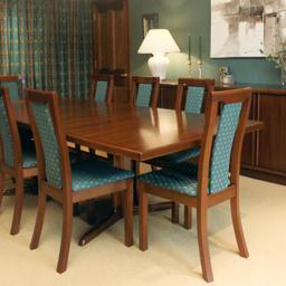}}
    \vfill
    
    \subfloat[]{\includegraphics[width=  0.195\textwidth, height=  0.155\textwidth]{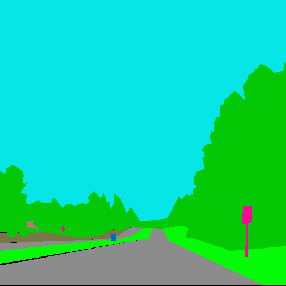}} \hfill
    \subfloat[]{\includegraphics[width=  0.195\textwidth, height=  0.155\textwidth]{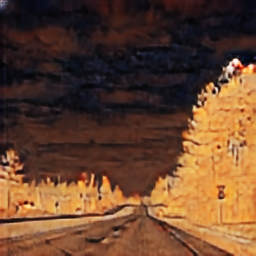}} \hfill
    \subfloat[]{\includegraphics[width=  0.195\textwidth, height=  0.155\textwidth]{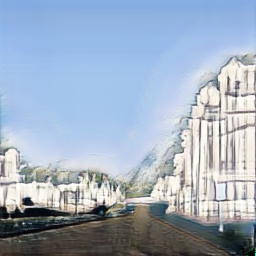}} \hfill
    \subfloat[]{\includegraphics[width=  0.195\textwidth, height=  0.155\textwidth]{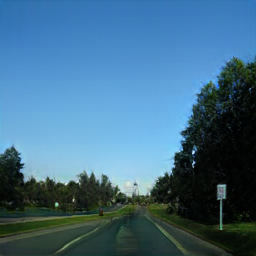}} \hfill
    \subfloat[]{\includegraphics[width=  0.195\textwidth, height=  0.155\textwidth]{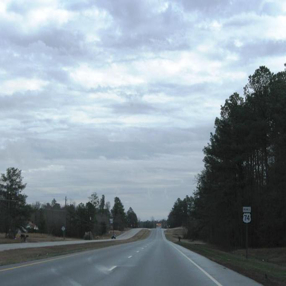}}
    \vfill
    
    \subfloat[]{\includegraphics[width=  0.195\textwidth, height=  0.155\textwidth]{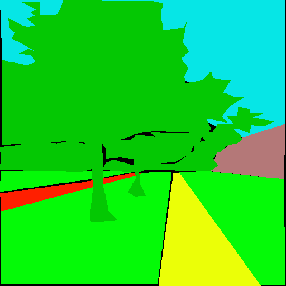}} \hfill
    \subfloat[]{\includegraphics[width=  0.195\textwidth, height=  0.155\textwidth]{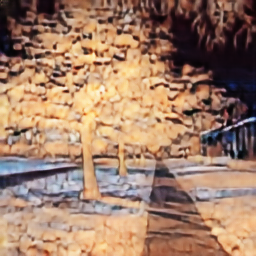}} \hfill
    \subfloat[]{\includegraphics[width=  0.195\textwidth, height=  0.155\textwidth]{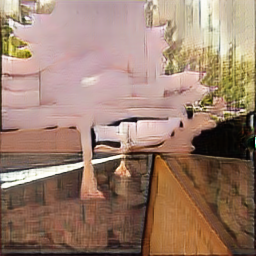}} \hfill
    \subfloat[]{\includegraphics[width=  0.195\textwidth, height=  0.155\textwidth]{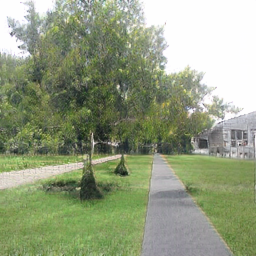}} \hfill
    \subfloat[]{\includegraphics[width=  0.195\textwidth, height=  0.155\textwidth]{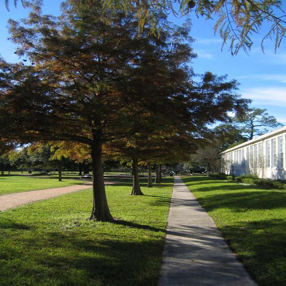}}
    \vfill
    
    \subfloat[]{\includegraphics[width=  0.195\textwidth, height=  0.155\textwidth]{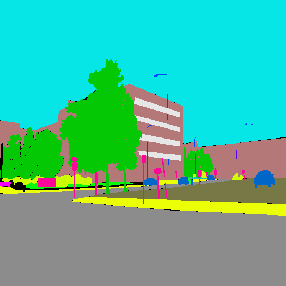}} \hfill
    \subfloat[]{\includegraphics[width=  0.195\textwidth, height=  0.155\textwidth]{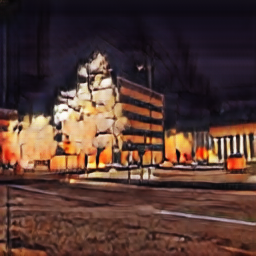}} \hfill
    \subfloat[]{\includegraphics[width=  0.195\textwidth, height=  0.155\textwidth]{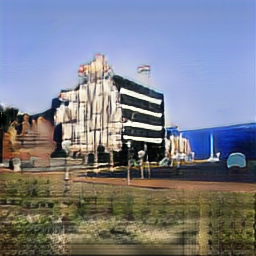}} \hfill
    \subfloat[]{\includegraphics[width=  0.195\textwidth, height=  0.155\textwidth]{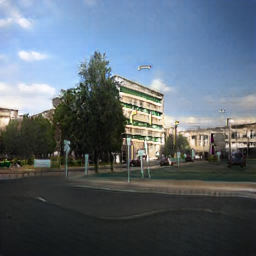}} \hfill
    \subfloat[]{\includegraphics[width=  0.195\textwidth, height=  0.155\textwidth]{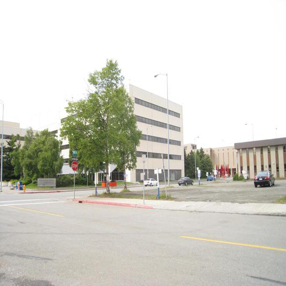}}
    \vfill
    
    \subfloat[]{\includegraphics[width=  0.195\textwidth, height=  0.155\textwidth]{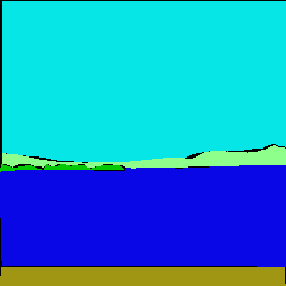}} \hfill
    \subfloat[]{\includegraphics[width=  0.195\textwidth, height=  0.155\textwidth]{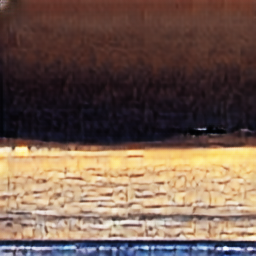}} \hfill
    \subfloat[]{\includegraphics[width=  0.195\textwidth, height=  0.155\textwidth]{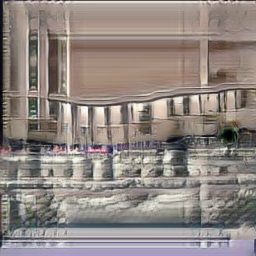}} \hfill
    \subfloat[]{\includegraphics[width=  0.195\textwidth, height=  0.155\textwidth]{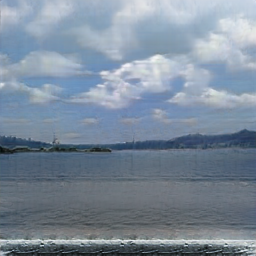}} \hfill
    \subfloat[]{\includegraphics[width=  0.195\textwidth, height=  0.155\textwidth]{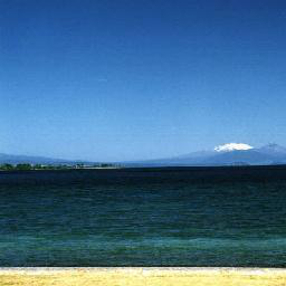}}
    \vfill
    
    \subfloat[]{\includegraphics[width=  0.195\textwidth, height=  0.155\textwidth]{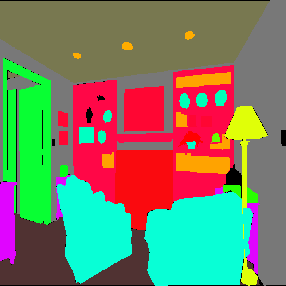}} \hfill
    \subfloat[]{\includegraphics[width=  0.195\textwidth, height=  0.155\textwidth]{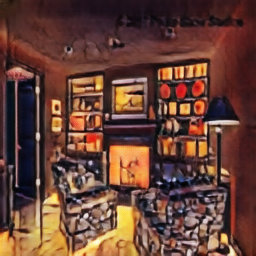}} \hfill
    \subfloat[]{\includegraphics[width=  0.195\textwidth, height=  0.155\textwidth]{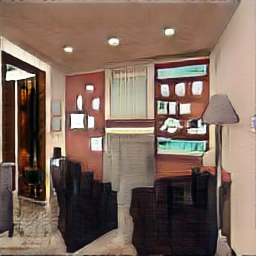}} \hfill
    \subfloat[]{\includegraphics[width=  0.195\textwidth, height=  0.155\textwidth]{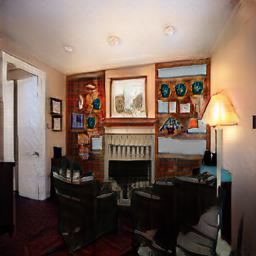}} \hfill
    \subfloat[]{\includegraphics[width=  0.195\textwidth, height=  0.155\textwidth]{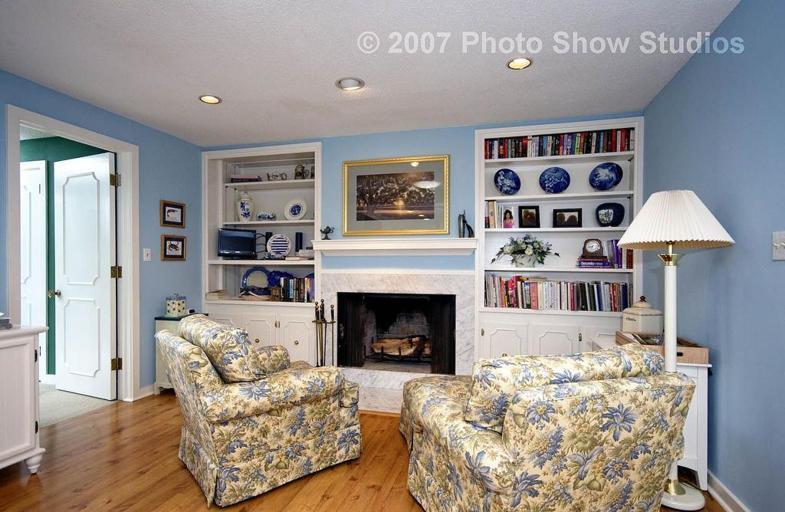}}
    \vfill
    
    \subfloat[]{\includegraphics[width=  0.195\textwidth, height=  0.155\textwidth]{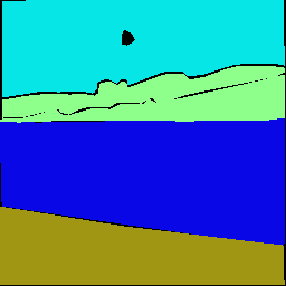}} \hfill
    \subfloat[]{\includegraphics[width=  0.195\textwidth, height=  0.155\textwidth]{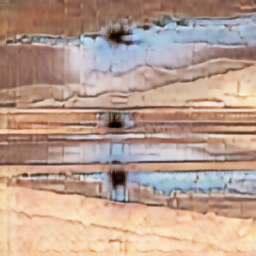}} \hfill
    \subfloat[]{\includegraphics[width=  0.195\textwidth, height=  0.155\textwidth]{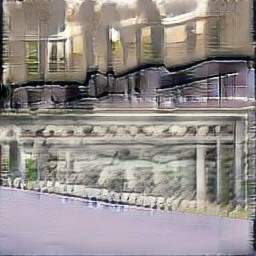}} \hfill
    \subfloat[]{\includegraphics[width=  0.195\textwidth, height=  0.155\textwidth]{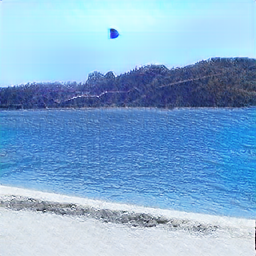}} \hfill
    \subfloat[]{\includegraphics[width=  0.195\textwidth, height=  0.155\textwidth]{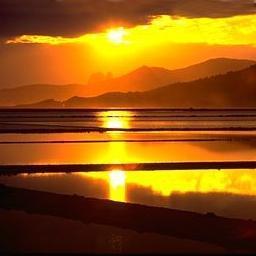}}
    \vfill

    \end{minipage}
    
   \caption{Qualitative Comparison on ADE20K Dataset}
   \label{fig:ade20k_results}
\end{figure*}

In Figure \ref{fig:multimodal_Multimodal}, we showcase the ability of our model to generate multimodal images by sampling several times from the 3D noise at the input of the generator. We perform this experiment on ADE20K. In Figures \ref{fig:cityscapes_results}, \ref{fig:ade20k_results} and \ref{fig:cocostuff_results}, we show more qualitative results of our model against other baselines, on Cityscapes, ADE20K and Cocostuff respectively.

\end{appendices}
\end{sloppypar}
\end{document}